\documentclass[11pt]{article}

\usepackage[final]{acl}

\usepackage{times}
\usepackage{latexsym}
\usepackage{amsmath}
\usepackage{booktabs}
\usepackage{multirow}

\usepackage{hyperref}
\usepackage{url}

\usepackage{multicol}

\usepackage[table]{xcolor}
\usepackage{colortbl}

\usepackage{array}
\usepackage{makecell}
\usepackage{siunitx}
\usepackage{tabularx}

\usepackage{pgfplots}
\pgfplotsset{compat=1.18}
\usepackage{rotating}

\usepackage{algorithm}
\usepackage{algpseudocode}

\usepackage[most]{tcolorbox}
\usepackage{subcaption}
\usepackage{pifont}

\usepackage{caption}
\usepackage{amssymb}

\newcommand{\method}{\textsc{Tale }}
\usepackage[T1]{fontenc}
\DeclareFontShape{T1}{ptm}{m}{scit}{<->ssub*ptm/m/sc}{}

\usepackage[utf8]{inputenc}

\usepackage{microtype}

\usepackage{inconsolata}
\providecommand{\hidden}[1]{}
\usepackage{graphicx}

\title{TELL-TALE:  Task Efficient LLMs with Task Aware Layer Elimination}

\author{
Omar Naim\thanks{Equal contribution} \\
IRIT \\
{\small Université de Toulouse} \\
{\small \texttt{omar.naim.docs@gmail.com}}
\And
Krish Sharma\footnotemark[1] \\
IRIT \\
{\small Université de Toulouse} \\
{\small \texttt{krish.sharma@irit.fr}}
\And
Niyar R Barman \\
IRIT \\
{\small Université de Toulouse} \\
{\small \texttt{niyar.barman@irit.fr}}
\And
Nicholas Asher \\
IRIT \\
{\small CNRS} \\
{\small \texttt{nicholas.asher@irit.fr}}
}

\begin{document}
\maketitle
\begin{abstract}
Large Language Models (LLMs) typically come with a fixed architecture, despite growing evidence that not all layers contribute equally to every downstream task. We introduce TALE (Task-Aware Layer Elimination), an inference-time method that improves task performance by selectively removing layers that are irrelevant or detrimental for a given task. TALE optimizes task-specific performance, yielding a task-optimized architecture without retraining. 
Across 9 tasks and 5 model families, under both zero-shot and few-shot settings, TALE consistently matches or surpasses baseline performance while simultaneously reducing computational costs.
TALE also synergizes with fine-tuning, leading to further performance improvements. Computing TALE for a new task requires modest resources, making it a practical and deployable solution for task-specialized LLM inference.

\end{abstract}

\section{Introduction}

Substantial computational costs of Large Language Models (LLMs) can limit their use in resource-constrained settings and high-throughput applications. This has motivated a search for approaches that improve task performance while reducing computational cost. The growing use of multi-agent systems, where LLMs are specialized for particular roles, further amplifies this need. However, developing methods that address it remains challenging. Fine-tuning can improve task performance but does not reduce inference cost and requires substantial training data and compute. General pruning methods reduce computational cost but typically require retraining and often degrade performance on downstream tasks.

To address this need, we introduce {\sc Tale} (Task-Aware Layer Elimination), a simple, greedy, and iterative method that removes layers based on their impact on task-specific validation accuracy. At each step, {\sc Tale} evaluates all possible single-layer removals and selects the one that maximizes validation performance, repeating this process until performance falls below a user-defined threshold. The method is hardware-agnostic, requires no retraining, and operates entirely at inference time. Unlike existing pruning approaches, {\sc Tale} directly optimizes task-specific accuracy at each pruning step, enabling it to improve performance over the original model while remaining complementary to fine-tuning.

Our study of state-of-the-art, mid-sized transformers extends prior work showing that not all layers contribute equally to downstream tasks \cite{devries:etal:2020,dalvi:etalL:2020,sajjad:etal:2023}. We find that layer importance is highly task-dependent, and that removing task-misaligned layers can improve both efficiency and accuracy.

We showcase {\sc Tale} on five modern LLMs (LLaMA 3.1 8B, Qwen 2.5 7B, Qwen 2.5 0.5B, Mistral 7B, and Lucie 7B) across nine diverse benchmarks. We compare against prior training-free pruning methods on LLaMA 2 (7B and 13B), showing that {\sc Tale} achieves higher accuracy with lower computational cost. These improvements arise without retraining, architectural modification, or changes to the underlying model weights; they persist under alternative data splits, random seeds, and evaluation protocols.  Overall, {\sc Tale} provides a practical approach for building task-specialized LLMs: it can be applied \emph{post hoc} to existing checkpoints and complements fine-tuning.

\begin{figure*}[t]
\centering
\includegraphics[width=0.8\linewidth]{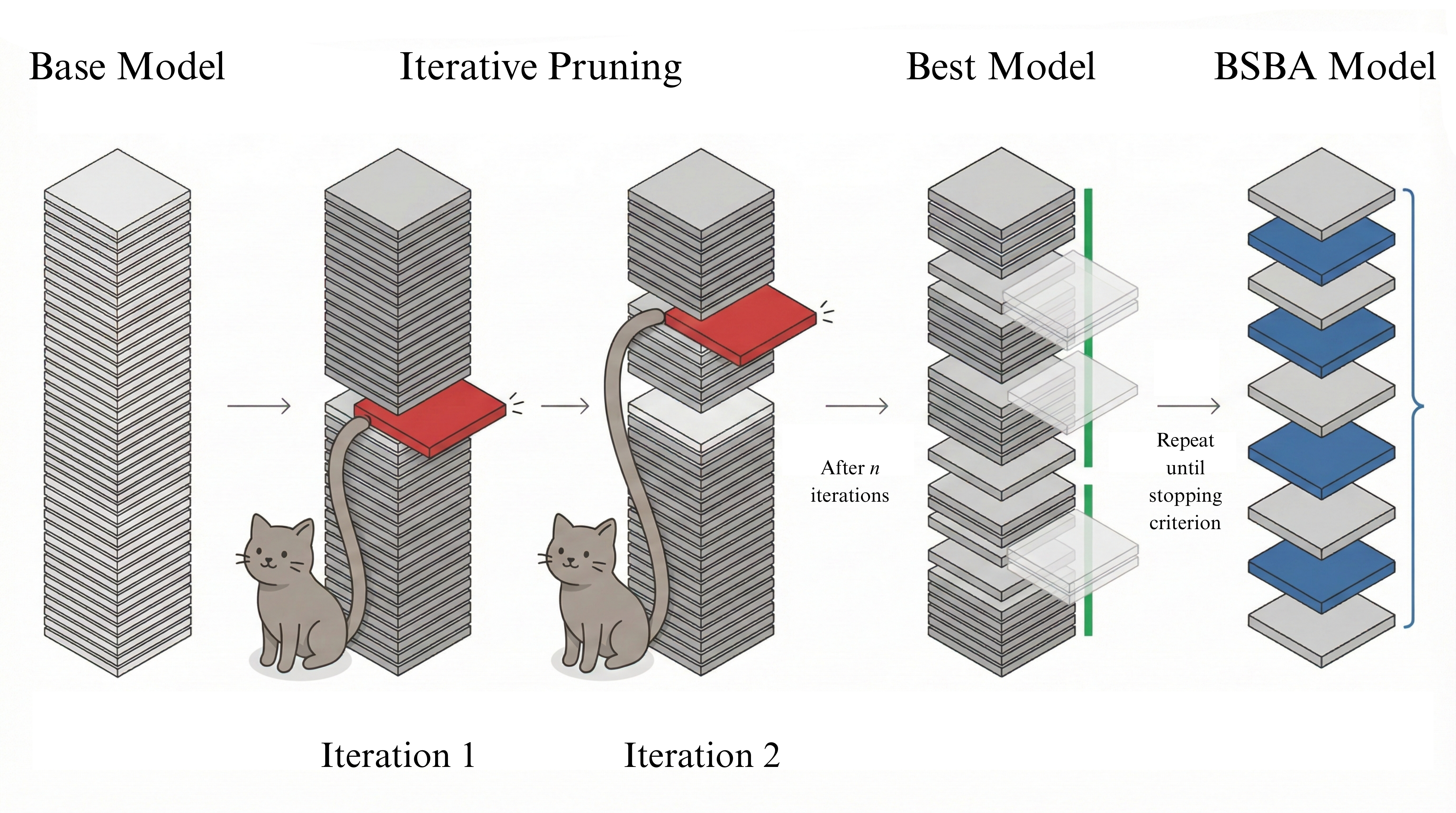}
\caption{
How TALE works to produce BEST and Best Speedup above Baseline (BSBA) models.
}
\label{fig:tale-demo}
\end{figure*}

\section{Related work}

While a large body of work focuses on improving the efficiency of LLMs through pruning, sparsity, and model compression, most approaches prioritize reducing computational cost, often at the expense of downstream performance.  Moreover, relatively little work explores using pruning to improve task-specific accuracy.

Structured pruning methods, which remove entire components such as neurons, attention heads, or layers \cite{structured2,structured3,structured4}, are the most closely related methods to {\sc Tale}. Recent work has explored both training-free and analysis-driven approaches to understanding and reducing redundancy in pretrained transformers. For instance,
\citet{zhang:etal:2024} propose training-free fine-tuning of sparse LLMs, while \citet{dalvi:etalL:2020} analyze redundancy using layer-level similarity and task-specific probing. \textbf{SLEB} \cite{sleb} removes layers based on cosine similarity of their representations and validates pruning decisions using perplexity on a linguistic dataset. Similarly, BlockPruner \cite{zhong2025blockprunerfinegrainedpruninglarge} segments Transformer layers into attention and MLP blocks and prunes them using perplexity-based heuristics. Because these methods rely on general-purpose metrics such as perplexity or representational similarity, they are largely task-agnostic.

Other approaches operate at different levels of granularity. \textbf{SliceGPT} \cite{slicegpt} reduces layer dimensionality via principal component analysis. Early-exit methods \cite{huang2026raeerobustretrievalaugmentedearly, wolczyk2021zerotimewasterecycling} dynamically determine stopping layers at inference time, typically using auxiliary predictors or retrieval mechanisms. At a finer granularity, \textbf{SparseGPT} \cite{sparsegpt} and \textbf{Wanda} \cite{wanda} perform unstructured pruning based on local reconstruction criteria or magnitude-activation products, achieving high sparsity but often degrading downstream performance. Methods such as OWL \citep{yin:etal:2024(owl)}, FLAP \citep{an:etal:2024(flap)}, and DarwinLM \citep{tang:etal:2025(darwinlm)} explore non-uniform sparsity or evolutionary search, primarily targeting efficiency gains.

\citet{peer:etal:2022} propose greedy layer pruning (GLP), which iteratively removes layers to meet a target accuracy. While superficially similar, GLP differs from {\sc Tale} in two key ways: it requires retraining on the downstream task and removes a predefined number of layers. In contrast, \method requires no retraining and adaptively determines the number of layers to remove based on validation performance.

Overall, \method differs from prior training-free approaches in both granularity and optimization objective. It operates at the layer level and directly optimizes task-specific validation accuracy, whereas existing methods typically rely on general-purpose criteria such as perplexity or reconstruction error. As shown in Section~\ref{overallperf}, this leads to consistent improvements in downstream accuracy while reducing computational cost. These results also suggest that commonly used proxies, such as representational similarity or perplexity, may be suboptimal for identifying task-relevant layers.

\section{Basics and Intuitions}

A transformer maps a sequence of input vectors $X^{(0)} = (x_1, \ldots, x_n)$ to a sequence
of hidden representations through a stack of $L$ layers. Each layer $\ell$ takes $X^{(\ell-1)}$
as input and produces
\[
    X^{(\ell)} = (x_1^{(\ell)}, \ldots, x_n^{(\ell)}),
\]
by applying self-attention and feed-forward blocks with residual connections. Concretely, each
layer updates the residual stream as
\[
    X^{(\ell)} = X^{(\ell-1)} + F_{\ell}(X^{(\ell-1)}), \quad \ell = 1, \ldots, L,
\]
where $F_{\ell}$ denotes the transformation implemented by layer $\ell$.
From this perspective, removing a layer $\ell$ corresponds to setting $F_\ell \equiv 0$,
omitting its residual contribution and reducing the update to the identity map:
\[
    X^{(\ell)} = X^{(\ell-1)}.
\]
This provides a natural interpretation of layer-wise pruning as deleting components of the
residual update while preserving the overall stream structure.

\paragraph{Intermediate representations.}
To better understand the role of individual layers, we examine intermediate hidden states.
Instead of using only the final representation $X^{(L)}$, we project intermediate
representations $X^{(k)}$ for $k < L$ directly into the vocabulary space using the
output projection matrix $W_{\text{out}}$:
\[
    \hat{y}^{(k)} = \mathrm{softmax}\!\left(W_{\text{out}}\, x_n^{(k)}\right).
\]

\hidden{
A transformer maps a sequence of input vectors $(x_1, \cdots, x_n)$ to a corresponding sequence of output vectors through a stack of $L$ layers. Each layer $\ell$ transforms the hidden representations $X^{(\ell)} = (x_1^{(\ell)}, \ldots, x_n^{(\ell)})$ into $X^{(\ell+1)}$ via attention and feed-forward blocks, connected by residual pathways. Removing layer $\ell$ redirects the computation such that $X^{(\ell-1)} \rightarrow X^{(\ell+1)}$, making the architecture naturally amenable to layer-wise pruning.

\paragraph{Intermediate representations.}
As a first pass to understanding the role of individual layers, we examine partial forward passes. Let $h^{(k)}$ denote the hidden representation after layer $k$. Instead of decoding only from the final representation $h^{(L)}$, we project intermediate representations $h^{(k)}$ for $k < L$ directly into the vocabulary space using the output projection:
\[
\hat{y}^{(k)} = \mathrm{softmax}(W_{\text{out}} h^{(k)}).
\]

}

While evaluating $\hat{y}^{(k)}$ across different values of $k$, we were surprised to find that for many tasks, intermediate layers ($k < L$) can achieve higher accuracy than the final layer (Figure~\ref{graph:oldexp}). This suggests that additional layers do not always improve task-specific performance: some layers contribute marginally, while others may introduce representational noise. Not all layers in an LLM are equally useful, and selectively removing redundant layers can preserve or even improve downstream accuracy.

\medskip
\noindent We formalize this idea in the next section as a greedy, task-aware layer pruning procedure.

\hidden{

Our intuition for \method is based on examining partial forward passes. Let $h^{(k)}$ denote the hidden representation after layer $k$. Instead of decoding only from the final representation $h^{(L)}$, we project intermediate representations $h^{(k)}$ for $k < L$ directly into the vocabulary space using the output projection $W_{\text{out}}$:
\[
\hat{y}^{(k)} = \mathrm{softmax}(W_{\text{out}} h^{(k)}).
\]
While evaluating $\hat{y}^{(k)}$ across different values of $k$, we sere surprised to find that for many tasks, intermediate layers ($k < L$) can achieve higher accuracy than the final layer (Figure~\ref{graph:oldexp}). This suggests that additional layers do not always improve task-specific performance: some layers contribute marginally, while others may introduce representational noise. Not all layers in an LLM are equally useful, and selectively removing redundant layers can preserve or even improve downstream accuracy. \method (Task-Aware Layer Elimination) formalizes this intuition as a principled, iterative optimization procedure.
}

\begin{figure*}[!ht]
    \centering
        \includegraphics[width=\linewidth]{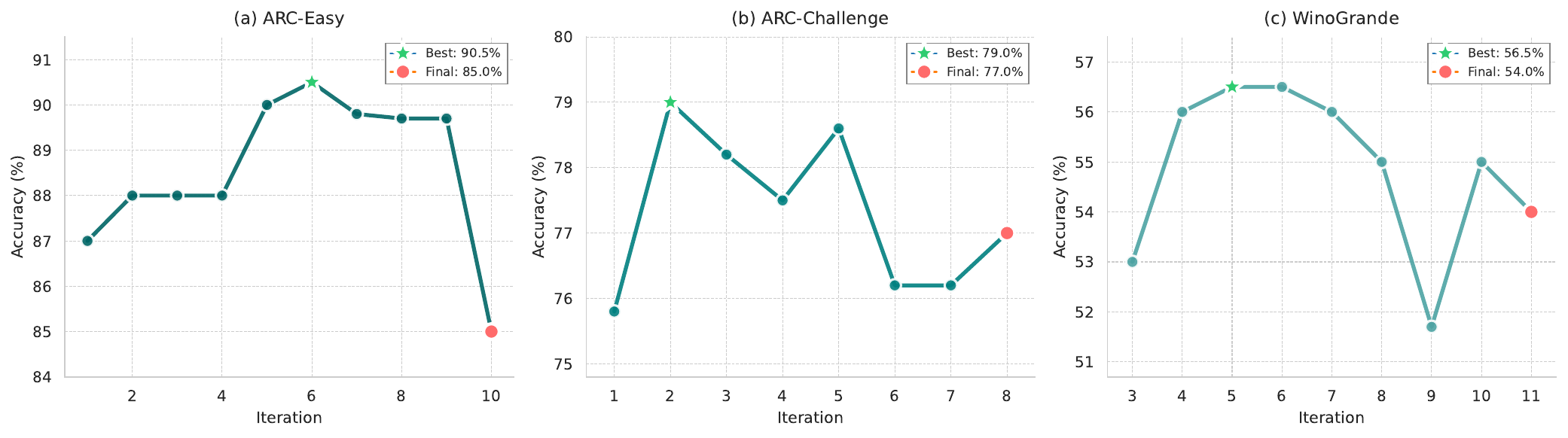}

\hidden{
    \vspace{0.3cm}
    \begin{subfigure}[t]{0.32\textwidth}
        \centering
        \includegraphics[width=\linewidth]{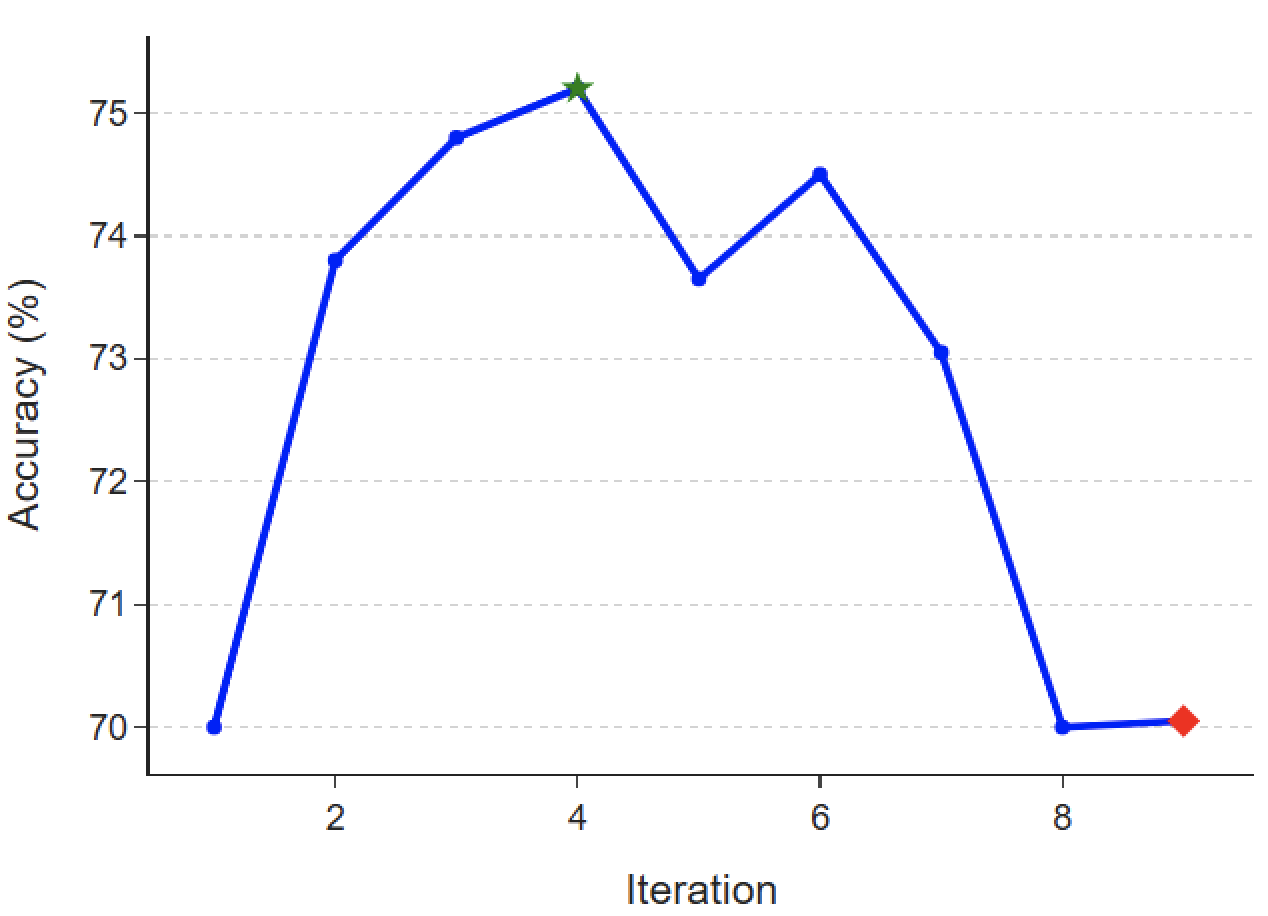}
        \caption{CommonQA}
    \end{subfigure}
    \hfill
    \begin{subfigure}[t]{0.32\textwidth}
        \centering
        \includegraphics[width=\linewidth]{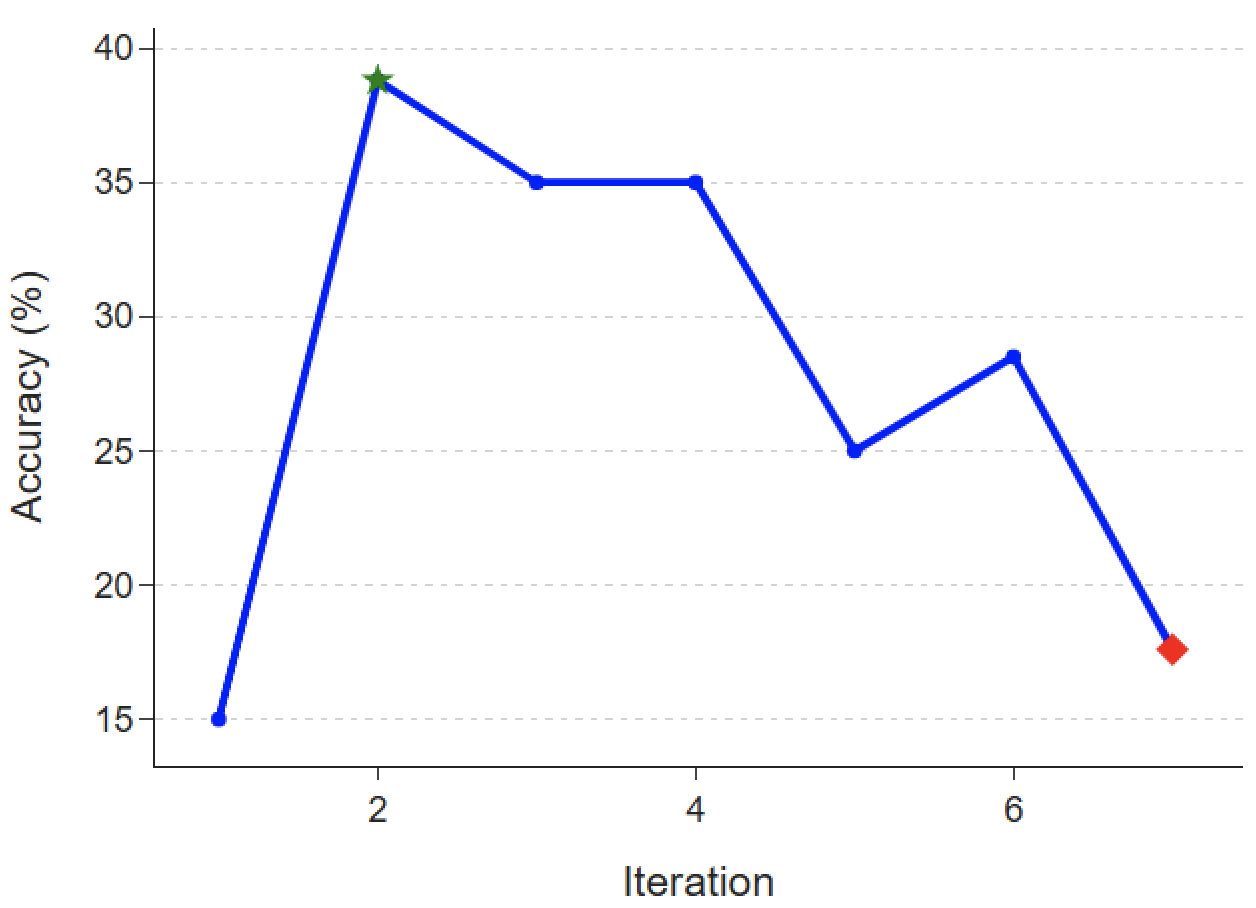}
        \caption{GSM8K-Hard}
    \end{subfigure}
    \hfill
    \begin{subfigure}[t]{0.32\textwidth}
        \centering
        \includegraphics[width=\linewidth]{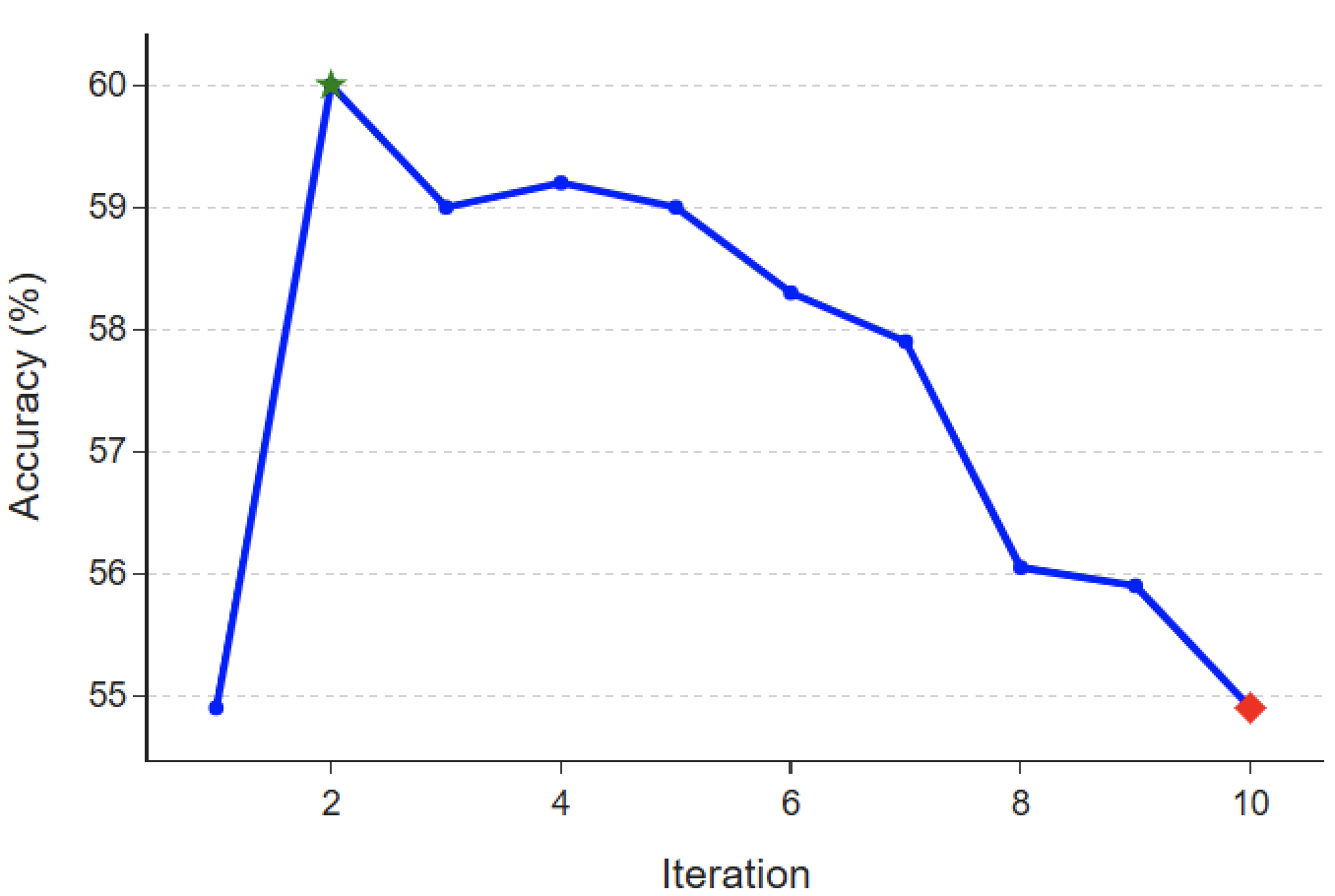}
        \caption{MMLU}
    \end{subfigure}
}
    \caption{Accuracy progression of \method across three benchmark datasets for LLaMA 3.1 8B. Each curve represents the accuracy at successive iterations. The \textcolor{green}{$\star$} denotes the best-performing layer drop configuration, while the \textcolor{red}{$\bullet$} highlights the best speedup with at least baseline accuracy (BSBA) configuration. Plots for all tasks are provided in Appendix \ref{appendix:dynamicss}.}
    \label{graph:dataset_results}
\end{figure*}

\begin{figure*}[t]
    \centering
    \includegraphics[width=\linewidth]{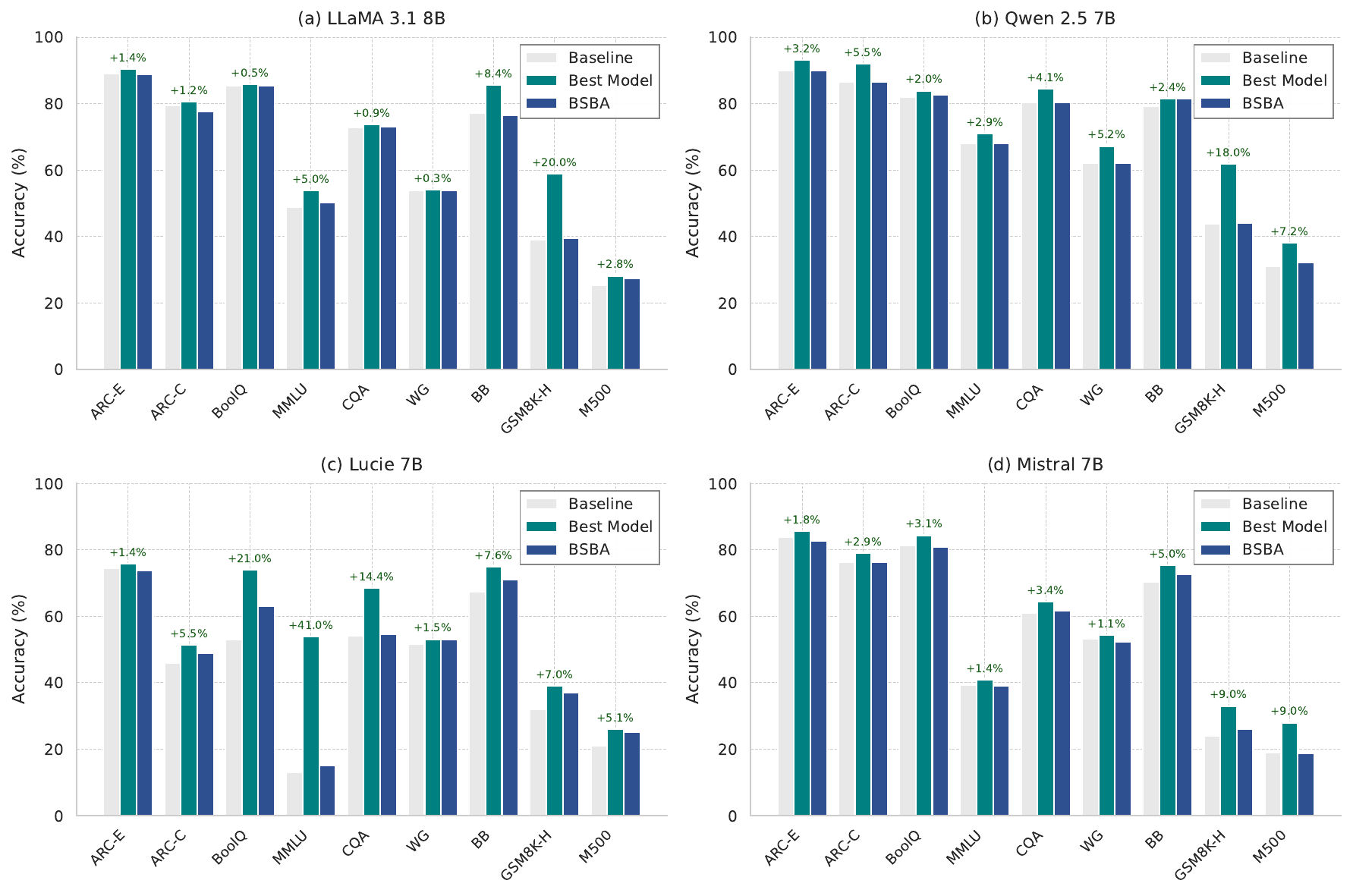}
    \caption{\textbf{Zero-shot performance comparison across four language models with layer dropping} 
    Light blue bars show baseline performance, while dark blue bars represent best model performance after strategic layer dropping. Percentages on top of the columns represent the absolute accuracy gained with best models over base models. 
    Results demonstrate that dropping 1 to 10 layers can improve or maintain performance across most benchmarks, with notable gains on mathematical reasoning tasks (e.g., LLaMA 3.1 8B: 39.0\% → 59.0\% on GSM8K-HARD with only 1 layer dropped). 
    This suggests significant redundancy in deeper language model architectures.}
    \label{fig:model_comparison}
\end{figure*}

\section{\method}
\method is a greedy layer-pruning algorithm for compressing pre-trained open-weight LLMs.  It systematically removes layers while preserving, and sometimes improving, task performance (Algorithm~\ref{algo}). Starting from the full pre-trained model, \method evaluates all possible single-layer removals at each step and computes the validation accuracy of each candidate architecture. The layer whose removal yields the highest validation accuracy is permanently eliminated, and the resulting compressed model becomes the starting point for the next step. This process continues until performance falls below a predefined threshold.

For this paper, we set the stopping threshold to 8\% below the baseline task accuracy, allowing the search to explore slightly lower-performing intermediate architectures in case later pruning steps yield larger gains. In practice, however, we did not observe any cases in which performance recovered after falling below the baseline. Once the threshold is reached, the algorithm returns the most compressed model whose performance remains above the threshold.

\begin{algorithm}[!ht]
\caption{\method: Iterative Layer Pruning}
\label{algo}
\begin{algorithmic}[1]
\Require Pre-trained model $\mathcal{M}$ with $L$ layers; validation set $\mathcal{D}_{val}$; performance threshold $\epsilon$
\Ensure Compressed model $\mathcal{M}^*$
\State Initialize $\mathcal{M}^* \gets \mathcal{M}$
\Repeat
    \For{each layer $\ell \in \{1,\dots,L\}$ of $\mathcal{M}^*$}
        \State Construct candidate model $\mathcal{M}_{-\ell}$ by removing layer $\ell$
        \State Compute the validation accuracy $A_\ell = \text{Acc}(\mathcal{M}_{-\ell}, \mathcal{D}_{val})$
    \EndFor
    \State Select $\ell^* = \arg\max_\ell A_\ell$
    \If{$A_{\ell^*} \geq \text{Acc}(\mathcal{M}^*, \mathcal{D}_{val}) - \epsilon$}
        \State Update $\mathcal{M}^* \gets \mathcal{M}_{-\ell^*}$
    \Else
        \State \textbf{break}
    \EndIf
\Until{all accuracies are below the threshold}
\State \Return $\mathcal{M}^*$
\end{algorithmic}
\end{algorithm}

\subsection{Benchmarks} \label{sec:tale}

We evaluate \method across a diverse suite of nine benchmarks spanning reasoning, language understanding, and commonsense reasoning tasks. For mathematical reasoning, we use \textbf{GSM8K-Hard}, a curated subset of GSM8K \cite{gsm8k} with more than five premises per question to increase difficulty, and \textbf{MATH500} \cite{math500}. 

For language understanding, common sense reasoning, and multi-task generalization, we use \textbf{MMLU} \cite{mmlu} and \textbf{BoolQ} \cite{boolq}; \textbf{Winogrande} \cite{winogrande}, \textbf{CommonsenseQA} \cite{commonsenseqa}, and \textbf{BIG-Bench} \cite{bigbench}, as well as \textbf{ARC-Easy} and \textbf{ARC-Challenge} \cite{ARC}.


\subsection{Evaluation protocol}
\method is in effect an {\bf end-to-end optimizer}; it optimizes a model for a particular task and a particular evaluation. So it is important to evaluate \method with respect to different evaluations procedures.

We consider two complementary evaluation methods. First, we use the standard LM-Eval framework, which selects the answer with the highest probability among the provided options for multiple-choice tasks. While this approach is widely adopted, it does not always capture the quality of the model’s generated outputs.

To complement this, we introduce an additional evaluation procedure, \textbf{Decoder Eval}, which directly evaluates the model’s generated responses. In this setting, the model is prompted to produce its answer in a structured format. We then extract the final answer from the generated output and compare it to the ground truth to compute accuracy.

\hidden{
We use two different evaluation methods.  We use the standard LM-Eval evaluation package, but we complement it with a different evaluation that directly examines model generated output and hence its decoder-layer representation interaction.  LM-Eval selects the answer with the highest probability from options provided in a data set with multiple choice answers but does not evaluate actual, generated output in those cases.  For data sets where generation is actually needed, it goes through the output of the model; if it finds the correct answer before the end token, it marks the answer as correct--if not it is incorrect. In response, we developed an automatic evaluation, {\bf Decoder Eval}, that tests the real generated output of the models. It uses a prompt to get the model to predict its answer in a particular format after reasoning steps, which indicates better whether the model is "understanding the problem."  In evaluation we look for the format and then calculates the accuracy from ground truth.  All base models could obey this prompt. 
}

\subsection{The dynamics of \method's iterations}
    
To better understand how \method behaves during optimization, we analyze how task performance evolves throughout the iterative pruning process.

Figure~\ref{graph:dataset_results} illustrates representative optimization trajectories for LLaMA~3.1 8B on three benchmark tasks. Each curve shows the change in accuracy as layers are progressively removed. Across tasks, we observe a consistent pattern: the first pruning step often yields a noticeable improvement in accuracy, followed by smaller gains or mild fluctuations. After this initial phase, performance typically decreases monotonically as additional layers are removed, eventually falling below the baseline. The full set of trajectories is shown in Figure~\ref{graph:dataset_results1}. These curves suggest structured, task-dependent redundancy patterns that merit further investigation.

From these dynamics, we identify three key properties of \method:

(i) \textbf{Existence of high-performing compressed models.} \method often identifies configurations that outperform the original model on a given task. We refer to these as \textbf{Best (BEST)} models. In addition, \textbf{Best Speedup with at least Baseline Accuracy (BSBA)} denotes the most compressed configurations that remain within the performance threshold defined by the baseline.

(ii) \textbf{Multi-step gains before degradation.} Performance improvements are not limited to the first pruning step; in many cases, accuracy improvements persist across multiple iterations before diminishing returns appear, indicating substantial redundancy in these pre-trained models.

(iii) \textbf{Task-dependent pruning dynamics.} The behavior of \method varies across tasks. For example, ARC-Easy and MMLU tolerate deeper pruning while maintaining or improving performance, whereas reasoning-intensive tasks such as GSM8K-Hard reach peak performance earlier and degrade more quickly, reflecting differences in layer importance across domains.

\hidden{
To better understand how \method works, we analyze the evolution of task performance during the pruning process.

    Figure~\ref{graph:dataset_results} visualizes three iterative optimization trajectories for LLaMA~3.1 8B performed with {\sc Tale}. Each curve tracks accuracy as layers are progressively removed. The results reveal a consistent pattern across most tasks: the first pruning step typically yields a substantial gain in accuracy, followed by smaller increases or decreases. Beyond this point, performance tends to follow a monotonically decreasing trend, eventually degrading well below the baseline.

    These trajectories provide three key elements of {\sc Tale}:  
    (i) \method identifies compressed {\bf best-performing (BEST)} models that \emph{outperform} the original across diverse tasks, indicated by green star markers lying above the baseline. {\bf Best Speedup with at least Baseline Accuracy (BSBA)} denotes the most compressed models that remain above TALE’s stopping threshold (red bullet markers).  
    (ii) Accuracy improvements often persist across multiple pruning steps before diminishing returns appear, indicating substantial redundancy even in carefully tuned pretrained models.  
    (iii) Pruning dynamics are task-specific: datasets such as ARC-Easy and MMLU tolerate deeper pruning while continuing to improve, whereas reasoning-heavy tasks such as GSM8K-Hard converge earlier, reflecting heterogeneous layer importance across domains.
 }   
\section{Results} \label{sec:results}

We evaluate \method along five dimensions: (i) overall performance across models and tasks, (ii) robustness across random seeds and evaluation protocols, (iii) comparison with state-of-the-art pruning methods, (iv) computational efficiency, and (v) interaction with fine-tuning and few-shot learning.


\subsection{Overall Performance}
\label{overallperf}

We first evaluate \method in a zero-shot setting on four medium-scale models (LLaMA~3.1 8B, Mistral~7B, Lucie~7B, Qwen~2.5 7B) and one smaller model (Qwen~2.5 0.5B). Figure~\ref{fig:model_comparison} and Table~\ref{tab:combined_model_comparison_placeholder} compare baseline models with their pruned counterparts. 

Across all models and benchmarks, we observe consistent improvements in accuracy after applying \method, although the magnitude of gains varies across tasks and architectures.\footnote{Code available at \url{https://github.com/omyokun/tale/}} On ARC-Challenge, improvements are modest for LLaMA (+1.6\%) but more pronounced for Qwen~2.5 7B (+6.3\%). Reasoning-intensive tasks benefit the most, with gains ranging from 23\% to 51\% on MATH500 and GSM8K across models. Results obtained under LM-Eval and Decoder Eval exhibit the same behavior (for details see Table~\ref{tab:robustness_study} in Appendix~\ref{appendix:robustness} and Table~\ref{tab:lmeval} in Appendix~\ref{appendix:lmeval}), indicating that improvements are not specific to a single evaluation protocol.

Beyond accuracy, \method also reduces computational cost by removing unnecessary layers, highlighting substantial redundancy in modern language models. Notably, selectively removing task-misaligned layers can \emph{improve} performance, rather than degrade it as commonly expected in pruning.

Taken together, these findings establish the central empirical result of this work: \textbf{task-aware layer pruning can simultaneously improve downstream performance and reduce model depth}.



\hidden{

\subsection{Overall Performance}

We begin by evaluating \method across the benchmarks described above. Figure~\ref{fig:model_comparison} and Table~\ref{tab:combined_model_comparison_placeholder} summarize the results.

Across all models and tasks, \method consistently produces pruned models that \textbf{match or exceed the baseline accuracy of the original model without retraining}. These improvements are observed across reasoning tasks such as GSM8K-Hard and MATH500, as well as language understanding and commonsense reasoning benchmarks.

{\color{olive}
We next evaluated \method on our benchmarks in 0-shot settings across four medium-scale models (LLaMA~3.1 8B, Mistral~7B, Lucie~7B, Qwen~2.5 7B) and one smaller model (Qwen~2.5 0.5B).

Figure \ref{fig:model_comparison} compares accuracy of four mid-sized baseline models against pruned counterparts selected using validation splits drawn from the training data; all reported results are evaluated on a disjoint test (or held-out validation) set. Across all benchmarks and base models, Best models yield consistent accuracy gains though percentage gains vary across models and tasks.\footnote{Code available at \url{https://anonymous.4open.science/r/tale/}}  On Arc  Challenge, Llama Best has the lowest increase in accuracy (though still significant) at 1.6\%, while Qwen 7b gains 6.3\%.  Reasoning tasks (Math500, GSM8K) benefited most from 
\method with gains ranging from 23\% to 51\% across all models.
Similar results on the larger models using LM-Eval are in the appendix. The fact that we see the same behavior under two different evaluations is evidence that \method is capturing a real phenomenon, not an artifact of evaluation. 
}

In addition to improving accuracy, \method reduces computational cost by removing unnecessary layers, demonstrating that substantial redundancy exists in modern language models. Notably, we observe that carefully removing task-misaligned layers can lead to \emph{improved} performance, rather than the degradation typically associated with pruning.

These results establish the central empirical finding of this work: \textbf{task-aware layer pruning can improve downstream performance while simultaneously reducing model depth}.

}

\paragraph{Remark on Validation Set Size}  
In our study, \method requires only modestly-sized sets for task-specific optimization, ranging from 500 to 1500 examples.  As seen in Table \ref{table_ablation} (Appendix \ref{appendix:ablation}), once the validation set size exceeds 500 examples, the set of layers dropped stabilizes across all tasks.

\subsection{Robustness}

\begin{figure}[!ht]
    \centering
    \includegraphics[width=0.9\columnwidth]{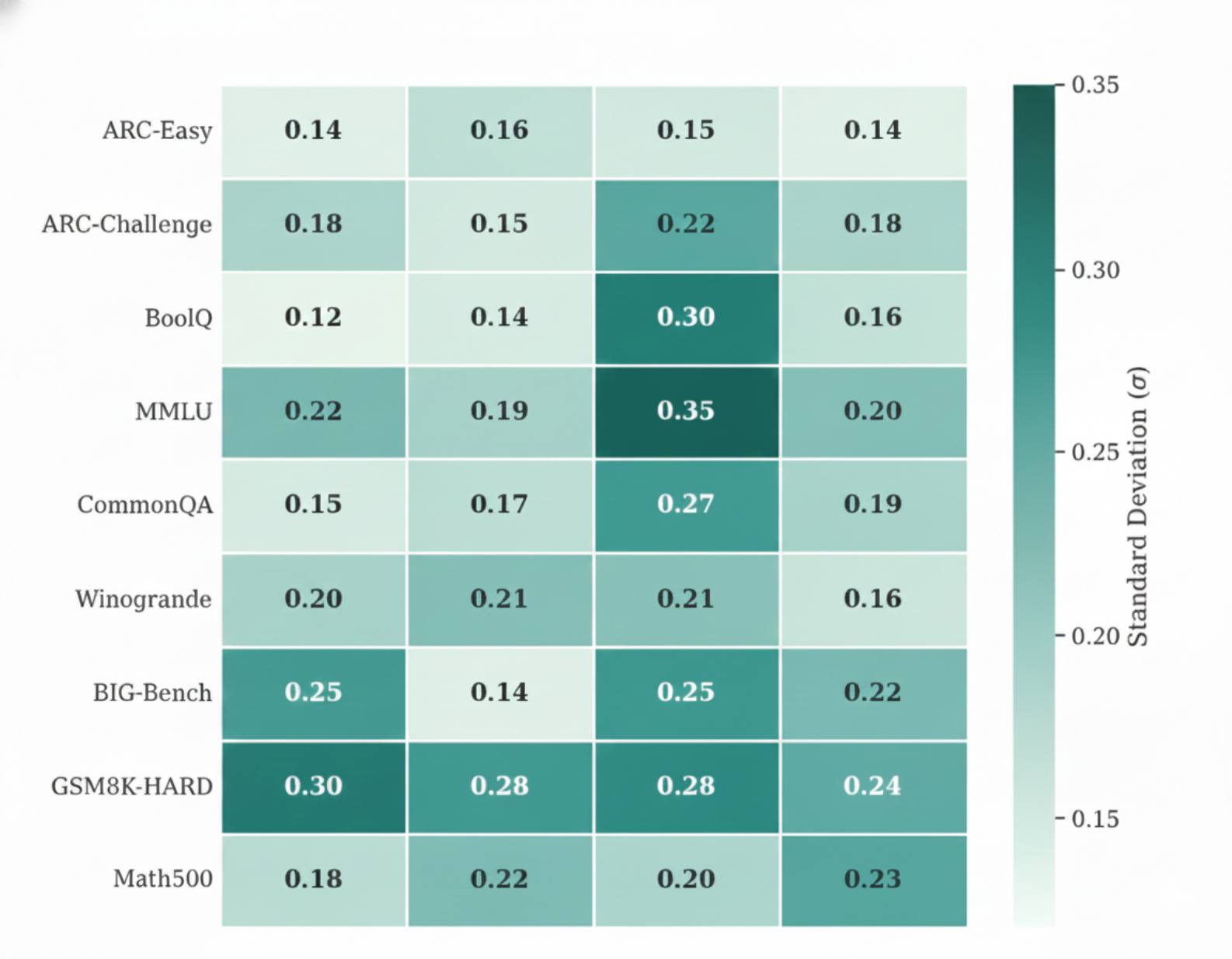}
    \caption{
        \textbf{        
        Model Stability Across Datasets.} 
        Heatmap visualizing the standard deviation ($\sigma$) of the final accuracy scores for the Best Model variant of Llama, Qwen, Lucie and Mistral, four different LLMs in the 7-8b range across nine datasets. 
        The results are aggregated from five independent runs (seeds). 
        Lighter green indicates higher stability (lower $\sigma$); darker green, lower stability (higher $\sigma$). 
        LLaMA 3.1 8B shows the overall lowest variance.
    }
    \label{fig:stability_heatmap}
\end{figure}

We evaluate the robustness of \method to random initialization by varying the pruning seed. Figure~\ref{fig:stability_heatmap} shows that performance variations across seeds are minimal. Table~\ref{tab:robustness_study} (Appendix \ref{appendix:robustness}) reports the mean and variance over five seeds for the best configurations of LLaMA, Qwen, Lucie, and Mistral.

Across all models, we observe consistently low variance, indicating that \method is stable with respect to seed choice and does not rely on favorable initialization. These results demonstrate that the performance gains of \method are robust and reproducible.

\hidden{
We next evaluated \method on our benchmarks in 0-shot settings across four medium-scale models (LLaMA~3.1 8B, Mistral~7B, Lucie~7B, Qwen~2.5 7B) and one smaller model (Qwen~2.5 0.5B).

Figure \ref{fig:model_comparison} compares accuracy of four mid-sized baseline models against pruned counterparts selected using validation splits drawn from the training data; all reported results are evaluated on a disjoint test (or held-out validation) set. Across all benchmarks and base models, Best models yield consistent accuracy gains though percentage gains vary across models and tasks.\footnote{Code available at \url{https://anonymous.4open.science/r/tale/}}  On Arc  Challenge, Llama Best has the lowest increase in accuracy (though still significant) at 1.6\%, while Qwen 7b gains 6.3\%.  Reasoning tasks (Math500, GSM8K) benefited most from 
\method with gains ranging from 23\% to 51\% across all models.
Similar results on the larger models using LM-Eval are in the appendix. The fact that we see the same behavior under two different evaluations is evidence that \method is capturing a real phenomenon, not an artifact of evaluation. 
}




\hidden{
For each benchmark, we partition the available data into three disjoint subsets: (i) a training pool, (ii) a held-out \emph{optimization split} drawn from the training pool, and (iii) a final \emph{evaluation split} (the official test set when available, or a held-out validation set otherwise).

TALE uses \emph{only} the optimization split to select which layers to remove for a given task. The evaluation split is never used during layer selection. All results reported throughout the paper correspond to performance on this unseen evaluation split.

When official test sets are unavailable, we follow prior work and report results on held-out validation splits that are disjoint from the data used for layer optimization. To see results with two different training subsets see Tables \ref{tab:robustness_study} and \ref{tab:combined_model_comparison_placeholder}.
}



\subsection{Comparisons}\label{sec:results}

\begin{figure*}[!ht]
    \centering
    \includegraphics[width=\linewidth]{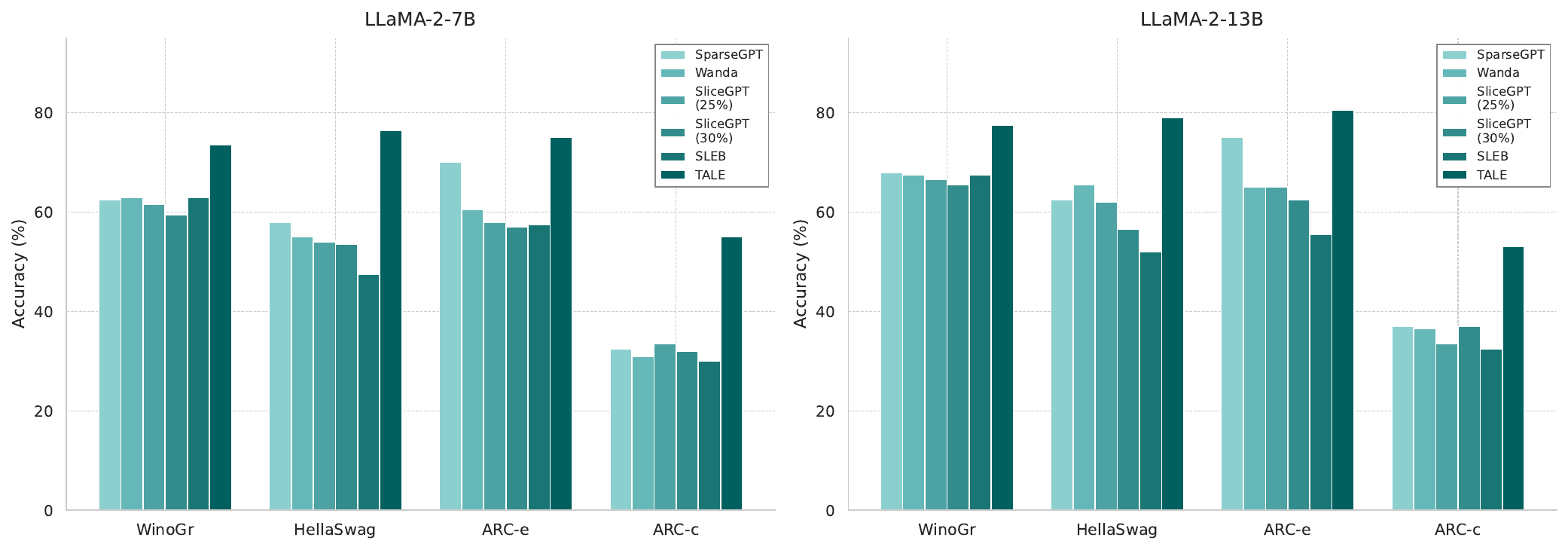}
    \caption{\textbf{Performance comparison of pruning methods on LLaMA-2 models.} 
    Evaluation of seven pruning approaches using LM-Eval accuracy across four zero-shot benchmarks for LLaMA-2-7B (left) and LLaMA-2-13B (right). 
    Methods are shown in a blue gradient from light (Baseline) to dark ({\sc Tale}). 
    \method achieves the highest accuracy across all tasks while maintaining equivalent sparsity to SLEB, which performed second best to {\sc Tale}.  Method outperformed all structured pruning methods (SpareGPT, Wanda, SliceGPT) on both model sizes, demonstrating effective layer dropping without accuracy loss.}
    \label{fig:method_comparison}
\end{figure*}

We compare \method{} against recent training-free pruning methods on LLaMA-2-7B and LLaMA-2-13B, including SLEB, SparseGPT, Wanda, and SliceGPT. Results are reported in Figure~\ref{fig:method_comparison}.

Across all benchmarks, \method consistently outperforms prior approaches at comparable sparsity levels and is the only method that reliably matches or exceeds the performance of the unpruned baseline across tasks. In contrast, existing pruning methods generally degrade downstream accuracy while reducing model depth.

A key distinction between \method{} and prior work is the optimization objective. While most existing methods rely on general-purpose signals such as perplexity or representation similarity to identify redundant layers, \method{} directly optimizes task-specific validation accuracy. This task-aware objective leads to more effective identification of layers that are detrimental for downstream performance.

 To show this, we used a representational similarity technique (cosine similarity or cossim) to guide {\sc Tale}.  \method registered drops similar to those seen with SLEB and other similarity driven optimization methods.  On ARC-Easy, for instance, cossim led TALE to drop 2 layers, {\bf dropping task accuracy} from Llama's baseline of {\bf 79.5} to a pruned model accuracy of {\bf 58.5} with a time speed up of 1.32.  Table \ref{tab:perplexity} in Appendix \ref{appendix:perplexity} shows that \method suffers similar drops in performance when perplexity is used as an optimizing objective. 

In addition, we evaluate task-aware variants of SLEB and BlockPruner, which are given access to the same validation data and evaluation metric as {\sc Tale}. As shown in Table~\ref{table:taskaware}, these methods still do not perform as well as \method{} across all tasks.\footnote{We also note that \method appears superior to early-exit methods we examined.  From the limited information we have about RAEE outside of classification tasks, RAEE gives a score on ARC easy of 65\%, while TALE substantially improves on the Llama baseline score of 76\%  and TALE on ARC Easy gives 79 \% using LM eval.}

\paragraph{Layer selection beyond fixed truncation.}
Our experiments indicate that for reasoning, intensive tasks (e.g., GSM8K, MATH500), removing early or intermediate layers can lead to performance gains, an effect that standard top-$k$ layer truncation strategies fail to capture. Additionally, a key strength of \method is that the number of pruned layers is not fixed \emph{a priori}. Instead, it adaptively determines when to stop pruning. Empirically, we observe that a specific number of removed layers $n$ may be optimal, while pruning further ($n+1$) can cause a sharp drop in performance. Moreover, this optimal point varies across datasets, making any fixed, pre-defined pruning budget potentially suboptimal or even harmful. This highlights a key limitation of approaches that fix the number of layers to prune from the outset.

\begin{table}[h]
\centering
\small
\resizebox{\linewidth}{!}{
\begin{tabular}{ll|ccc}
\hline
\textbf{Eval} & \textbf{Method } & \textbf{ARC-e} & \textbf{ARC-c} & \textbf{WinoGr} \\
\hline

dec & TALE    & 76.7 & 54.3 & 73.1 \\
    & SLEB-ta          & 61.0 & 38.0 & 66.5 \\
    & BlockPr-ta       & 64.6 & 39.6 & 65.59 \\
\hline

LM  & BlockPr-ta       & 65   & 41   & 66 \\
    & TALE             & 81   & 55   & 78 \\
\hline
\end{tabular}}
\caption{Comparison of \method to SLEB and BlockPruner using both decoder and LM eval evaluations}
\label{table:taskaware}
\end{table}

\hidden{
{\color {magenta} ***see appendix*** we have perplexity optimized models somewhere we can test them on a benchmark}
}

\subsection{Computational Cost and Amortized Efficiency}


\method requires a modest, one-time computation to identify the optimal layer-set for a given task, which is then amortized over the model’s entire inference lifetime. 
For an $L$-layer model and a validation set of size $V$, the time of pruning process is proportional to $O(I \cdot L \cdot V \cdot T_{\text{layer}})$, where $I$ is the number of pruning iterations. 
For LLaMA 3.1 8B across our benchmarks ($L=32$, $V \approx 500-1500$), the pruning required approximately 1 to 2 GPU-hours on a single A100. 

In Appendix \ref{appendix:computation}, we also report first-token latency and aggregate throughput for a subset of model/task pairs, measured on one A100-80GB NVIDIA GPU (with identical decoding settings for base and pruned models). In these experiments, the reported pruned model is the BEST variant, not the BSBA variant. TALE improves first-token latency in 9/9 settings (macro average: -14.3\%) and throughput in 9/9 settings (macro average: +17.9\%).

{\bf Takeaways.}  
\method consistently uncovers high accuracy and high accuracy/high efficiency models. By balancing task fidelity with computational savings, it enables both accuracy-focused and efficiency-focused deployment. Even strong, larger models like Qwen 7B see significant improvements, but so do small models (Qwen 0.5B). 



\subsection{Interactions between {\sc Tale}, few-shot learning and fine-tuning}

{\bf Few-shot setting.}  
As few-shot prompting improves baselines on many tasks,\footnote{ in particular reasoning tasks like GSM8K and Math500} we tested on Lucie and LLaMA models whether \method could synergize with few-shot prompting to bring higher gains (Tables~\ref{tab:lucie_fewshots} and \ref{tab:llamafewshots_results_colors}). {\sc Tale}-pruned variants still achieve higher accuracy in nearly all settings. This shows that {\sc Tale}-induced improvements are  complementary to gains from in-context learning. 

\hidden{We provide the results for Llama below
\begin{table*}[!ht]
\centering
\renewcommand{\arraystretch}{1.3}
\resizebox{0.8\linewidth}{!}{
\begin{tabular}{l|c|ccc|ccc}
\toprule
\multirow{3}{*}{\textbf{Dataset}} & 
\multicolumn{7}{c}{\textbf{LLaMA 3.1 8B few-shots}} \\
\cmidrule(lr){2-8}
 & \multicolumn{1}{c|}{Baseline} & \multicolumn{3}{c|}{Best Model} & \multicolumn{3}{c}{BSBA} \\
\cmidrule(lr){2-2} \cmidrule(lr){3-5} \cmidrule(lr){6-8}
& Perf. & Perf. & \#D & Sp. & Perf. & \#D & Sp. \\
\midrule
ARC-Easy      & 90.36  & \textbf{92.18}{\color {green} 2.01\% $\uparrow$} & 4 & 1.14 & 90.91 & 8 & 1.37 \\
ARC-Challenge & 78.2 & \textbf{83.10} {\color {green} 6.27\% $\uparrow$} & 3 & 1.17 & 78.62  & 9   & 1.42 \\
BoolQ         &  82.7 & 85.3 {\color {green} 3.1\% $\uparrow$} & 4 & 1.11 & 83.0 &  6 & 1.22 \\
MMLU          &  59.2 & 62.38{\color {green} 5.37\% $\uparrow$}  & 4 & 1.14 & 59.57 & 7 & 1.26 \\
COMMONQA      &  73.30 & 75.30{\color {green} 2.72\% $\uparrow$} & 6 & 1.22 & 73.80 & 7 & 1.32 \\
WINOGRANDE    & 57.01 & 60.1{\color {green} 5,26\% $\uparrow$} & 3 & 1.1 & 57.02 & 8 & 1.3 \\
BIG-Bench     & 70.0 & 83.60{\color {green} 19,43\% $\uparrow$} & 5 & 1.2 & 81.20 & 15 & 1.83 \\
GSM8K-HARD    & 60.67 & 60.67 & 0 & 1 & 60.67 & 0 & 1 \\
MATH500        & 44.00 & 49.00{\color {green} 11.36\% $\uparrow$} & 1 & 1.02 & 45.00 & 2 & 1.03 \\
\bottomrule
\end{tabular}
}
\caption{Results of \textbf{LLaMA 3.1 8B} across nine benchmarks. All tested on 5-shots, except gms8k and MATH500 on 8-shots 
}
\label{tab:llamafewshots_results_colors}
\end{table*}
}

\newcommand{\perc}[1]{\textcolor{green!50!black}{\small(+#1\% $\uparrow$)}}
\begin{table*}[!ht]
\centering
\renewcommand{\arraystretch}{1.3}
\resizebox{\linewidth}{!}{
\begin{tabular}{ll|cc|cc|cc|cc|cc|cc}
\toprule
\multirow{3}{*}{\textbf{Model}} & \multirow{3}{*}{\textbf{Dataset}} 
& \multicolumn{2}{c|}{Baseline} 
& \multicolumn{2}{c|}{\method} 
& \multicolumn{2}{c|}{FT Only} 
& \multicolumn{2}{c|}{\method $\rightarrow$ FT} 
& \multicolumn{2}{c|}{FT $\rightarrow$ \method} 
& \multicolumn{2}{c}{(\method $\rightarrow$ FT) $\rightarrow$ \method} \\
\cmidrule(lr){3-4} \cmidrule(lr){5-6} \cmidrule(lr){7-8} \cmidrule(lr){9-10} \cmidrule(lr){11-12} \cmidrule(lr){13-14}
& & Perf. & \#D & Perf. & \#D & Perf. & \#D & Perf. & \#D & Perf. & \#D & Perf. & \#D \\
\midrule
\multirow{3}{*}{Llama 3.1 8B} 
& Winogrande & 53.83 & 0 & 56.67 & 4 & 85.00 & 0 & 87.06 & 4 & 86.74 & 7 & 87.37 & 8 \\
& MMLU & 54.87 & 0 & 59.90 & 1 & 63.62 & 0 & 63.49 & 1 & 64.21 & 2 & 64.01 & 2 \\
& CommonQA & 72.20 & 0 & 75.30 & 3 & 81.88 & 0 & 81.80 & 3 & 83.40 & 3 & 82.90 & 6 \\
& GSM8K & 15.07 & 0 & 37.08 & 3 & 42.70 & 0 & 53.96 & 1 & 50.86 & 2 & 54.02 & 2 \\
\midrule
\multirow{2}{*}{Qwen 0.5B} 
& Winogrande & 49.86 & 0 & 51.88 & 5 & 50.43 & 0 & 50.43 & 5 & 50.49 & 2 & 52.49 & 9 \\
& MMLU & 31.48 & 0 & 39.98 & 2 & 44.87 & 0 & 43.76 & 2 & 45.53 & 2 & 45.58 & 3 \\
\bottomrule
\end{tabular}}
\caption{Comparison of \textbf{Llama 3.1 8B} and \textbf{Qwen 0.5B} across Winogrande, MMLU, and CommonQA under different pruning and fine-tuning regimes. 
Columns denote: (i) Baseline = original model, (ii) Pruned Only = \method without fine-tuning, (iii) FT Only = fine-tuned without pruning, (iv) Prune $\rightarrow$ FT = prune then fine-tune, (v) FT $\rightarrow$ Prune = fine-tune then prune, (vi) (Prune $\rightarrow$ FT) $\rightarrow$ Prune = best fine-tuned-pruned model further pruned. Perf. = performance score, \#D = number of deleted layers.}
\label{tab:model_comparison}
\end{table*}

\paragraph{Fine-tuning.}
We next investigate how layer pruning interacts with fine-tuning. Since pruning reduces model representational capacity, one might expect it to negatively impact fine-tuning performance compared to baseline instruct-tuned models. However, our results show the opposite: \method {\bf not only preserves fine-tuning results but can even improve accuracy and efficiency}.


We explored four settings: (i) fine-tuning the base model (FT), (ii) applying \method after fine-tuning (FT $\rightarrow$ \method), (iii) pruning first and then fine-tuning (\method $\rightarrow$ FT), and (iv) pruning first, then fine-tuning, and finally pruning again (\method $\rightarrow$ FT $\rightarrow$ \method). Across various benchmarks, we consistently observed  mostly moderate and sometimes significant gains after iterating pruning and fine-tuning, especially on Winogrande  and GSM8K (Table~\ref{tab:model_comparison}). This suggests that pruning can act as a regularizer, simplifying the optimization landscape by removing redundant layers.

\method also reduced computation costs for fine-tuning.
For example, pruning LLaMA-3.1 8B prior to fine-tuning reduces training time by approximately 18.5\% (2--2.5 GPU hours on an A100) while improving Winogrande performance by +2.4\%.
Iteratively applying pruning and fine-tuning allowed us to prune up to 8 layers achieving still higher accuracy (87.37\%) than the full fine-tuned model (85.00\%). Similarly, pruning the fully fine-tuned model yielded a 7-layer reduction while maintaining strong accuracy (86.66\%). 



\paragraph{Summary.}
Overall, these results reveal a consistent pattern: \method {\bf synergizes with few-shot learning and fine-tuning}. Rather than limiting model capacity in a harmful way, task-aware pruning removes detrimental components, leading to models that are both more efficient and, in many cases, more accurate.
\section{Discussion}
\label{sec:discussion}

We summarize six key observations from our experiments below.  

\paragraph{a. Task dependency of layer importance} 
The literature offers differing views on layer importance. Some argue that early layers are essential \citep{dalvi:etalL:2020}, while others emphasize the importance of later layers \cite{ellie, latestlayers1, latestlayers2}. Our findings show that layer importance is fundamentally task-specific; for example, removing early layers reduces accuracy to near zero on commonsense reasoning tasks (Figure~\ref{drop1}), whereas removing LLaMA’s layer 3 improves performance on GSM8K-Hard. 

\paragraph{b. Related tasks often exhibit similar layer dependencies} 
Common sense reasoning tasks (see Figure~\ref{drop1}) show importance concentrated in comparable regions of the network. All models show sizable accuracy boosts in mathematical reasoning tasks after pruning between one and three early-to-middle layers (e.g., LLaMA layer 3, Mistral layers 6 and 22, Lucie layer 12) (Figures~\ref{deletedllama}, \ref{deletedqwen}, \ref{deletedlucie}). By contrast, knowledge-intensive tasks (ARC, BoolQ, CommonsenseQA, Winogrande, and BIG-Bench) exhibit more modest improvements (although LLaMA shows an 11\% gain on BIG-Bench) and benefit more from removing later layers.  
These results may aid model interpretability, as plotting performance degradation from layer ablations helps localize task-specific capabilities within the network. 

Initial multilingual testing of \method on Lucie, which is tuned for French conversational proficiency \cite{gouvert:etal:2025}, using bilingual versions of the same dataset, shows that optimal pruning is task-specific rather than language-specific.  

\paragraph{c. Layer redundancy persists even in single-task training.}  
To isolate this effect, we trained a transformer from scratch on a single task (in-context learning of linear functions), hypothesizing that layer redundancy might arise primarily from multi-task pretraining. However, even in this controlled setting, several layers proved redundant, and some even degraded performance (Figure~\ref{evolution}).

\paragraph{d. Generality} 
In principle, \method can combine tasks to produce more general optimized models. A LLaMA math model without layer 12 improves over the baseline LLaMA on Math500 and GSM8K tasks. A promising direction is to prune models jointly across multiple tasks using different mixtures of data to guide the pruning process. 

\paragraph{e. Model-specific effects with \method}  
\method affects different models differently. While LLaMA benefits the least from \method, Lucie achieves large gains on MMLU and double-digit improvements on ARC-Challenge, CommonsenseQA, BoolQ, and GSM8K-hard. \method confers more modest but still substantial gains on Qwen-7B and Mistral. Lucie also benefits from more aggressive pruning than the other models.  

The fact that Lucie was trained on a much smaller dataset (3T tokens vs.\ 15T for LLaMA and 13T for Qwen) suggests intriguing interactions between pretraining and \method improvements. We hypothesize that models trained close to their performance ceiling (via large-scale pretraining, instruction tuning, or RLHF) yield smaller gains from \method, whereas models trained under more limited objectives may benefit more.

\paragraph{f. MI Analysis} 
We use Mutual Information (MI) to investigate why selectively removing layers can improve accuracy, focusing on how information about the output evolves as it propagates through the network. Unlike correlation, MI captures non-linear statistical dependencies and thus provides a more complete measure of dependence \cite{kinney}. We estimate MI using MINE \cite{belghazi2018mine}, a widely used approximation method. 

Our analysis reveals that many layers exhibit a pronounced drop in MI (Figure~\ref{MIllama}). \method removes some of these layers, but not all; overall, it reduces the peaks and valleys in the MI profile across layers. However, removing all layers associated with decreases in MI leads to very poor performance. This suggests that some local decreases in MI across adjacent layers are necessary for proper model functioning (see Appendix~\ref{sec:MI} for details).

\section{Conclusions}

We introduced {\sc Tale}, a simple and effective task-aware layer elimination strategy that removes layers irrelevant to a target task $T$. Across a wide range of models and benchmarks, \method improves the performance of the base model while reducing computational cost, outperforming prior training-free pruning approaches.

Unlike existing methods that rely on task-agnostic heuristics, \method directly optimizes task-specific validation accuracy, enabling it to identify compact, task-specialized architectures without retraining. We further show that \method complements both few-shot prompting and fine-tuning, often yielding additional gains in both accuracy and efficiency.

Beyond performance improvements, \method provides a practical and flexible approach to adapting large language models post hoc. It can be applied to pre-trained, instruction-tuned, and fine-tuned models, making it suitable for a wide range of deployment scenarios. In particular, \method is well-suited for high-throughput and resource-constrained settings, such as multi-agent systems and interactive applications, where balancing model capability and computational efficiency is critical.

Overall, our results highlight the importance of task-aware model adaptation and suggest that selectively removing misaligned components can be as beneficial as adding capacity, opening new directions for efficient and specialized use of large language models.

\section*{Limitations}

While TALE demonstrates consistent gains across a broad range of models and tasks, we note several limitations.

First, TALE operates at the level of \emph{entire transformer layers}. This design choice prioritizes simplicity, transparency, and compatibility with existing checkpoints, but it does not exploit finer-grained structure such as attention heads, blocks, or token-level adaptivity. More granular structured pruning or adaptive computation methods may provide complementary benefits, particularly when retraining or architectural modification is permissible.

Second, TALE performs \emph{task-specific} layer selection using a held-out optimization split. As a result, the resulting pruned architectures are specialized to individual tasks rather than universally optimal across tasks. While this specialization aligns with deployment scenarios where the target task is known in advance, it may be less suitable in settings that require a single model to perform well across many heterogeneous tasks without reconfiguration.

Third, TALE relies on a greedy elimination procedure and a stopping tolerance hyperparameter to determine when further layer removal becomes detrimental. Although we observe stable behavior across random seeds and alternative data splits, more principled global optimization strategies or adaptive stopping criteria could further improve robustness and are left for future work.

Finally, our empirical comparisons focus on training-free, layer-level pruning methods that preserve the original model architecture. Approaches that rely on retraining, block-level restructuring, or task-adaptive control flow address a different point in the design space and are therefore not directly comparable within our experimental scope.

Overall, these limitations reflect deliberate design choices rather than deficiencies of the approach. We view TALE as complementary to retraining-based and fine-grained pruning methods, and believe that combining task-aware layer selection with such approaches is a promising direction for future research.

\section*{Acknowledgments}
 
We gratefuly acknowledge the support of the grants SARER, Summ-RE (ANR-20-CE23-0017), and the AI Cluster ANITI (ANR-19-PI3A-0004). This work used the HPC resources from CALMIP (Grant 2016-P23060 and M24047).

\bibliography{custom}

\appendix

\section{Implementation Details}
\label{appendix:impl}

\providecommand{\perc}[1]{\textcolor{green!50!black}{\small(+#1\% $\uparrow$)}}
\providecommand{\percd}[1]{\textcolor{green!50!black}{\small-#1\%}}
\providecommand{\percdd}[1]{\textcolor{red!50!black}{\small(-#1\% $\downarrow$)}}
\providecommand{\del}[2]{\colorbox{#1!20}{\texttt{\textcolor{#1!90!black}{#2}}}}

\vspace{0.5cm}


\paragraph{Hardware.} 
All experiments were conducted on 1 NVIDIA A100 GPU with 80GB memory.  

\paragraph{Models.} 
We applied \method to five open-weights LLMs of varying scales: \textbf{Qwen2.5-0.5B-Instruct}, \textbf{Qwen2.5-7B-Instruct}, \textbf{Lucie-7B-Instruct}, \textbf{Mistral-7B-Instruct}, and \textbf{Llama-3.1-8B-Instruct}.  

\paragraph{Datasets for \method pruning.} 
The greedy layer-pruning algorithm was evaluated across nine widely used benchmarks covering reasoning, commonsense, and knowledge-intensive tasks: \textbf{ARC-Challenge}, \textbf{ARC-Easy}, \textbf{MMLU}, \textbf{Winogrande}, \textbf{GSM8K}, \textbf{MATH500}, \textbf{CommonQA}, \textbf{BIG-Bench}, and \textbf{BoolQ}.  

\paragraph{Pruning setup.} 
At each iteration, \method evaluates all candidate single-layer deletions with respect to validation accuracy. The pruning threshold was defined as the baseline accuracy -8\% of the full model, ensuring that pruning never reduces performance relative to the original unpruned model. The iterative procedure terminates once no further layer removals satisfy this criterion.

\paragraph{Fine-tuning setup.} 
For fine-tuning experiments, we focused on \textbf{Winogrande} and \textbf{MMLU}. We employed LoRA with rank 64, a batch size of 4, and the optimizer \texttt{paged\_adamw\_32bit}. A cosine learning rate scheduler was used, and models were trained for 10 epochs.  

\hidden{
\subsection{Data splits and leakage prevention}
\label{sec:data_splits}

For each task, we explicitly separate the data used for layer optimization from the data used for reporting results. Let $D$ denote the full dataset. We partition $D$ into two disjoint subsets: a training set $D_{\text{train}}$ and an evaluation set $D_{\text{eval}}$. From $D_{\text{train}}$, we further sample a small held-out subset $D_{\text{opt}} \subset D_{\text{train}}$, which is used exclusively by TALE to guide layer elimination.

At no point does TALE access $D_{\text{eval}}$ during optimization. Final accuracy is computed only on $D_{\text{eval}}$. When official test sets are unavailable, $D_{\text{eval}}$ corresponds to a held-out validation split that is strictly disjoint from $D_{\text{opt}}$.
}

\paragraph{Evaluation.} 
 The LM-Eval  methodology presents a significant limitation: it selects the answer with the highest probability among the provided options rather than assessing what the model would actually generate. This approach ignores hallucination behavior and systematically inflates scores; for example, in a two-choice setting, a hallucinated answer still has a 50\% chance of being counted as correct. Furthermore, LM-Eval often assigns relatively high scores to weak models, compressing performance differences and making stronger approaches appear only marginally better despite substantial real-world gains. This produces a misleading picture of model capability, as high LM-Eval results do not guarantee that a model will produce correct, coherent outputs in practice. For these reasons, we relied primarily on Decoder Eval that measures actual accuracy based on the model’s generated outputs, which we implemented for each task.

\paragraph{Prompting.} 
For zero-shot and few-shot evaluation, we used task-specific prompts. Below we show the prompt used for datasets, consisting of a system instruction :  
\begin{tcolorbox}[colback=teal!3!white,
                  colframe=teal!60!black,
                  title=ARC-E \& ARC-C  System Prompt,
                  fonttitle=\bfseries,
                  boxrule=0.3pt,
                  arc=2pt,
                  left=4pt,
                  right=4pt,
                  top=4pt,
                  bottom=4pt]
\small
You are a Science expert assistant.  
Your task is to answer multiple-choice science questions at grade-school level.  
Each question has four answer choices, labeled A, B, C, and D.  

For each question:  
- Carefully read the question and all answer choices.  
- Select the single best answer from the options (A, B, C, or D).  
- Respond only with the letter of the correct answer, and nothing else—no explanation or extra words.  

Be precise and consistent: Only the answer letter.
\end{tcolorbox}

\begin{tcolorbox}[colback=teal!3!white,
                  colframe=teal!60!black,
                  title= Bigbench System Prompt,
                  fonttitle=\bfseries,
                  boxrule=0.3pt,
                  arc=2pt,
                  left=4pt,
                  right=4pt,
                  top=4pt,
                  bottom=4pt]
\small
"You are a boolean expression evaluator. You must respond with exactly one word: either 'True' or 'False'. 
Do not provide explanations, steps, or any other text. Only respond with 'True' or 'False'."
\end{tcolorbox}

\begin{tcolorbox}[colback=teal!3!white,
                  colframe=teal!60!black,
                  title= BOOLQ System Prompt,
                  fonttitle=\bfseries,
                  boxrule=0.3pt,
                  arc=2pt,
                  left=4pt,
                  right=4pt,
                  top=4pt,
                  bottom=4pt]
\small
"You are a helpful assistant that answers True/False questions based on given passages. 
Read the passage carefully and determine if the question can be answered as True or False based on the information in the passage. 
"Respond with only 'A' for True or 'B' for False."
\end{tcolorbox}


\begin{tcolorbox}[colback=teal!3!white,
                  colframe=teal!60!black,
                  title= CommonQA System Prompt,
                  fonttitle=\bfseries,
                  boxrule=0.3pt,
                  arc=2pt,
                  left=4pt,
                  right=4pt,
                  top=4pt,
                  bottom=4pt]
\small
"You are a helpful assistant that answers multiple-choice questions requiring commonsense knowledge and reasoning. Read each question carefully and select the most logical answer from the given options based on common knowledge and reasoning. Respond with only the letter of your chosen answer (A, B, C, D, or E)."
\end{tcolorbox}

\begin{tcolorbox}[colback=teal!3!white,
                  colframe=teal!60!black,
                  title= GSM8K System Prompt,
                  fonttitle=\bfseries,
                  boxrule=0.3pt,
                  arc=2pt,
                  left=4pt,
                  right=4pt,
                  top=4pt,
                  bottom=4pt]
\small
"You are a math problem solver. Solve the given math problem step by step. "
"Show your complete reasoning and calculations. "
"At the end, write your final answer after '\#\#\#\#' like this: \#\#\#\# [your final numerical answer]""
\end{tcolorbox}

\begin{tcolorbox}[colback=teal!3!white,
                  colframe=teal!60!black,
                  title= MMLU System Prompt,
                  fonttitle=\bfseries,
                  boxrule=0.3pt,
                  arc=2pt,
                  left=4pt,
                  right=4pt,
                  top=4pt,
                  bottom=4pt]
\small
"You are a helpful assistant that answers multiple-choice questions across various academic subjects including humanities, social sciences, STEM, and professional fields. Read each question carefully and select the best answer from the given options. Respond with only the letter of your chosen answer (A, B, C, or D)."
\end{tcolorbox}

\begin{tcolorbox}[colback=teal!3!white,
                  colframe=teal!60!black,
                  title=MATH500 System prompt,
                  fonttitle=\bfseries,
                  boxrule=0.4pt,
                  arc=3pt,
                  left=6pt,
                  right=6pt,
                  top=6pt,
                  bottom=6pt]
\small
You are a careful math problem solver. Show complete step-by-step reasoning
and all calculations needed to arrive at the answer. Use clear, numbered or
labeled steps so the reasoning is easy to follow.

\textbf{IMPORTANT (formatting):}
\begin{itemize}
    \item After the full reasoning, write the \textbf{final answer on a new line by itself} in exactly this format:
    \begin{quote}
        \#\#\#\# \\
        \textit{integer}
    \end{quote}
    \item \texttt{<integer>} must be digits only, optionally with a leading ``-'' for negatives (e.g., \texttt{-7}).
    \item Do \textbf{not} add words, punctuation, units, or commentary on the same line as the \#\#\#\# line.
    \item The \#\#\#\# line must be the \textbf{final line of the output} (nothing may follow it).
    \item Assume all problems expect integer answers; ensure the final line contains a single integer.
\end{itemize}
\end{tcolorbox}
\subsection{Layer Removal Implementation}
\label{sec:layer_removal}

Layer pruning was implemented through a custom model wrapper that reconstructs the architecture excluding specified layers. Given delete indices $\mathcal{D}$, we create a \texttt{ModifiedModel} that:

\begin{enumerate}
    \item Preserves the embedding layer, final normalization, and language modeling head from the original model
    \item Constructs a new layer sequence $\mathcal{L}' = \{l_i \mid i \notin \mathcal{D}\}$ by filtering out deleted layers while maintaining the order of retained layers
    \item Updates the model configuration to reflect the new layer count
\end{enumerate}

The forward pass implements standard transformer computation: input embeddings are passed through the retained layers sequentially with causal attention masking, then normalized and projected to vocabulary logits. Position embeddings are generated automatically if not provided. This architecture is fully compatible with the Hugging Face training pipeline and can be directly used with LoRA fine-tuning without requiring custom training loops.
\subsection{Fine-tuning Training Details}
\label{sec:finetuning_details}

All fine-tuning experiments were conducted using Parameter-Efficient Fine-Tuning (PEFT) with Low-Rank Adaptation (LoRA). We employed 4-bit quantization using the BitsAndBytes library to reduce memory footprint during training. The quantization configuration used NF4 (4-bit NormalFloat) quantization type with float16 compute dtype, without nested quantization.

\paragraph{LoRA Configuration:} 
We applied LoRA to all linear layers in the model with the following hyperparameters: rank $r=64$, alpha $\alpha=16$, and dropout rate of $0.1$. These parameters were kept consistent across both full and pruned models to ensure fair comparison.

\paragraph{Optimization Settings:}
Training was performed for 10 epochs using the paged AdamW optimizer (32-bit) with a learning rate of $2 \times 10^{-4}$ and weight decay of $0.001$. We used a cosine learning rate schedule with a warmup ratio of $0.03$. Gradient clipping was applied with a maximum gradient norm of $0.3$. The effective batch size was 60 (per-device batch size of 2 with gradient accumulation steps of 30). Gradient checkpointing was enabled to reduce memory consumption during training.

\paragraph{Data Processing:}
All sequences were truncated or padded to a maximum length of 300 tokens. We used right-side padding with a special padding token (ID: 128004). Packing was disabled to maintain sequence boundaries, and we removed unused columns from the dataset. The dataloader used 4 workers with the last incomplete batch dropped to ensure consistent batch sizes.

\paragraph{Hardware and Implementation:}
All experiments were conducted on NVIDIA A100 GPUs. We used mixed-precision training without fp16 or bf16 enabled at the trainer level, relying instead on the 4-bit quantization for memory efficiency. Training logs were reported to TensorBoard every 25 steps.


\section{Intuition behind \method}
\begin{figure}[!ht]
    \centering
    \begin{subfigure}[t]{0.48\textwidth}
        \centering
        \includegraphics[width=0.8\linewidth]{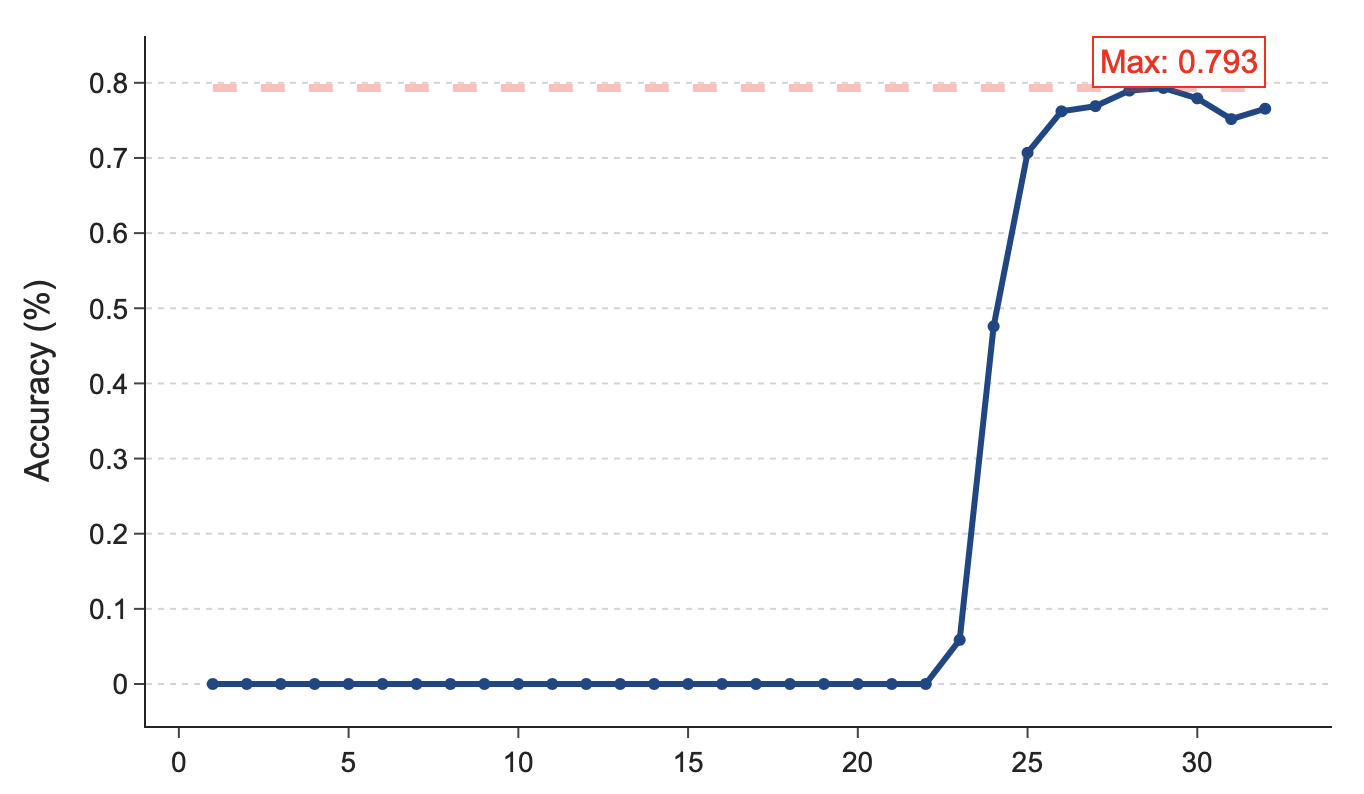}
        \caption{ARC-Challenge}
    \end{subfigure}
    \hfill
    \begin{subfigure}[t]{0.48\textwidth}
        \centering
        \includegraphics[width=0.8\linewidth]{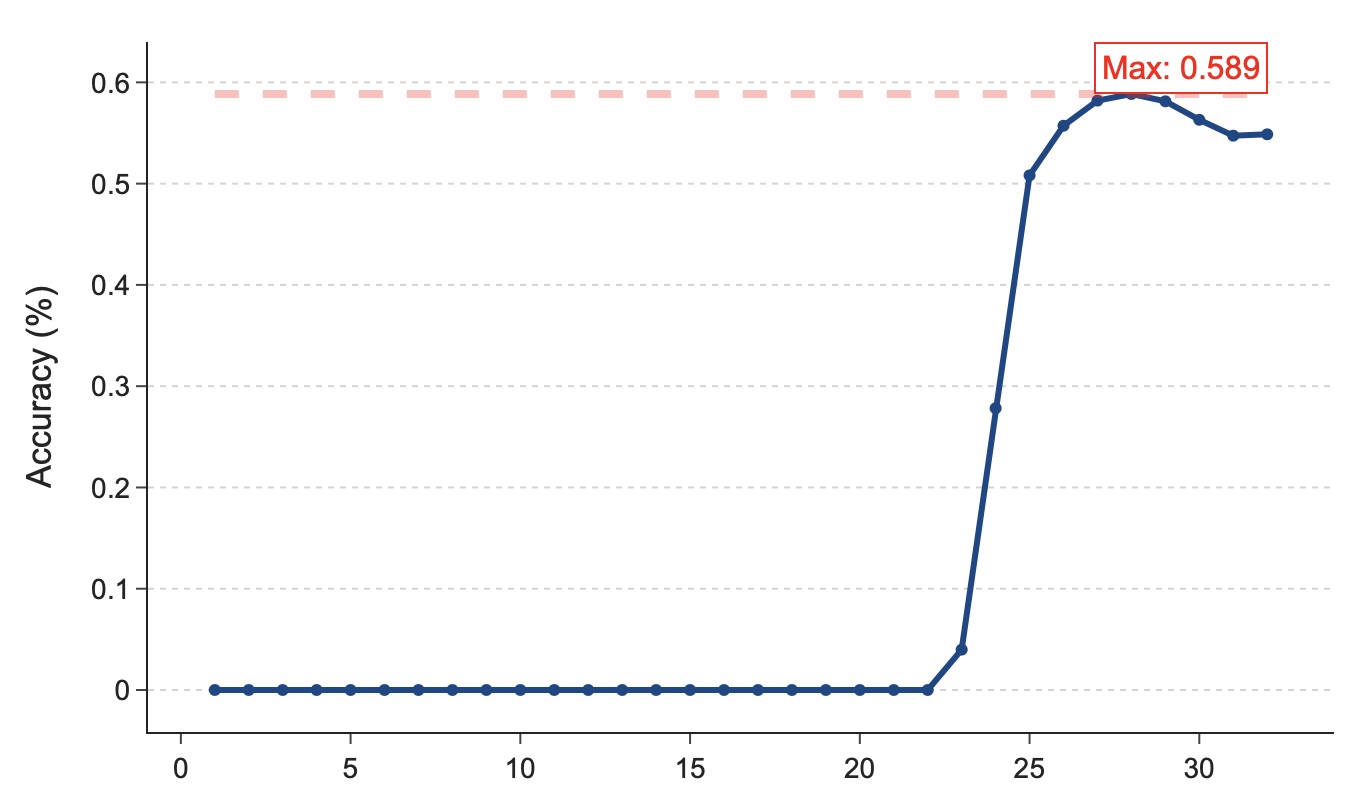}
        \caption{MMLU}
    \end{subfigure}
    \caption{Layer-wise output performance for LLaMA models: results when generating predictions from intermediate layers 1 through 32 on two different datasets.}
    \label{graph:oldexp}
\end{figure}

\newpage
\section{Number of parameters per layer for each model}
\label{appendix:nbparameters}

\begin{table}[!ht]
\centering
\renewcommand{\arraystretch}{1.3}
\begin{tabular}{l|r|c}
\toprule
\textbf{Model} & \textbf{Params/Layer} & \textbf{Layers} \\
\midrule
LLaMA 3.1 8B & 218,112,000 & 32 \\
Qwen 2.5 7B & 233,057,792 & 28 \\
Mistral 7B & 218,112,000 & 32 \\
Lucie 7B & 192,946,176 & 32 \\
Qwen 2.5 0.5B & 14,912,384 & 24 \\
\bottomrule
\end{tabular}
\caption{Model parameter counts comparison showing parameters per layer and total number of layers.}
\label{tab:model_parameters}
\end{table}

\newpage

\section{Ablation study on validation Set of Pruning}
\label{appendix:ablation}
We analyze the effect of validation set size on TALE's layer selection. Table \ref{table_ablation} reports the specific layers dropped for different validation set sizes across three tasks (ARC-Easy, MMLU, GSM8K) and two models (Llama 3.1 8B, Qwen 2.5 7B). 
\begin{table}[ht!]
\centering
\small
\setlength{\tabcolsep}{4pt}
\renewcommand{\arraystretch}{1.25}

\resizebox{\linewidth}{!}{
\begin{tabular}{lccc}
\toprule
\textbf{Model} & \textbf{Val Size} & \textbf{Task} & \textbf{Dropped Layers} \\
\midrule

\multirow{9}{*}{\textbf{Llama 3.1 8B}}
 & \multirow{3}{*}{200} 
    & ARC-E   & \{19, 20, 22, 29, 32  \} \\
 & & MMLU    & \{  21 \} \\
 & & GSM8K   & \{ 3 \} \\
\cmidrule(lr){2-4}
 & \multirow{3}{*}{500} 
    & ARC-E   & \{ 19, 20, 21, 29, 32 \} \\
 & & MMLU    & \{ 21 \} \\
 & & GSM8K   & \{ 3 \} \\
\cmidrule(lr){2-4}
 & \multirow{3}{*}{1000} 
    & ARC-E   & \{ 19, 20, 21, 29, 32 \} \\
 & & MMLU    & \{ 21 \} \\
 & & GSM8K   & \{ 3 \} \\
\midrule

\multirow{9}{*}{\textbf{Qwen 2.5 7B}}
 & \multirow{3}{*}{100} 
    & ARC-E   & \{ 22 , 27 , 28 \} \\
 & & MMLU    & \{ 18 , 22 , 24 , 27 , 28 \} \\
 & & GSM8K   & \{ 19 \} \\
\cmidrule(lr){2-4}
 & \multirow{3}{*}{500} 
    & ARC-E   & \{ 19, 22 , 28 \} \\
 & & MMLU    & \{ 22  , 23 , 26 , 27 , 28 \} \\
 & & GSM8K   & \{ 19 \} \\
\cmidrule(lr){2-4}
 & \multirow{3}{*}{1000} 
    & ARC-E   & \{ 19 , 22 , 28 \} \\
 & & MMLU    & \{ 22 , 23 , 26 , 27 , 28 \} \\
 & & GSM8K   & \{ 19 \} \\
\bottomrule
\end{tabular}
}

\caption{
Layers removed by TALE for different validation-set sizes 
across three tasks.  
This reveals the stability of pruning decisions directly.
}\label{table_ablation}
\end{table}

\newpage

\providecommand{\neutral}[1]{\textcolor{gray!70!black}{#1}}

\begin{table*}[!ht]
 \renewcommand{\arraystretch}{1.3}
 \setlength{\tabcolsep}{6pt}

 \begin{subtable}[t]{\textwidth}
 \centering
 \resizebox{\linewidth}{!}{
 \begin{tabular}{l|c|ccc|ccc|c|ccc|ccc|}
 \toprule
 \rowcolor{teal!30}
 \multirow{2}{*}{\textbf{Dataset}} & 
 \multicolumn{7}{c|}{\textbf{LLaMA 3.1 8B (0-shot)}} & 
 \multicolumn{7}{c}{\textbf{Qwen 2.5 7B (0-shot)}}\\
 \cmidrule(lr){2-8} \cmidrule(lr){9-15} 
  & Baseline & \multicolumn{3}{c|}{Best Model} & \multicolumn{3}{c|}{BSBA} 
  & Baseline & \multicolumn{3}{c|}{Best Model} & \multicolumn{3}{c}{BSBA} \\
 \cmidrule(lr){2-2} \cmidrule(lr){3-5} \cmidrule(lr){6-8} 
 \cmidrule(lr){9-9} \cmidrule(lr){10-12} \cmidrule(lr){13-15}
  & Perf. & Perf. $\pm$ Std & \#D & Sp. & Perf. & \#D & Sp. 
  & Perf. & Perf. $\pm$ Std & \#D & Sp. & Perf. & \#D & Sp. \\
 \midrule
 ARC-Easy       & 89 & 90.4 $\pm$ 0.14 & 5 & \percd{14.6} & 88.8 & 8 & \percd{23.5} 
                & 90.04 & 93.2 $\pm$ 0.16 & 3 & \percd{10.0} & 90.08 & 7 & \percd{30.3} \\
 ARC-Challenge  & 79.4 & 80.6 $\pm$ 0.18 & 4 & \percd{11.7} & 77.6 & 7 & \percd{20.5} 
                & 86.55 & 92.00 $\pm$ 0.15 & 2 & \percd{6.7} & 86.55 & 6 & \percd{19.9} \\
 BoolQ          & 85.4 & 85.9 $\pm$ 0.12 & 3 & \percd{8.8} & 85.4 & 7 & \percd{17.6} 
                & 81.90 & 83.90 $\pm$ 0.14 & 4 & \percd{13.3} & 82.70 & 5 & \percd{23.2} \\
 MMLU           & 48.8 & 53.8 $\pm$ 0.22 & 1 & \percd{2.9} & 50.2 & 9 & \percd{26.4} 
                & 68.10 & 71.00 $\pm$ 0.19 & 5 & \percd{16.6} & 68.13 & 6 & \percd{19.9} \\
 CommonQA       & 72.9 & 73.8 $\pm$ 0.15 & 3 & \percd{8.8} & 73.10 & 6 & \percd{17.6} 
                & 80.30 & 84.40 $\pm$ 0.17 & 2 & \percd{6.6} & 80.50 & 6 & \percd{19.9} \\
 Winogrande     & 53.8 & 54.1 $\pm$ 0.20 & 4 & \percd{11.7} & 53.83 & 12 & \percd{32.2} 
                & 62.04 & 67.25 $\pm$ 0.21 & 3 & \percd{10.0} & 62.19 & 6 & \percd{19.9} \\
 BIG-Bench      & 77.2 & 85.6 $\pm$ 0.25 & 5 & \percd{14.4} & 76.4 & 11 & \percd{32.2} 
                & 79.20 & 81.60 $\pm$ 0.14 & 6 & \percd{19.9} & 81.60 & 6 & \percd{19.9} \\
 GSM8K-HARD     & 39.0 & 59.0 $\pm$ 0.30 & 1 & \percd{2.9} & 39.4 & 4 & \percd{11.7} 
                & 43.80 & 61.80 $\pm$ 0.28 & 2 & \percd{43.6} & 43.99 & 5 & \percd{17.6} \\
 Math500        & 25.4 & 28.2 $\pm$ 0.18 & 2 & \percd{6.0} & 27.4 & 3 & \percd{9.1} 
                & 31.00 & 38.20 $\pm$ 0.22 & 2 & \percd{6.6} & 32.10 & 4 & \percd{13.3} \\
 \bottomrule
 \end{tabular}
 }
 \end{subtable}
 
 \vspace{1em}
 
 \begin{subtable}[t]{\textwidth}
 \centering
 \resizebox{\linewidth}{!}{
 \begin{tabular}{l|c|ccc|ccc|c|ccc|ccc|}
 \toprule
 \rowcolor{teal!30}
 \multirow{2}{*}{\textbf{Dataset}} & 
 \multicolumn{7}{c|}{\textbf{Lucie 7B (0-shot)}} & 
 \multicolumn{7}{c}{\textbf{Mistral 7B (0-shot)}}\\
 \cmidrule(lr){2-8} \cmidrule(lr){9-15}
  & Baseline & \multicolumn{3}{c|}{Best Model} & \multicolumn{3}{c|}{BSBA} 
  & Baseline & \multicolumn{3}{c|}{Best Model} & \multicolumn{3}{c}{BSBA} \\
 \cmidrule(lr){2-2} \cmidrule(lr){3-5} \cmidrule(lr){6-8}
 \cmidrule(lr){9-9} \cmidrule(lr){10-12} \cmidrule(lr){13-15}
  & Perf. & Perf. $\pm$ Std & \#D & Sp. & Perf. & \#D & Sp.
  & Perf. & Perf. $\pm$ Std & \#D & Sp. & Perf. & \#D & Sp. \\
 \midrule
 ARC-Easy        & 74.4 & 75.8 $\pm$ 0.15 & 6 & \percd{18.1} & 73.8 & 8 & \percd{23.5}
                 & 83.8 & 85.6 $\pm$ 0.14 & 5 & \percd{15.4} & 82.8 & 9 & \percd{27.7} \\
 ARC-Challenge   & 46.0 & 51.45 $\pm$ 0.22 & 7 & \percd{22.1} & 48.8 & 11 & \percd{33.1}
                 & 76.2 & 79.1 $\pm$ 0.18 & 6 & \percd{18.5} & 76.2 & 8 & \percd{24.6} \\
 BoolQ           & 53.0 & 74.0 $\pm$ 0.30 & 5 & \percd{17.2} & 63.0 & 19 & \percd{54.2}
                 & 81.3 & 84.4 $\pm$ 0.16 & 4 & \percd{18.5} & 80.8 & 5 & \percd{27.7} \\
 MMLU            & 13.0 & 54.0 $\pm$ 0.35 & 8 & \percd{24.1} & 15 & 22 & \percd{60.2}
                 & 39.4 & 40.8 $\pm$ 0.20 & 2 & \percd{6.2} & 39.0 & 8 & \percd{24.6} \\
 CommonQA        & 54.2 & 68.6 $\pm$ 0.27 & 3 & \percd{9.1} & 54.6 & 17 & \percd{48.2}
                 & 61.0 & 64.4 $\pm$ 0.19 & 4 & \percd{12.3} & 61.6 & 7 & \percd{21.5} \\
 Winogrande      & 51.6 & 53.1 $\pm$ 0.21 & 5 & \percd{27.1} & 53.0 & 15 & \percd{45.2}
                 & 53.2 & 54.3 $\pm$ 0.16 & 10 & \percd{30.7} & 52.4 & 13 & \percd{40.0} \\
 BIG-Bench       & 67.4 & 75.0 $\pm$ 0.25 & 9 & \percd{27.1} & 71 & 15 & \percd{45.1}
                 & 70.4 & 75.4 $\pm$ 0.22 & 9 & \percd{28.0} & 72.6 & 11 & \percd{33.8} \\
 GSM8K-HARD      & 32 & 39.0 $\pm$ 0.28 & 1 & \percd{3.1} & 37 & 3 & \percd{9.1}
                 & 24 & 33 $\pm$ 0.24 & 2 & \percd{6.2} & 26.1 & 4 & \percd{12.3} \\
 Math500         & 21.0 & 26.1 $\pm$ 0.20 & 2 & \percd{6.0} & 25.1 & 3 & \percd{9.1}
                 & 19 & 28 $\pm$ 0.23 & 1 & \percd{3.1} & 18.8 & 4 & \percd{12.3} \\
 \bottomrule
 \end{tabular}
 }
\end{subtable}
\caption{Robustness study of the proposed layer-dropping method across multiple language models under zero-shot evaluation. For each dataset and model, results are reported over five random seeds to account for variability in decoding and sampling. We present the baseline model accuracy and the accuracy of the best pruned configuration, along with their corresponding standard deviations computed across the 5 seeds. The table also includes the number of transformer layers removed in the best-performing configuration (\textbf{\#D}) and the resulting inference speedup (\textbf{Sp.}) expressed as the percentage of total TFlops saved during evaluation.  Bold values indicate the highest mean accuracy for each dataset.}
\label{tab:robustness_study}
\end{table*}



\section{Robustness Study on TALE}
\label{appendix:robustness}
See Table \ref{tab:robustness_study}.

\hidden{

\begin{table}[!ht]
\centering

\resizebox{\linewidth}{!}{
\begin{tabular}{l|ccccc}
\toprule
\rowcolor{teal!30}
\textbf{Dataset} & Baseline & Best Perf. & Best (alt val) & \#D & Sp. \\
 & (mean ± std) & (mean $\pm$ std) & (mean $\pm$ std) & & \\
\midrule
ARC-Easy & $87.00 \pm 0.11$ & $90.55 \pm 0.14$ & $90.55 \pm 0.18$ & 5 & \percd{14.6} \\
ARC-Challenge & $75.86 \pm 0.12$ & $78.62 \pm 0.15$ & $78.62 \pm 0.19$ & 4 & \percd{11.7} \\
BoolQ & $85.00 \pm 0.10$ & $86.20 \pm 0.13$ & $86.20 \pm 0.16$ & 3 & \percd{8.8} \\
MMLU & $54.87 \pm 0.12$ & $59.90 \pm 0.16$ & $59.90 \pm 0.20$ & 1 & \percd{2.9} \\
CommonQA & $72.20 \pm 0.10$ & $75.30 \pm 0.14$ & $75.30 \pm 0.17$ & 3 & \percd{8.8} \\
Winogrande & $53.83 \pm 0.11$ & $56.67 \pm 0.15$ & $56.67 \pm 0.18$ & 4 & \percd{11.7} \\
BIG-Bench & $75.20 \pm 0.13$ & $83.60 \pm 0.18$ & $83.60 \pm 0.22$ & 5 & \percd{14.4} \\
GSM8K-HARD & $15.07 \pm 0.09$ & $37.08 \pm 0.20$ & $37.08 \pm 0.24$ & 1 & \percd{2.9} \\
Math500 & $20.50 \pm 0.10$ & $26.00 \pm 0.14$ & $26.00 \pm 0.17$ & 1 & \percd{2.9} \\
\bottomrule
\end{tabular}
}
\caption{Robustness evaluation for LLaMA 3.1 8B (zero-shot) for Best model across 5 random seeds on the train test split from Table \ref{tab:robustness_study}. Best (alt val)" reports the performance when using an alternative validation split (stds slightly larger to reflect extra variability).}
\label{tab:robustness2llama}
\end{table}

\begin{table}[!ht]
\centering
\resizebox{\linewidth}{!}{
\begin{tabular}{l|ccccc}
\toprule
\rowcolor{teal!30}
\textbf{Dataset} & Baseline & Best Perf. & Best (alt val) & \#D & Sp. \\
 & (mean ± std) & (mean $\pm$ std) & (mean ± std) & & \\
\midrule
ARC-Easy & $91.01 \pm 0.10$ & $91.82 \pm 0.13$ & $91.82 \pm 0.16$ & 2 & \percd{10.0} \\
ARC-Challenge & $86.55 \pm 0.09$ & $92.00 \pm 0.16$ & $92.00 \pm 0.20$ & 2 & \percd{6.7} \\
BoolQ & $84.10 \pm 0.11$ & $86.90 \pm 0.15$ & $86.90 \pm 0.18$ & 4 & \percd{13.3} \\
MMLU & $68.10 \pm 0.14$ & $71.00 \pm 0.18$ & $71.00 \pm 0.22$ & 5 & \percd{16.6} \\
CommonQA & $80.30 \pm 0.12$ & $84.40 \pm 0.16$ & $84.40 \pm 0.19$ & 2 & \percd{6.6} \\
Winogrande & $62.04 \pm 0.12$ & $67.25 \pm 0.17$ & $67.25 \pm 0.20$ & 3 & \percd{10.0} \\
BIG-Bench & $79.20 \pm 0.11$ & $81.60 \pm 0.14$ & $81.60 \pm 0.18$ & 6 & \percd{19.9} \\
GSM8K-HARD & $7.90 \pm 0.08$ & $27.00 \pm 0.17$ & $27.00 \pm 0.21$ & 2 & \percd{6.6} \\
Math500 & $18.00 \pm 0.10$ & $27.00 \pm 0.16$ & $27.00 \pm 0.19$ & 2 & \percd{6.6} \\
\bottomrule
\end{tabular}
}
\caption{Robustness evaluation for Qwen 2.5 7B (zero-shot) for Best model across 5 random seeds using a different train test split from Table \ref{tab:robustness2llama}. Best (alt val)" reports the performance when using an alternative validation split (stds slightly larger to reflect extra variability).}
\label{tab:robustness2qwen}
\end{table}

\begin{table}[!ht]
\centering
\resizebox{\linewidth}{!}{
\begin{tabular}{l|c|ccc|ccc}
\toprule
\rowcolor{teal!30}
\textbf{Dataset} & Baseline & \multicolumn{3}{c|}{Best Model} & \multicolumn{3}{c}{BSBA} \\
\cmidrule(lr){3-5} \cmidrule(lr){6-8}
 & Perf. & Perf. $\pm$ Std & \#D & Sp. & Perf. & \#D & Sp. \\
\midrule
ARC-Easy & 74.4 & 75.8 $\pm$ 0.15 & 6 & \percd{18.1} & 73.8 & 8 & \percd{23.5} \\
ARC-Challenge & 46.0 & 51.45 $\pm$ 0.22 & 7 & \percd{22.1} & 48.8 & 11 & \percd{33.1} \\
BoolQ & 53.0 & 74.0 $\pm$ 0.30 & 5 & \percd{17.2} & 63.0 & 19 & \percd{54.2} \\
MMLU & 13.0 & 54.0 $\pm$ 0.35 & 8 & \percd{24.1} & 15 & 22 & \percd{60.2} \\
CommonQA & 54.2 & 68.6 $\pm$ 0.27 & 3 & \percd{9.1} & 54.6 & 17 & \percd{48.2} \\
Winogrande & 51.6 & 53.1 $\pm$ 0.21 & 5 & \percd{27.1} & 53.0 & 15 & \percd{45.2} \\
BIG-Bench & 67.4 & 75.0 $\pm$ 0.25 & 9 & \percd{27.1} & 71 & 15 & \percd{45.1} \\
GSM8K-HARD & 32.0 & 39.0 $\pm$ 0.28 & 1 & \percd{3.1} & 37 & 3 & \percd{9.1} \\
Math500 & 21.0 & 26.1 $\pm$ 0.20 & 2 & \percd{6.0} & 25.1 & 3 & \percd{9.1} \\
\bottomrule
\end{tabular}
}
\caption{Robustness evaluation for Lucie 7B (zero-shot) on Best model across 5 random seeds using the train test split from Table \ref{tab:robustness2llama}. BSBA model values also included. }
\label{tab:robustness2lucie}
\end{table}

\begin{table}[!ht]
\centering
\resizebox{\linewidth}{!}{
\begin{tabular}{l|c|ccc|ccc}
\toprule
\rowcolor{teal!30}
\textbf{Dataset} & Baseline & \multicolumn{3}{c|}{Best Model} & \multicolumn{3}{c}{BSBA} \\
\cmidrule(lr){3-5} \cmidrule(lr){6-8}
 & Perf. & Perf. $\pm$ Std & \#D & Sp. & Perf. & \#D & Sp. \\
\midrule
ARC-Easy & 83.8 & 85.6 $\pm$ 0.14 & 5 & \percd{15.4} & 82.8 & 9 & \percd{27.7} \\
ARC-Challenge & 76.2 & 79.1 $\pm$ 0.18 & 6 & \percd{18.5} & 76.2 & 8 & \percd{24.6} \\
BoolQ & 81.3 & 84.4 $\pm$ 0.16 & 4 & \percd{18.5} & 80.8 & 5 & \percd{27.7} \\
MMLU & 39.4 & 40.8 $\pm$ 0.20 & 2 & \percd{6.2} & 39.0 & 8 & \percd{24.6} \\
CommonQA & 61.0 & 64.4 $\pm$ 0.19 & 4 & \percd{12.3} & 61.6 & 7 & \percd{21.5} \\
Winogrande & 53.2 & 54.3 $\pm$ 0.16 & 10 & \percd{30.7} & 52.4 & 13 & \percd{40.0} \\
BIG-Bench & 70.4 & 75.4 $\pm$ 0.22 & 9 & \percd{28.0} & 72.6 & 11 & \percd{33.8} \\
GSM8K-HARD & 24 & 33 $\pm$ 0.24 & 2 & \percd{6.2} & 26.1 & 4 & \percd{12.3} \\
Math500 & 19 & 28 $\pm$ 0.23 & 1 & \percd{3.1} & 18.8 & 4 & \percd{12.3} \\
\bottomrule
\end{tabular}
}
\caption{Robustness evaluation for Mistral 7B (zero-shot) on Best model across 5 random seeds using the different train test split from Table \ref{tab:robustness2llama}. BSBA values also included.}
\label{tab:robustness2mistral}
\end{table}

\begin{table*}[!ht]
\centering
\begin{subtable}[t]{\textwidth}  

\resizebox{\linewidth}{!}{
\begin{tabular}{l|c|ccccc|c|ccccc|}
\toprule
\rowcolor{teal!30}
\multirow{2}{*}{\textbf{Dataset}} & 
\multicolumn{6}{c|}{\textbf{LLaMA 3.1 8B (0-shot)}} & 
\multicolumn{6}{c}{\textbf{Qwen 2.5 7B (0-shot)}}\\
\cmidrule(lr){2-7} \cmidrule(lr){8-13}
 & Baseline & Best Perf. & Best (alt val) & \#D & Sp. & & Baseline & Best Perf. & Best (alt val) & \#D & Sp. & \\
 & (mean ± std) & (mean $\pm$ std) & (mean $\pm$ std) & & & & (mean $\pm$ std) & (mean $\pm$ std) & (mean ± std) & & & \\
\midrule
ARC-Easy      
& $87.00 \pm 0.11$ 
& $90.55 \pm 0.14$ 
& $90.55 \pm 0.18$ & 5 & \percd{14.6} & 
& $91.01 \pm 0.10$ 
& $91.82 \pm 0.13$ 
& $91.82 \pm 0.16$ & 2 & \percd{10.0} & \\

ARC-Challenge 
& $75.86 \pm 0.12$
& $78.62 \pm 0.15$ 
& $78.62 \pm 0.19$ & 4 & \percd{11.7}
& & $86.55 \pm 0.09$
& $92.00 \pm 0.16$ 
& $92.00 \pm 0.20$ & 2 & \percd{6.7} & \\

BoolQ         
& $85.00 \pm 0.10$
& $86.20 \pm 0.13$ 
& $86.20 \pm 0.16$ & 3 & \percd{8.8}
& & $84.10 \pm 0.11$
& $86.90 \pm 0.15$ 
& $86.90 \pm 0.18$ & 4 & \percd{13.3} & \\

MMLU          
& $54.87 \pm 0.12$
& $59.90 \pm 0.16$ 
& $59.90 \pm 0.20$ & 1 & \percd{2.9}
& & $68.10 \pm 0.14$
& $71.00 \pm 0.18$ 
& $71.00 \pm 0.22$ & 5 & \percd{16.6} & \\

CommonQA      
& $72.20 \pm 0.10$
& $75.30 \pm 0.14$ 
& $75.30 \pm 0.17$ & 3 & \percd{8.8}
& & $80.30 \pm 0.12$
& $84.40 \pm 0.16$ 
& $84.40 \pm 0.19$ & 2 & \percd{6.6} & \\

Winogrande    
& $53.83 \pm 0.11$
& $56.67 \pm 0.15$ 
& $56.67 \pm 0.18$ & 4 & \percd{11.7}
& & $62.04 \pm 0.12$
& $67.25 \pm 0.17$ 
& $67.25 \pm 0.20$ & 3 & \percd{10.0} & \\

BIG-Bench     
& $75.20 \pm 0.13$
& $83.60 \pm 0.18$ 
& $83.60 \pm 0.22$ & 5 & \percd{14.4}
& & $79.20 \pm 0.11$
& $81.60 \pm 0.14$ 
& $81.60 \pm 0.18$ & 6 & \percd{19.9} & \\

GSM8K-HARD    
& $15.07 \pm 0.09$
& $37.08 \pm 0.20$ 
& $37.08 \pm 0.24$ & 1 & \percd{2.9}
& & $7.90 \pm 0.08$
& $27.00 \pm 0.17$ 
& $27.00 \pm 0.21$ & 2 & \percd{6.6} & \\

Math500       
& $20.50 \pm 0.10$
& $26.00 \pm 0.14$ 
& $26.00 \pm 0.17$ & 1 & \percd{2.9}
& & $18.00 \pm 0.10$
& $27.00 \pm 0.16$ 
& $27.00 \pm 0.19$ & 2 & \percd{6.6} & \\
\bottomrule
\end{tabular}
}
\end{subtable}

 \vspace{0.5cm}

 \begin{subtable}[t]{\textwidth}

 \resizebox{\linewidth}{!}{
 \begin{tabular}{l|c|ccc|ccc|c|ccc|ccc|}
 \toprule
 \rowcolor{teal!30}
 \multirow{2}{*}{\textbf{Dataset}} & 
 \multicolumn{7}{c|}{\textbf{Lucie 7B (0-shot)}} & 
 \multicolumn{7}{c}{\textbf{Mistral 7B (0-shot)}}\\
 \cmidrule(lr){2-8} \cmidrule(lr){9-15}
  & Baseline & \multicolumn{3}{c|}{Best Model} & \multicolumn{3}{c|}{BSBA} 
  & Baseline & \multicolumn{3}{c|}{Best Model} & \multicolumn{3}{c}{BSBA} \\
 \cmidrule(lr){2-2} \cmidrule(lr){3-5} \cmidrule(lr){6-8}
 \cmidrule(lr){9-9} \cmidrule(lr){10-12} \cmidrule(lr){13-15}
  & Perf. & Perf. $\pm$ Std & \#D & Sp. & Perf. & \#D & Sp.
  & Perf. & Perf. $\pm$ Std & \#D & Sp. & Perf. & \#D & Sp. \\
 \midrule
 ARC-Easy        & 74.4 & 75.8 $\pm$ 0.15 & 6 & \percd{18.1} & 73.8 & 8 & \percd{23.5}
                 & 83.8 & 85.6 $\pm$ 0.14 & 5 & \percd{15.4} & 82.8 & 9 & \percd{27.7} \\
 ARC-Challenge   & 46.0 & 51.45 $\pm$ 0.22 & 7 & \percd{22.1} & 48.8 & 11 & \percd{33.1}
                 & 76.2 & 79.1 $\pm$ 0.18 & 6 & \percd{18.5} & 76.2 & 8 & \percd{24.6} \\
 BoolQ           & 53.0 & 74.0 $\pm$ 0.30 & 5 & \percd{17.2} & 63.0 & 19 & \percd{54.2}
                 & 81.3 & 84.4 $\pm$ 0.16 & 4 & \percd{18.5} & 80.8 & 5 & \percd{27.7} \\
 MMLU            & 13.0 & 54.0 $\pm$ 0.35 & 8 & \percd{24.1} & 15 & 22 & \percd{60.2}
                 & 39.4 & 40.8 $\pm$ 0.20 & 2 & \percd{6.2} & 39.0 & 8 & \percd{24.6} \\
 CommonQA        & 54.2 & 68.6 $\pm$ 0.27 & 3 & \percd{9.1} & 54.6 & 17 & \percd{48.2}
                 & 61.0 & 64.4 $\pm$ 0.19 & 4 & \percd{12.3} & 61.6 & 7 & \percd{21.5} \\
 Winogrande      & 51.6 & 53.1 $\pm$ 0.21 & 5 & \percd{27.1} & 53.0 & 15 & \percd{45.2}
                 & 53.2 & 54.3 $\pm$ 0.16 & 10 & \percd{30.7} & 52.4 & 13 & \percd{40.0} \\
 BIG-Bench       & 67.4 & 75.0 $\pm$ 0.25 & 9 & \percd{27.1} & 71 & 15 & \percd{45.1}
                 & 70.4 & 75.4 $\pm$ 0.22 & 9 & \percd{28.0} & 72.6 & 11 & \percd{33.8} \\
 GSM8K-HARD      & 32.0 & 39.0 $\pm$ 0.28 & 1 & \percd{3.1} & 37 & 3 & \percd{9.1}
                 & 24 & 33 $\pm$ 0.24 & 2 & \percd{6.2} & 26.1 & 4 & \percd{12.3} \\
 Math500         & 21.0 & 26.1 $\pm$ 0.20 & 2 & \percd{6.0} & 25.1 & 3 & \percd{9.1}
                 & 19 & 28 $\pm$ 0.23 & 1 & \percd{3.1} & 18.8 & 4 & \percd{12.3} \\
 \bottomrule
 \end{tabular}
 }
 \end{subtable}
\caption{Robustness evaluation across 5 random seeds (mean ± std) for LLaMA 3.1 8B, Qwen 2.5 7B, Lucie 7b and Mistral 7b. "Best (alt val)" reports the performance on LLama and Qwen 7b when using an alternative validation split (stds slightly larger to reflect extra variability).}\label{tab:robustness2}
\end{table*}
}
\hidden{
\begin{table}[t]
\centering
\resizebox{\linewidth}{!}{
\begin{tabular}{l|c|ccc|c|ccc|}
\toprule
\rowcolor{orange!30}
\multirow{2}{*}{\textbf{Dataset}} & 
\multicolumn{4}{c|}{\textbf{LLaMA 3.1 8B (0-shot)}} & 
\multicolumn{4}{c}{\textbf{Qwen 2.5 7B (0-shot)}}\\
\cmidrule(lr){2-5} \cmidrule(lr){6-9}
 & Baseline & Best Perf. & \#D & Sp. 
 & Baseline & Best Perf. & \#D & Sp. \\
\midrule
ARC-Easy      
& $87.00 \pm 0.11$ 
& $90.55 \pm 0.14$\perc{4.1} & 5 & \percd{14.6} 
& $91.01 \pm 0.10$ 
& $91.82 \pm 0.13$\perc{2.0} & 2 & \percd{10.0} \\

ARC-Challenge 
& $75.86 \pm 0.12$
& $78.62 \pm 0.15$\perc{3.7} & 4 & \percd{11.7}
& $86.55 \pm 0.09$
& $92.00 \pm 0.16$ & 2 & \percd{6.7} \\

BoolQ         
& $85.00 \pm 0.10$
& $86.20 \pm 0.13$ & 3 & \percd{8.8}
& $84.10 \pm 0.11$
& $86.90 \pm 0.15$ & 4 & \percd{13.3} \\

MMLU          
& $54.87 \pm 0.12$
& $59.90 \pm 0.16$ & 1 & \percd{2.9}
& $68.10 \pm 0.14$
& $71.00 \pm 0.18$ & 5 & \percd{16.6} \\

CommonQA      
& $72.20 \pm 0.10$
& $75.30 \pm 0.14$ & 3 & \percd{8.8}
& $80.30 \pm 0.12$
& $84.40 \pm 0.16$ & 2 & \percd{6.6} \\

Winogrande    
& $53.83 \pm 0.11$
& $56.67 \pm 0.15$ & 4 & \percd{11.7}
& $62.04 \pm 0.12$
& $67.25 \pm 0.17$ & 3 & \percd{10.0} \\

BIG-Bench     
& $75.20 \pm 0.13$
& $83.60 \pm 0.18$ & 5 & \percd{14.4}
& $79.20 \pm 0.11$
& $81.60 \pm 0.14$ & 6 & \percd{19.9} \\

GSM8K-HARD    
& $15.07 \pm 0.09$
& $37.08 \pm 0.20$ & 1 & \percd{2.9}
& $7.90 \pm 0.08$
& $27.00 \pm 0.17$ & 2 & \percd{6.6} \\

Math500       
& $20.50 \pm 0.10$
& $26.00 \pm 0.14$ & 1 & \percd{2.9}
& $18.00 \pm 0.10$
& $27.00 \pm 0.16$ & 2 & \percd{6.6} \\
\bottomrule
\end{tabular}
}
 \caption{Performance comparison for LLaMA 3.1 8B and Qwen 2.5 7B over 5 random seeds (mean $\pm$ std) under 0-shot evaluation with a different training set A for optimization from the training portions in Table \ref{tab:robustness2}.  Testing on data set Test portions.}
\label{tab:robustness3}
\end{table}
}

\newpage

\hidden{
Table~\ref{tab:model_parameters} summarizes the total number of parameters for each model, along with the number of layers.  

\begin{itemize}
    \item \textbf{LLaMA 3.1 8B}: 218,112,000 parameters across 32 layers
    \item \textbf{Qwen 2.5 7B}: 233,057,792 parameters across 28 layers
    \item \textbf{Mistral 7B}: 218,112,000 parameters across 32 layers
    \item \textbf{Lucie 7B}: 192,946,176 parameters across 32 layers
    \item \textbf{Qwen 2.5 0.5B}: 14,912,384 parameters across 24 layers
\end{itemize}

\begin{table}[!ht]
\centering
\renewcommand{\arraystretch}{1.3}
\resizebox{0.8\linewidth}{!}{
\begin{tabular}{l|c|c|c|c|c}
\toprule
\textbf{Model} & \textbf{LLaMA 3.1 8B} & \textbf{Qwen 2.5 7B} & \textbf{Mistral 7B} & \textbf{Lucie 7B} & \textbf{Qwen 2.5 0.5B} \\
\midrule
Parameters & 218,112,000 & 233,057,792 & 218,112,000 & 192,946,176 & 14,912,384 \\
\bottomrule
\end{tabular}
}
\caption{Model parameter counts comparison. LLaMA 3.1 8B, Mistral 7B and Lucie 7B has 32 layers, Qwen 2.5 7B has 28 layers and Qwen 2.5 0.5B has 24 layers. }
\label{tab:model_parameters}
\end{table}
}
\newpage

\hidden{
\begin{figure*}[!ht]
    \centering
    \hidden{
    \begin{subfigure}[t]{0.42\textwidth}
        \centering
        \includegraphics[width=\linewidth]{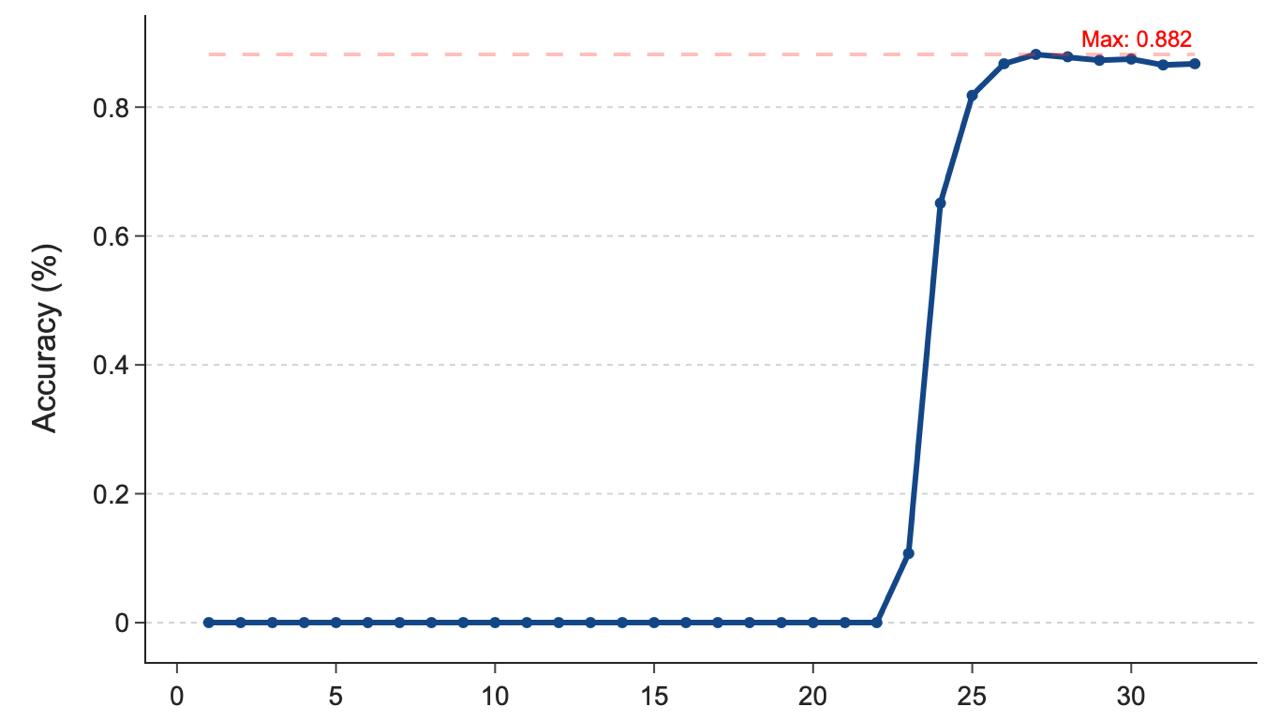}
        \caption{ARC-Easy}
    \end{subfigure}
    \hfill}
    \begin{subfigure}[t]{0.52\textwidth}
        \centering
        \includegraphics[width=\linewidth]{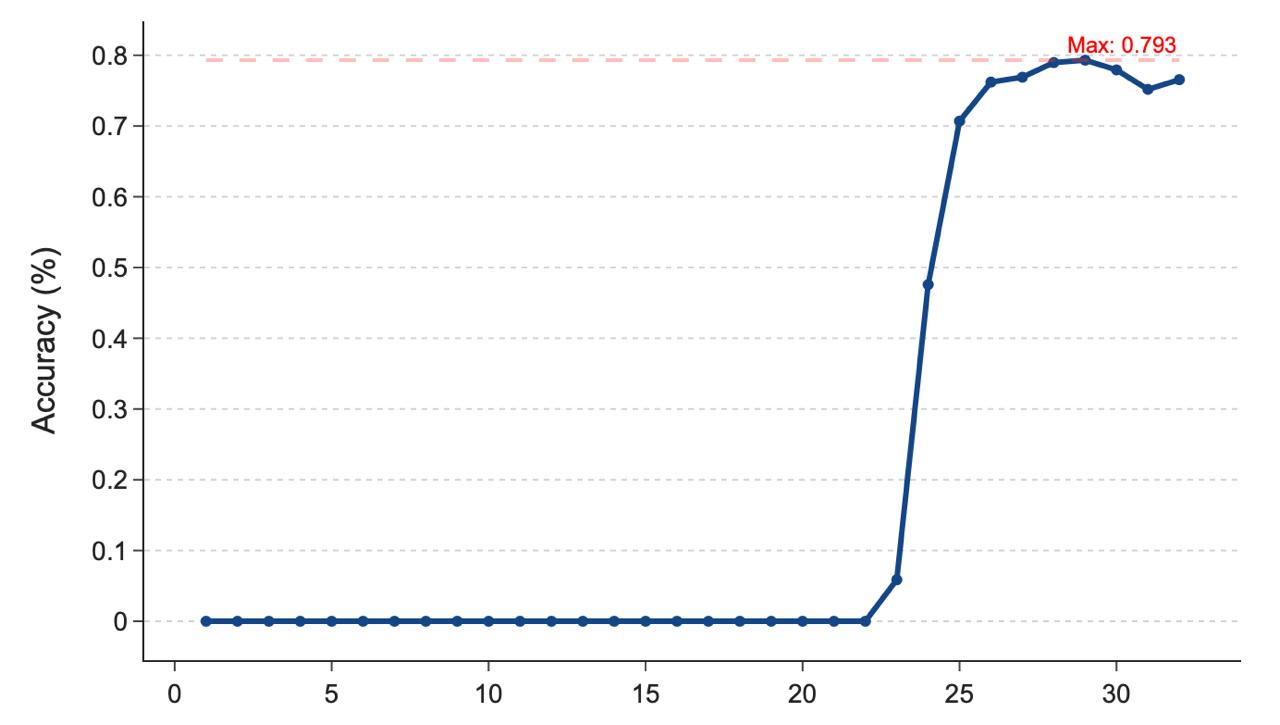}
        \caption{ARC-Challenge}
    \end{subfigure}
    \hfill
    \begin{subfigure}[t]{0.52\textwidth}
        \centering
        \includegraphics[width=\linewidth]{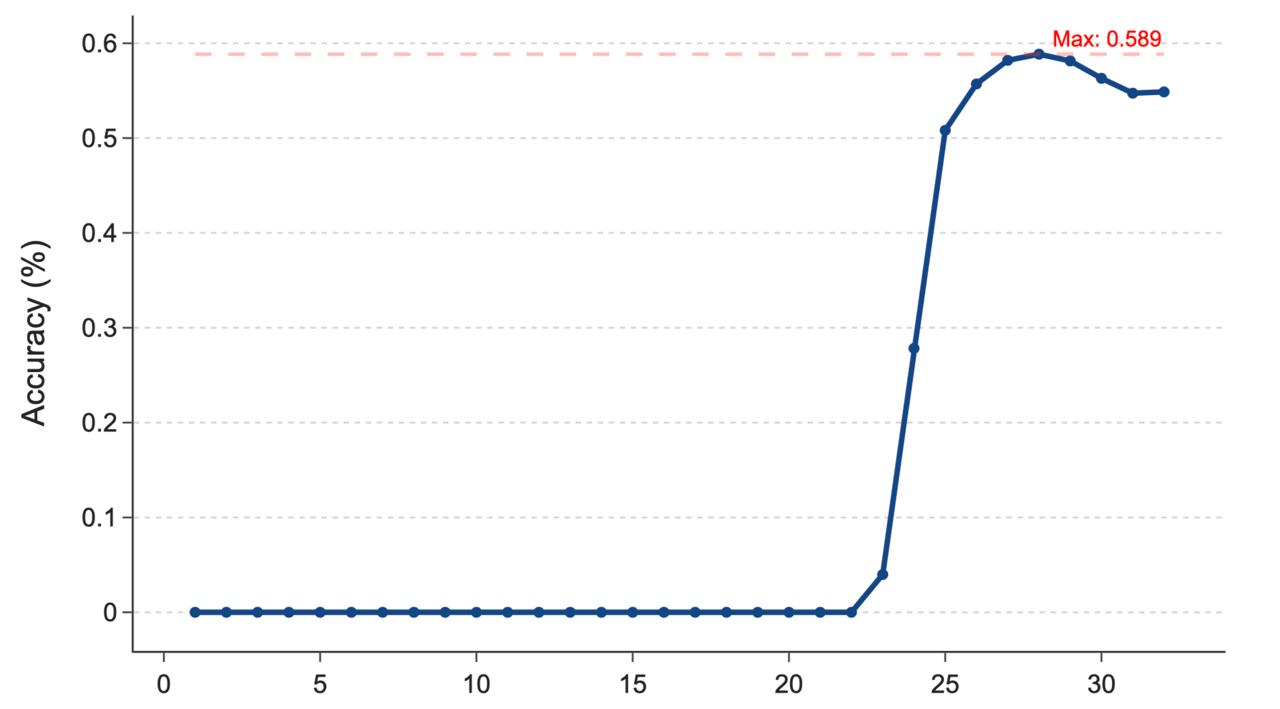}
        \caption{MMLU}
    \end{subfigure}

    \caption{Layer-wise output performance for LLaMA models: results when generating predictions from intermediate layers 1 through 32 on three different datasets.}
    \label{graph:oldexp}
\end{figure*}
}

\newpage

\begin{table*}[!ht]
\centering
\small
\setlength{\tabcolsep}{4pt}
\begin{tabular}{llccc}
\toprule
\textbf{Model} & \textbf{Dataset} & \textbf{Latency (s)} $\downarrow$ & \textbf{Throughput (tok/s)} $\uparrow$ & \textbf{Speedup} \\
\midrule

\multirow{9}{*}{Llama-3.1-8B-Instruct}
& ARC-Easy      & 0.0402 $\rightarrow$ 0.0343 (-14.8\%) & 24.83 $\rightarrow$ 29.13 (+17.3\%) & 1.17$\times$ \\
& ARC-Challenge & 0.0403 $\rightarrow$ 0.0355 (-11.9\%) & 24.78 $\rightarrow$ 28.13 (+13.5\%) & 1.14$\times$ \\
& BoolQ         & 0.0409 $\rightarrow$ 0.0373 (-8.8\%)  & 24.41 $\rightarrow$ 26.76 (+9.6\%)  & 1.10$\times$ \\
& MMLU          & 0.0408 $\rightarrow$ 0.0396 (-2.9\%)  & 24.50 $\rightarrow$ 25.24 (+3.0\%)  & 1.03$\times$ \\
& CommonQA      & 0.0403 $\rightarrow$ 0.0367 (-8.9\%)  & 24.78 $\rightarrow$ 27.21 (+9.8\%)  & 1.10$\times$ \\
& Winogrande    & 0.0404 $\rightarrow$ 0.0356 (-11.9\%) & 24.71 $\rightarrow$ 28.04 (+13.5\%) & 1.13$\times$ \\
& BIG-Bench     & 0.0404 $\rightarrow$ 0.0344 (-15.0\%) & 24.72 $\rightarrow$ 29.08 (+17.6\%) & 1.18$\times$ \\
& GSM8K-Hard    & 0.0406 $\rightarrow$ 0.0394 (-3.0\%)  & 23.87 $\rightarrow$ 24.61 (+3.1\%)  & 1.03$\times$ \\
& Math500       & 0.0407 $\rightarrow$ 0.0383 (-6.0\%)  & 23.85 $\rightarrow$ 25.35 (+6.3\%)  & 1.06$\times$ \\

\midrule

\multirow{9}{*}{Qwen-2.5-7B-Instruct}
& ARC-Easy      & 0.0356 $\rightarrow$ 0.0320 (-10.2\%) & 28.03 $\rightarrow$ 31.21 (+11.3\%) & 1.11$\times$ \\
& ARC-Challenge & 0.0357 $\rightarrow$ 0.0333 (-6.8\%)  & 27.99 $\rightarrow$ 30.02 (+7.3\%)  & 1.07$\times$ \\
& BoolQ         & 0.0366 $\rightarrow$ 0.0316 (-13.5\%) & 27.31 $\rightarrow$ 31.57 (+15.6\%) & 1.16$\times$ \\
& MMLU          & 0.0359 $\rightarrow$ 0.0298 (-17.0\%) & 27.80 $\rightarrow$ 33.46 (+20.4\%) & 1.20$\times$ \\
& CommonQA      & 0.0354 $\rightarrow$ 0.0330 (-6.7\%)  & 28.24 $\rightarrow$ 30.25 (+7.1\%)  & 1.07$\times$ \\
& Winogrande    & 0.0354 $\rightarrow$ 0.0318 (-10.2\%) & 28.21 $\rightarrow$ 31.39 (+11.3\%) & 1.11$\times$ \\
& BIG-Bench     & 0.0357 $\rightarrow$ 0.0284 (-20.4\%) & 27.94 $\rightarrow$ 35.40 (+26.7\%) & 1.27$\times$ \\
& GSM8K-Hard    & 0.0361 $\rightarrow$ 0.0337 (-6.6\%)  & 25.88 $\rightarrow$ 27.73 (+7.1\%)  & 1.07$\times$ \\
& Math500       & 0.0361 $\rightarrow$ 0.0337 (-6.7\%)  & 26.32 $\rightarrow$ 28.20 (+7.1\%)  & 1.07$\times$ \\

\midrule

\multirow{9}{*}{Lucie-7B-Instruct-v1.1}
& ARC-Easy      & 0.0399 $\rightarrow$ 0.0327 (-18.1\%) & 25.03 $\rightarrow$ 30.54 (+22.0\%) & 1.22$\times$ \\
& ARC-Challenge & 0.0398 $\rightarrow$ 0.0314 (-21.0\%) & 25.11 $\rightarrow$ 31.79 (+26.6\%) & 1.27$\times$ \\
& BoolQ         & 0.0402 $\rightarrow$ 0.0341 (-15.1\%) & 24.87 $\rightarrow$ 29.27 (+17.7\%) & 1.18$\times$ \\
& MMLU          & 0.0401 $\rightarrow$ 0.0304 (-24.1\%) & 24.89 $\rightarrow$ 32.79 (+31.7\%) & 1.32$\times$ \\
& CommonQA      & 0.0400 $\rightarrow$ 0.0364 (-9.1\%)  & 24.96 $\rightarrow$ 27.46 (+10.0\%) & 1.10$\times$ \\
& Winogrande    & 0.0400 $\rightarrow$ 0.0339 (-15.1\%) & 25.00 $\rightarrow$ 29.43 (+17.7\%) & 1.18$\times$ \\
& BIG-Bench     & 0.0402 $\rightarrow$ 0.0292 (-27.3\%) & 24.86 $\rightarrow$ 34.17 (+37.4\%) & 1.37$\times$ \\
& GSM8K-Hard    & 0.0399 $\rightarrow$ 0.0388 (-2.9\%)  & 24.60 $\rightarrow$ 25.33 (+3.0\%)  & 1.03$\times$ \\
& Math500       & 0.0402 $\rightarrow$ 0.0378 (-6.0\%)  & 24.46 $\rightarrow$ 26.00 (+6.3\%)  & 1.06$\times$ \\

\midrule

\multirow{9}{*}{Mistral-7B-Instruct-v0.3}
& ARC-Easy      & 0.0403 $\rightarrow$ 0.0342 (-15.1\%) & 24.78 $\rightarrow$ 29.19 (+17.8\%) & 1.18$\times$ \\
& ARC-Challenge & 0.0403 $\rightarrow$ 0.0330 (-18.2\%) & 24.77 $\rightarrow$ 30.29 (+22.2\%) & 1.22$\times$ \\
& BoolQ         & 0.0407 $\rightarrow$ 0.0334 (-18.1\%) & 24.51 $\rightarrow$ 29.91 (+22.0\%) & 1.22$\times$ \\
& MMLU          & 0.0407 $\rightarrow$ 0.0382 (-6.0\%)  & 24.56 $\rightarrow$ 26.12 (+6.3\%)  & 1.06$\times$ \\
& CommonQA      & 0.0405 $\rightarrow$ 0.0356 (-12.1\%) & 24.68 $\rightarrow$ 28.07 (+13.8\%) & 1.14$\times$ \\
& Winogrande    & 0.0404 $\rightarrow$ 0.0282 (-30.2\%) & 24.71 $\rightarrow$ 35.37 (+43.1\%) & 1.43$\times$ \\
& BIG-Bench     & 0.0400 $\rightarrow$ 0.0289 (-27.6\%) & 24.99 $\rightarrow$ 34.69 (+38.8\%) & 1.39$\times$ \\
& GSM8K-Hard    & 0.0401 $\rightarrow$ 0.0377 (-6.0\%)  & 24.60 $\rightarrow$ 26.14 (+6.2\%)  & 1.06$\times$ \\
& Math500       & 0.0405 $\rightarrow$ 0.0392 (-3.2\%)  & 24.36 $\rightarrow$ 25.17 (+3.3\%)  & 1.03$\times$ \\

\bottomrule
\end{tabular}
\caption{Latency and throughput comparison between baseline and TALE-pruned (BEST) models across tasks. TALE consistently reduces first-token latency and improves throughput.}
\label{tab:latency_throughput}
\end{table*}

\section{Practical computing savings and scaling}
\label{appendix:computation}
We quantify TALE's inference-cost reduction by measuring TFLOPs (tera-FLOPs) drop per removed layer. Across models and tasks, removing a single transformer layer yields a mean TFLOPs reduction of $3.00\%\pm 0.20\%$. Because TALE removes entire layers sequentially, the total TFLOPs reduction scales essentially linearly with the number of iterations (layers removed). In practice this means only a few iterations are required to reach common sparsity targets: e.g., three iterations remove roughly $\approx$ 9\% TFLOPs, sufficient to realize $\approx$ 10\% sparsity in our settings.

\paragraph{Wall-clock runtime and end-to-end cost.}
In addition to FLOPs-based estimates, we report wall-clock runtime measurements under the exact decoding settings used for evaluation. TALE incurs a one-time optimization cost per task, after which the pruned model is used for standard inference with reduced depth.

For all non-reasoning benchmarks (e.g., ARC, BoolQ, MMLU, CommonsenseQA), we use greedy decoding with \texttt{max\_new\_tokens = 1}, reflecting single-step answer generation. Under this setting, the total wall-clock time required to complete a full TALE optimization run is approximately 1 hour per dataset on a single NVIDIA A100 GPU.

For reasoning-heavy benchmarks (GSM8K-Hard and Math500), we use \texttt{max\_new\_tokens = 200} to allow full chain-of-thought generation. Due to longer decoding sequences, the average wall-clock time for TALE on these datasets is approximately 3 hours, measured over the evaluated subset.

Importantly, this optimization cost is incurred only once per task. During deployment, inference is performed using the pruned model, yielding reduced per-example latency and throughput improvements proportional to the number of eliminated layers. These gains are amortized over all subsequent inference calls.

\newpage
\section{Layer Redundancy Is Not a Multi-Task Artifact}
\subsection{Training details}

\begin{figure*}[!ht]
\centering
\includegraphics[width=0.8\linewidth]{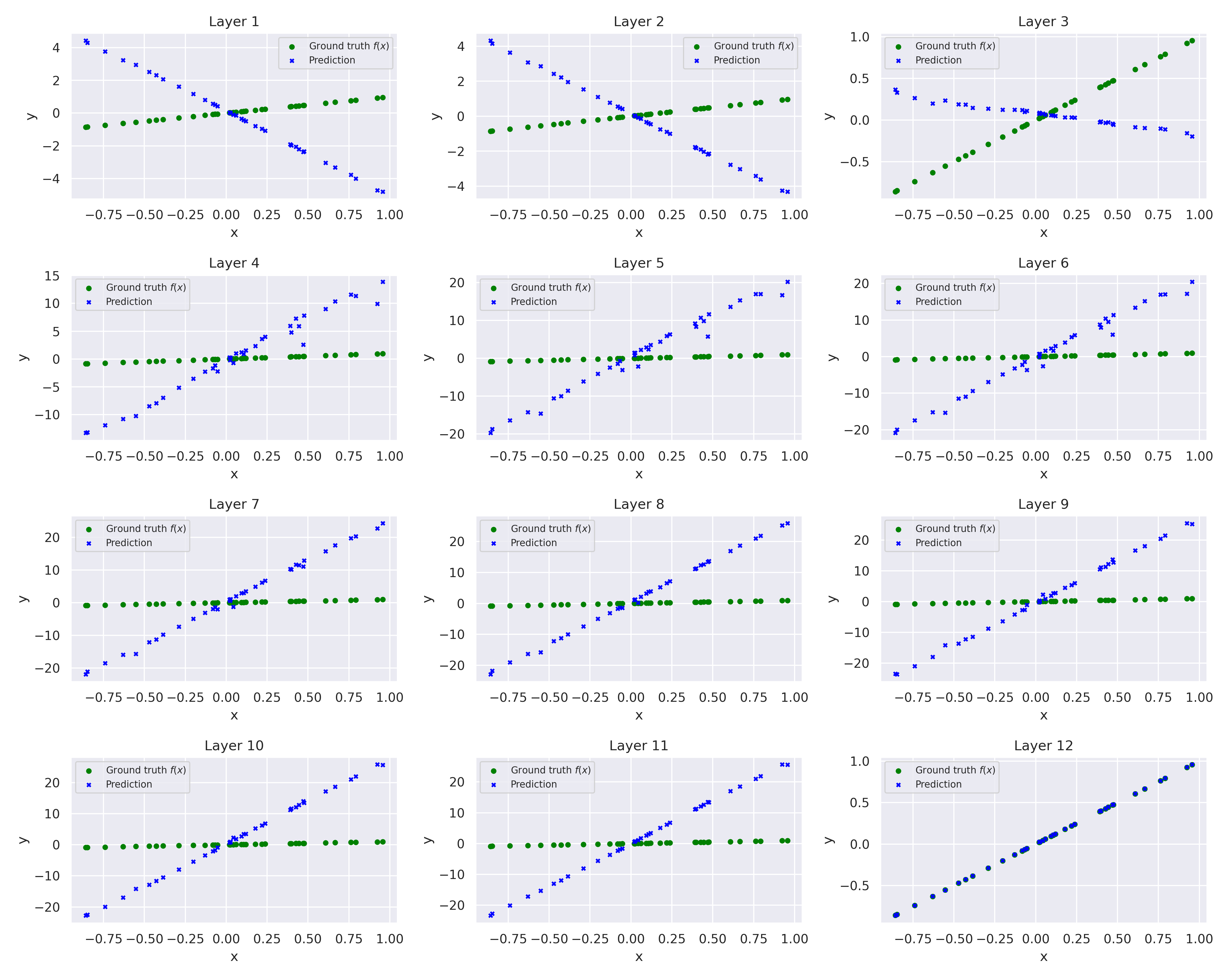}
\caption{Plots showing evolution of the predictions over layers for $f(x)=x$ for a model trained on degree 1.}\label{evolution}
\end{figure*}

We train a transformer model to perform in-context learning over the class of linear functions. 
Given a prompt consisting of a small set of input--output examples $(x_1, g(x_1), \dots, x_p, g(x_p), x),$
the model is tasked with predicting the value \( g(x) \) for the final input \(x\).
We follow the training and evaluation setup of \cite{naim2025analyzing, naim2025re}

\hidden{
Our architecture is a 12-layer transformer with 8 attention heads, trained \textbf{from scratch} exclusively on the linear function family. 
The training objective follows the standard ICL formulation introduced in \cite{garg:etal:2022, naim2025re}:

\begin{equation}
\label{eq:autoregressive}
\begin{aligned}
\min_{\theta}\; & \mathbb{E}_{g \sim \mathcal{D}_\mathcal{F}}\Bigg[
    \mathbb{E}_{x_1, \dots, x_p \sim \mathcal{D}_\mathcal{I}}
    \sum_{i=0}^k 
    \ell\!\Big(
        y_{i+1}, \\
& \qquad f^{\theta}\!\big(x_1, g(x_1), \dots, x_{i+1}\big)
    \Big)
\Bigg]
\end{aligned}
\end{equation}

where \(\ell\) is the squared error loss. 
Here, \(y_{i+1} = g(x_{i+1})\) denotes the ground-truth output of the underlying linear function \(g\). 

}
\subsection{Plots}
Figure \ref{evolution}

\newpage

\section{LM-Eval results}
\label{appendix:lmeval}




\begin{table}[!ht]
\centering
\renewcommand{\arraystretch}{1.3}
\resizebox{\linewidth}{!}{
\begin{tabular}{l|c|ccc|ccc}
\toprule
\multirow{3}{*}{\textbf{Dataset}} & 
\multicolumn{7}{c}{\textbf{LLaMA 3.1 8B 0-shot}} \\
\cmidrule(lr){2-8}
 & \multicolumn{1}{c|}{Baseline} & \multicolumn{3}{c|}{Best Model} & \multicolumn{3}{c}{BSBA} \\
\cmidrule(lr){2-2} \cmidrule(lr){3-5} \cmidrule(lr){6-8}
& Perf. & Perf. & \#D & Sp. & Perf. & \#D & Sp. \\
\midrule
BoolQ         &  82.4 & 87.63 & 3 & \percd{8.8} & 85.62 &  7 & \percd{20.5} \\
Hellaswag     &  52.5 & 55.5  & 3 & \percd{8.8} & 54.5 & 5 &  \percd{14.6} \\
COMMONQA      &  77.2 & 81.61 & 6 & \percd{17.6} & 80.27 & 7 & \percd{20.5} \\
WINOGRANDE    & 75.92 & 78.93 & 4 & \percd{11.7} & 76.59 & 5 &  \percd{14.6}\\
GSM8k         & 43 & 58.5 & 2 & \percd{6.0}  & 58.5 & 2 & \percd{6.0} \\
\bottomrule
\end{tabular}}
\caption{Results of \textbf{LLaMA 3.1 8B} across benchmarks. All tested on 0-shot and evaluated with LM Eval.}
\label{tab:lmeval}
\end{table}


\section{Few-shot Learning Results}

\begin{table}[!ht]
\centering
\renewcommand{\arraystretch}{1.3}
\resizebox{0.8\linewidth}{!}{
\begin{tabular}{l|c|ccc|ccc}
\toprule
\multirow{3}{*}{\textbf{Dataset}} & 
\multicolumn{7}{c}{\textbf{Lucie 7B few-shots}} \\
\cmidrule(lr){2-8}
 & \multicolumn{1}{c|}{Baseline} & \multicolumn{3}{c|}{Best Model} & \multicolumn{3}{c}{BSBA} \\
\cmidrule(lr){2-2} \cmidrule(lr){3-5} \cmidrule(lr){6-8}
& Perf. & Perf. & \#D & Sp. & Perf. & \#D & Sp. \\
\midrule
ARC-Easy      & 69.2 & 72.36 & 9  & 1.41 & 71.27 & 12  & 1.68 \\
ARC-Challenge & 49.31& 55,17 & 9 & 1.39 & 51.72 &  13 & 1.67 \\
BoolQ         &  77.6 & 79.10  &  6 & 1.22 & 78.5 & 10 & 1.27  \\
MMLU          & 41.02  &  43.44  & 7  & 1.26 & 41.48  & 11 & 1.55 \\
COMMONQA      &  55.4  & 69.7 & 3 & 1.22 & 57.10 & 17 & 2.02  \\
WINOGRANDE    &  52.8 & 56.90  & 12 & 1.58 & 53.30  & 17  & 1.74 \\
BIG-Bench     &  68.8  & 77.20  &  9 & 1.61 & 72 & 15 & 2.23  \\
GSM8K-HARD    & 26.97   & 29.21 & 1 & 1.03 & 26.97 & 2 & 1.1 \\
\bottomrule
\end{tabular}}
\caption{Results of \textbf{Lucie 7B} across nine benchmarks. All tested on 5-shots, except gms8k on 8-shots.}
\label{tab:lucie_fewshots}
\end{table}

\begin{table}[!ht]
\centering
\renewcommand{\arraystretch}{1.3}
\resizebox{0.8\linewidth}{!}{
\begin{tabular}{l|c|ccc|ccc}
\toprule
\multirow{3}{*}{\textbf{Dataset}} & 
\multicolumn{7}{c}{\textbf{LLaMA 3.1 8B few-shots}} \\
\cmidrule(lr){2-8}
 & \multicolumn{1}{c|}{Baseline} & \multicolumn{3}{c|}{Best Model} & \multicolumn{3}{c}{BSBA} \\
\cmidrule(lr){2-2} \cmidrule(lr){3-5} \cmidrule(lr){6-8}
& Perf. & Perf. & \#D & Sp. & Perf. & \#D & Sp. \\
\midrule
ARC-Easy      & 90.36  & \textbf{92.18} & 4 & 1.14 & 90.91 & 8 & 1.37 \\
ARC-Challenge & 78.2 & \textbf{83.10}  & 3 & 1.17 & 78.62  & 9   & 1.42 \\
BoolQ         &  82.7 & 85.3  & 4 & 1.11 & 83.0 &  6 & 1.22 \\
MMLU          &  59.2 & 62.38  & 4 & 1.14 & 59.57 & 7 & 1.26 \\
COMMONQA      &  73.30 & 75.30 & 6 & 1.22 & 73.80 & 7 & 1.32 \\
WINOGRANDE    & 57.01 & 60.1 & 3 & 1.1 & 57.02 & 8 & 1.3 \\
BIG-Bench     & 70.0 & 83.60 & 5 & 1.2 & 81.20 & 15 & 1.83 \\
GSM8K-HARD    & 60.67 & 60.67 & 0 & 1 & 60.67 & 0 & 1 \\
MATH500        & 44.00 & 49.00  & 1 & 1.02 & 45.00 & 2 & 1.03 \\
\bottomrule
\end{tabular}
}
\caption{Results of \textbf{LLaMA 3.1 8B} across nine benchmarks. All tested on 5-shots, except gms8k and MATH500 on 8-shots 
}
\label{tab:llamafewshots_results_colors}
\end{table}

\newpage

\section{Information theory:
Why pruned models might perform better. 
}
\label{sec:MI}


Our results pose a puzzle: the increase in accuracy with \method
is counterintuitive: why would removing parts of a carefully trained model lead to better performance? One way to explore this question is mutual information.

\cite{alemi2016deep} use information theory  
to analyze how neural networks learn and represent data.  \citet{fano1961transmission} defines $\text{I}(\text{X};\text{Y})$, the mutual information between two random variables $X$ and $Y$, with the equation: 
\begin{equation}\label{mutual}
\begin{aligned}
\text{I}(\text{X};\text{Y})
&= \text{H}(\text{Y}) - \text{H}(\text{Y} \mid \text{X}) \\
&= \text{H}(\text{X}) - \text{H}(\text{X} \mid \text{Y}) \\
&= \sum_{x \in \mathcal{X}} \sum_{y \in \mathcal{Y}}
p(x,y) \log \frac{p(x,y)}{p(x)\,p(y)}
\end{aligned}
\end{equation}
 where $p(x,y)$ is the joint distribution of $\text{X}$ and $\text{Y}$, and $p(x), p(y)$ are their marginals and where $\text{H}(\text{X}) = -\sum_x p(x)\log p(x)$ is the \cite{shannon1948mathematical} entropy. $\text{I}(\text{X};\text{Y})$ measures how much knowing $\text{X}$ reduces uncertainty about $\text{Y}$.  To attempt to explain why accuracy increases through task pruning we also use MI.

A major challenge of this approach is that it requires information about true distributions, which are infeasible to compute. As a result, researchers typically assume a Gaussian distribution \cite{gabrie2019entropy, gao2015efficient, wen2024gaussian} or approximate the probe using a classifier \cite{belinkov2022probing, alain2016understanding} or an MLP \cite{belghazi2018mine}. These approximations can yield useful insights. In our case, the Gaussian assumption did not fit our datasets. Since we evaluate on QA tasks, we used a trainable classifier to approximate the probes and estimate $I(X^\ell, \text{Y})$ at each layer, where $X^\ell$ denotes the contextualized representations at layer $\ell$ and Y denotes the target answer. This approximates how much information the layer $\ell$ representations contain about the answer. The goal is then to examine whether some layers exhibit a sharp drop in information and whether those layers coincide with the ones whose removal leads to improved performance.

\begin{figure}[!ht]
\centering
\begin{subfigure}[t]{0.30\textwidth}
\centering
\includegraphics[width=\linewidth]{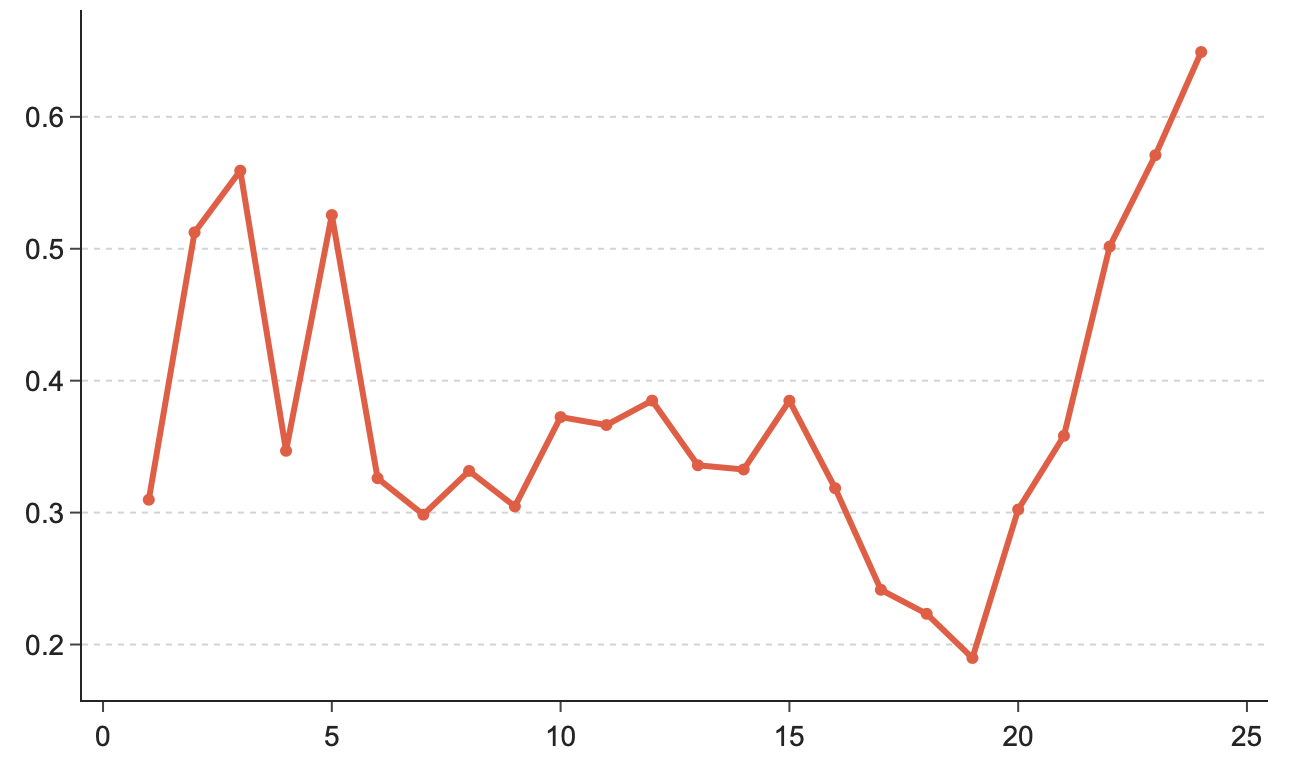}
\caption{ARC-Easy (Qwen 0.5B)}
\label{fig:arc_easy}
\end{subfigure}
\hfill
\begin{subfigure}[t]{0.30\textwidth}
\centering
\includegraphics[width=\linewidth]{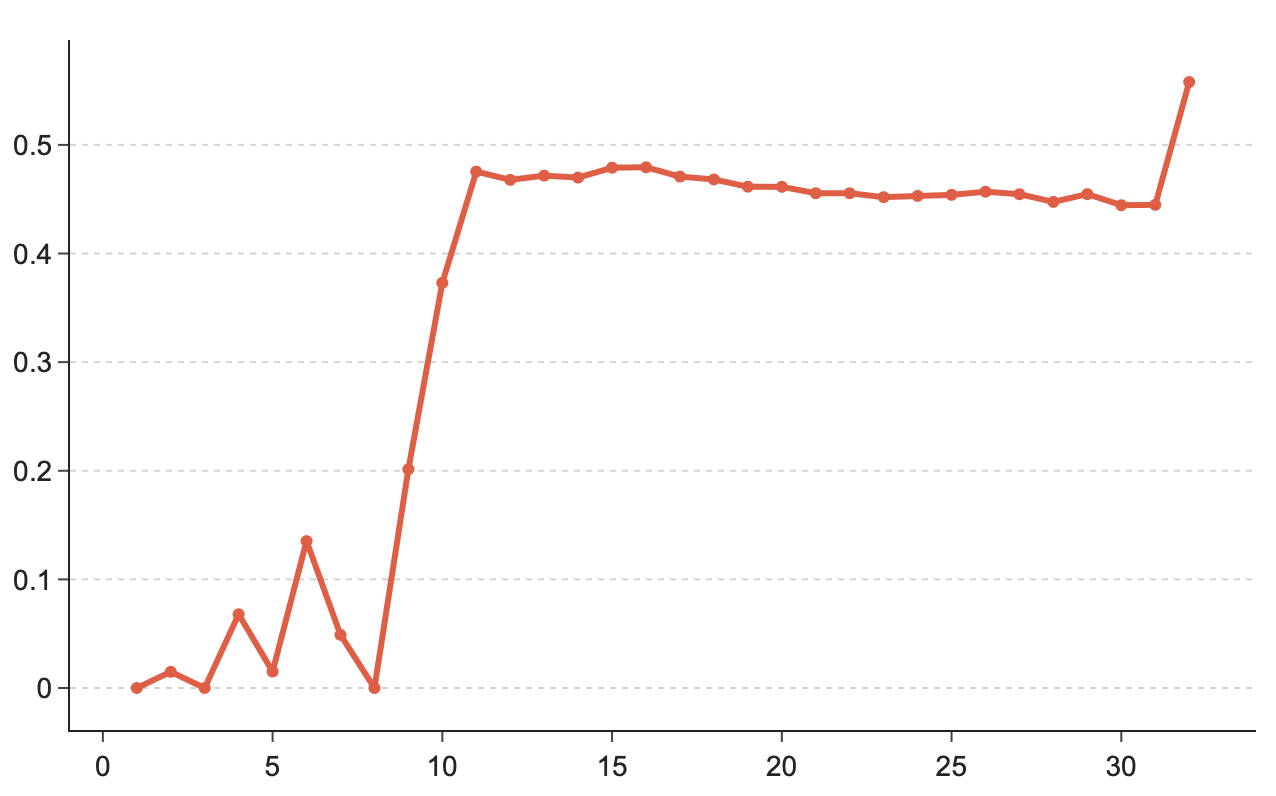}
\caption{BoolQ (Lucie 7B)}
\label{fig:boolq}
\end{subfigure}
\hfill
\begin{subfigure}[t]{0.30\textwidth}
\centering
\includegraphics[width=\linewidth] {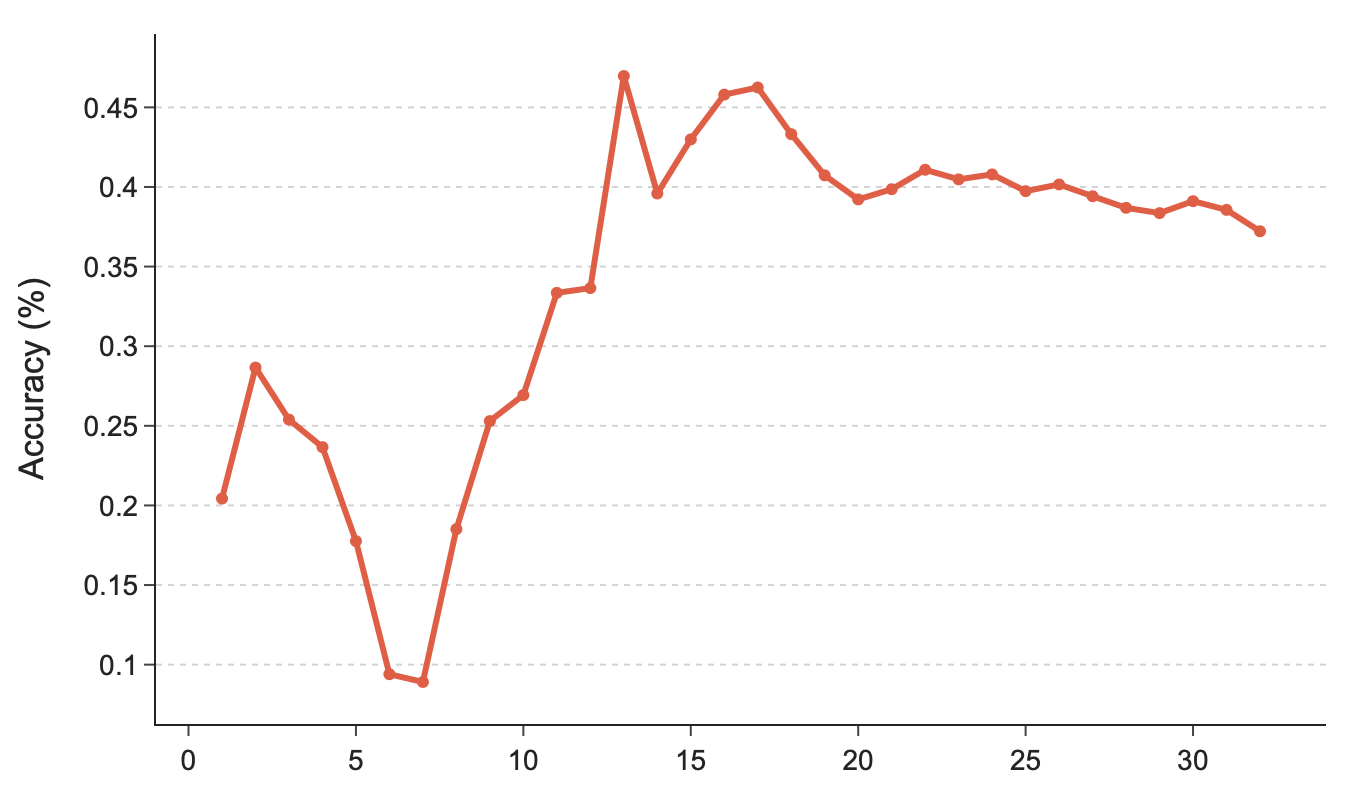}
\caption{BigBench (Llama 8B)}
\label{fig:bigbench}
\end{subfigure}

\caption{Evolution of mutual information (MI) across transformer layers for different benchmark datasets and different models. Each subplot shows how information is processed and transformed as it flows through the network layers, demonstrating distinct patterns of information propagation for (a) ARC-Easy on Qwen 0.5B, (b) BoolQ on Lucie 7B, and (c) BigBench on LLaMA 8B.}
\label{fig:datasets_comparison}
\end{figure}

Our findings, summarized in Figure~\ref{fig:datasets_comparison} and Table~\ref{deletedllama}, reveal two key patterns:
(i) several layers in large pre-trained transformers exhibit a pronounced drop in mutual information;
(ii) removing layers dictated by \method consistently increases the mutual information at the subsequent layer across tasks.
Together, these results suggest that certain layers act more as bottlenecks than as contributors to task-relevant representations, providing a rationale for why pruning can lead to improved accuracy.

\begin{figure}[!ht]
    \centering
    \includegraphics[width=\linewidth]{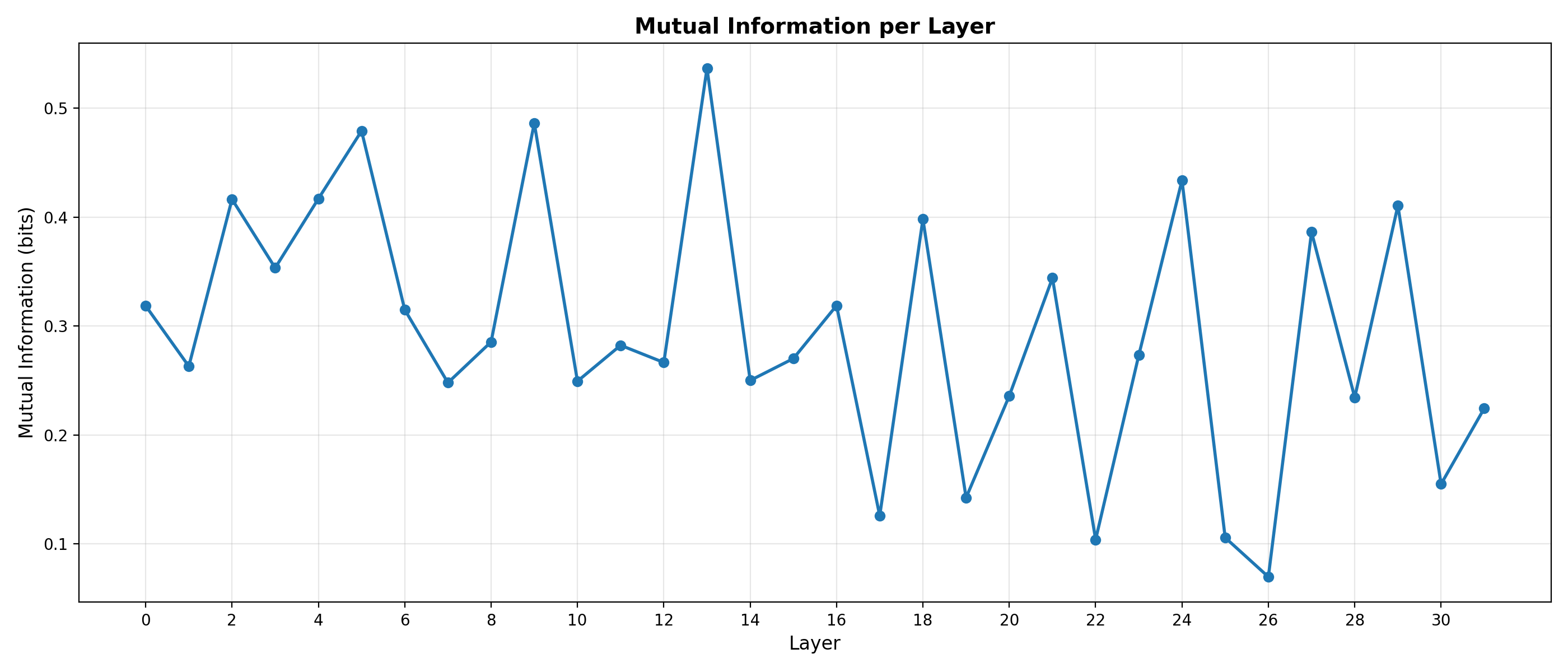}
    \caption{Evolution of Mutual Information about the output through layers for Llama3-8B.}
    \label{MIllama}
\end{figure}

\hidden{
{\color{orange}To apply these notions in our case, we estimate $\text{I}(\text{X}^{(l)}, \text{Y})$ at each layer using trained probes as mutual information approximators. While this approach yields $\text{I}_{\text{probe}}(\text{X}^{(l)}, \text{Y})$, an approximation of the true mutual information, probe-based estimates capture the linearly accessible task-relevant information at each layer, which is meaningful since many downstream tasks employ linear classification heads \cite{belinkov2022probing}.

This estimate together with these three observations yields an information-theoretic explanation for why architectural pruning can improve model performance. (i) Certain layers in large pre-trained transformers decrease mutual information between the layer's representation and the target task (Figure \ref{fig:datasets_comparison} and Table \ref{deletedllama}). (ii) 
From \method we select the layer $\ell$ removed at the first iteration, and compare 
$\mathrm{I}\!\left(X^{(\ell+1)}, Y\right)$ before deletion of layer $\ell$ to its value after deletion. Deleting this layer always increases the mutual information at the subsequent layer on the tasks, 
effectively preventing the previous layer, which acts as noise, from obstructing the flow of information through the network.  
Although our findings depend on the approximative nature of probe-based MI estimation, 
they provide evidence that certain layers in over-parameterized transformers act as information bottlenecks, which degrade rather than refine task-relevant representations. By removing layers that decrease mutual information with the target task, we enable  representations with higher task-relevant information to flow directly to subsequent layers using the residual connection.  This improves the information flow through the remaining architecture, results in representations that are more predictive of the target task, and yields improved accuracy. }

}

\section{\method Dynamics}
\label{appendix:dynamicss}
\begin{figure*}[!ht]
    \centering
    \begin{subfigure}[t]{0.32\textwidth}
        \centering
        \includegraphics[width=\linewidth]{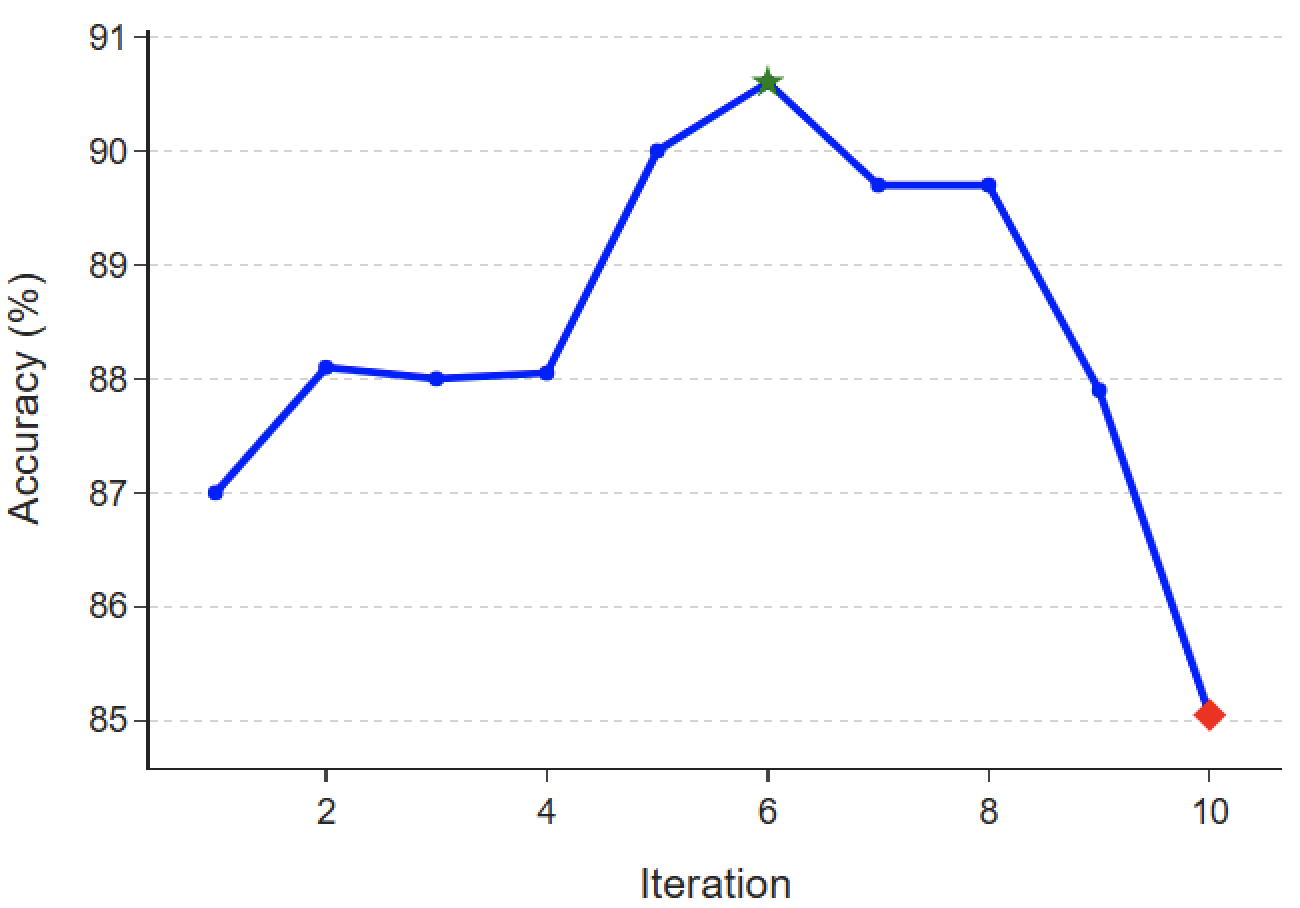}
        \caption{ARC-Easy}
    \end{subfigure}
    \hfill
    \begin{subfigure}[t]{0.32\textwidth}
        \centering
        \includegraphics[width=\linewidth]{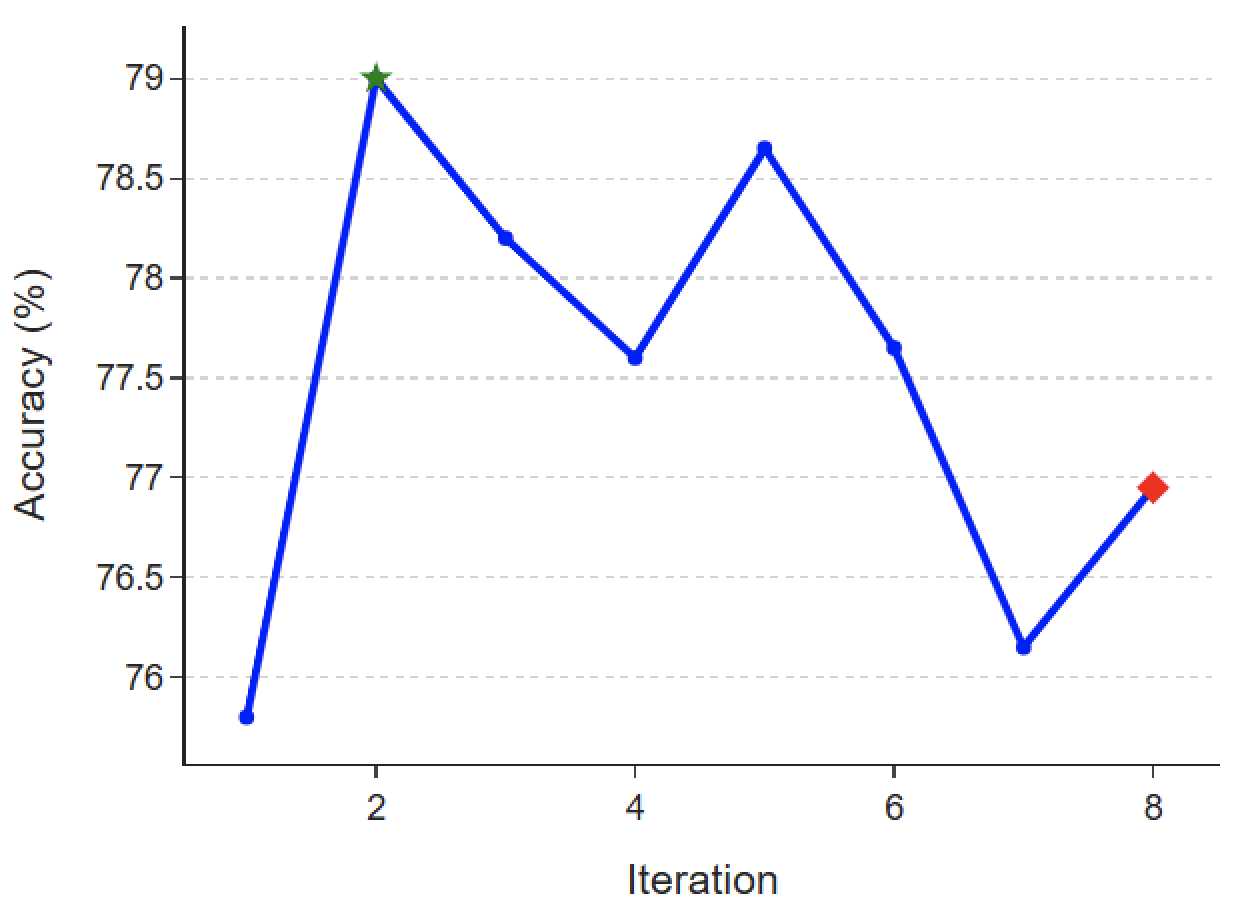}
        \caption{ARC-Challenge}
    \end{subfigure}
    \hfill
    \begin{subfigure}[t]{0.32\textwidth}
        \centering
        \includegraphics[width=\linewidth]{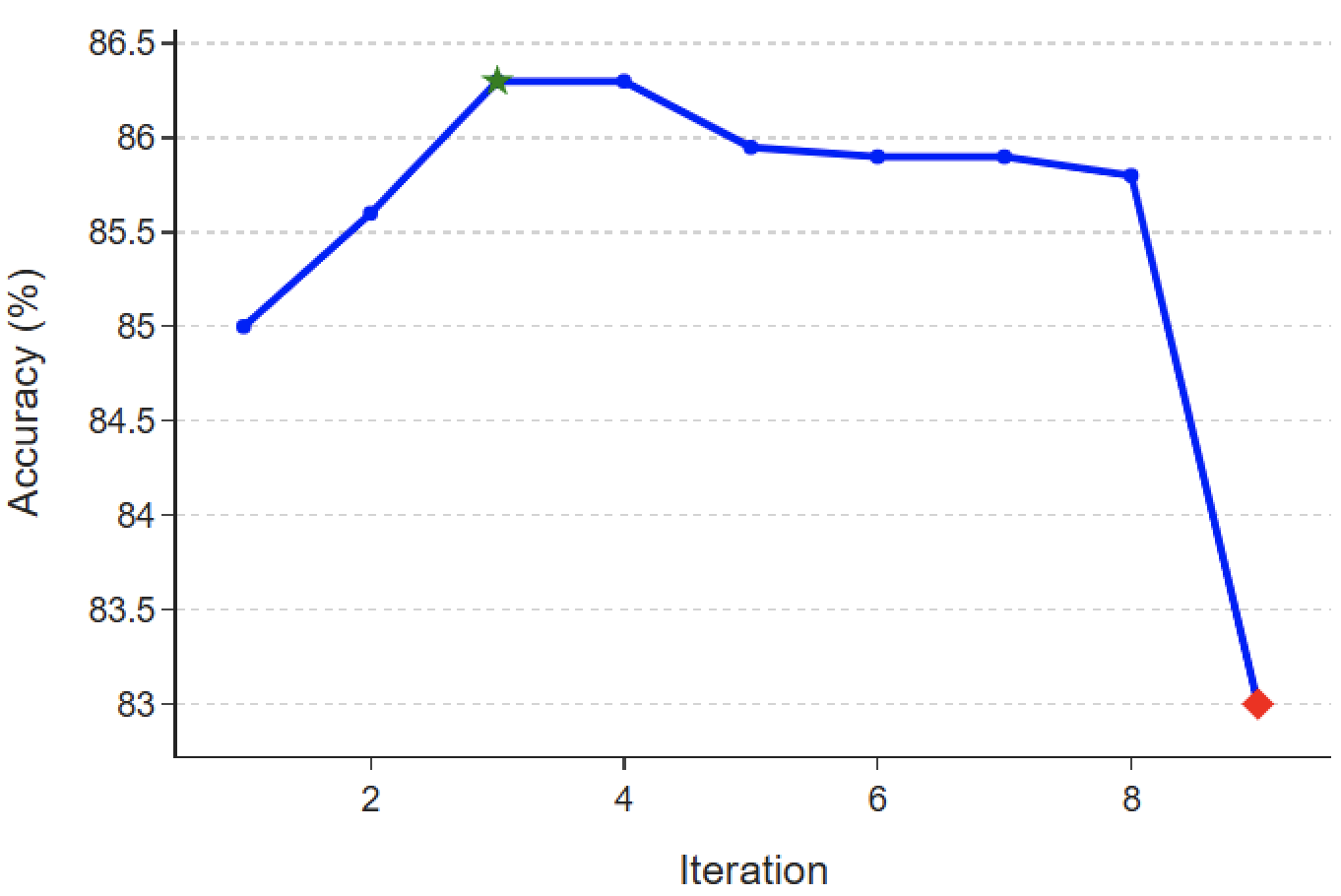}
        \caption{BoolQ}
    \end{subfigure}

    \vspace{0.3cm}
    \begin{subfigure}[t]{0.32\textwidth}
        \centering
        \includegraphics[width=\linewidth]{img/plots/mmlu.png}
        \caption{MMLU}
    \end{subfigure}
    \hfill
    \begin{subfigure}[t]{0.32\textwidth}
        \centering
        \includegraphics[width=\linewidth]{img/plots/commonqa.png}
        \caption{CommonQA}
    \end{subfigure}
    \hfill
    \begin{subfigure}[t]{0.32\textwidth}
        \centering
        \includegraphics[width=\linewidth]{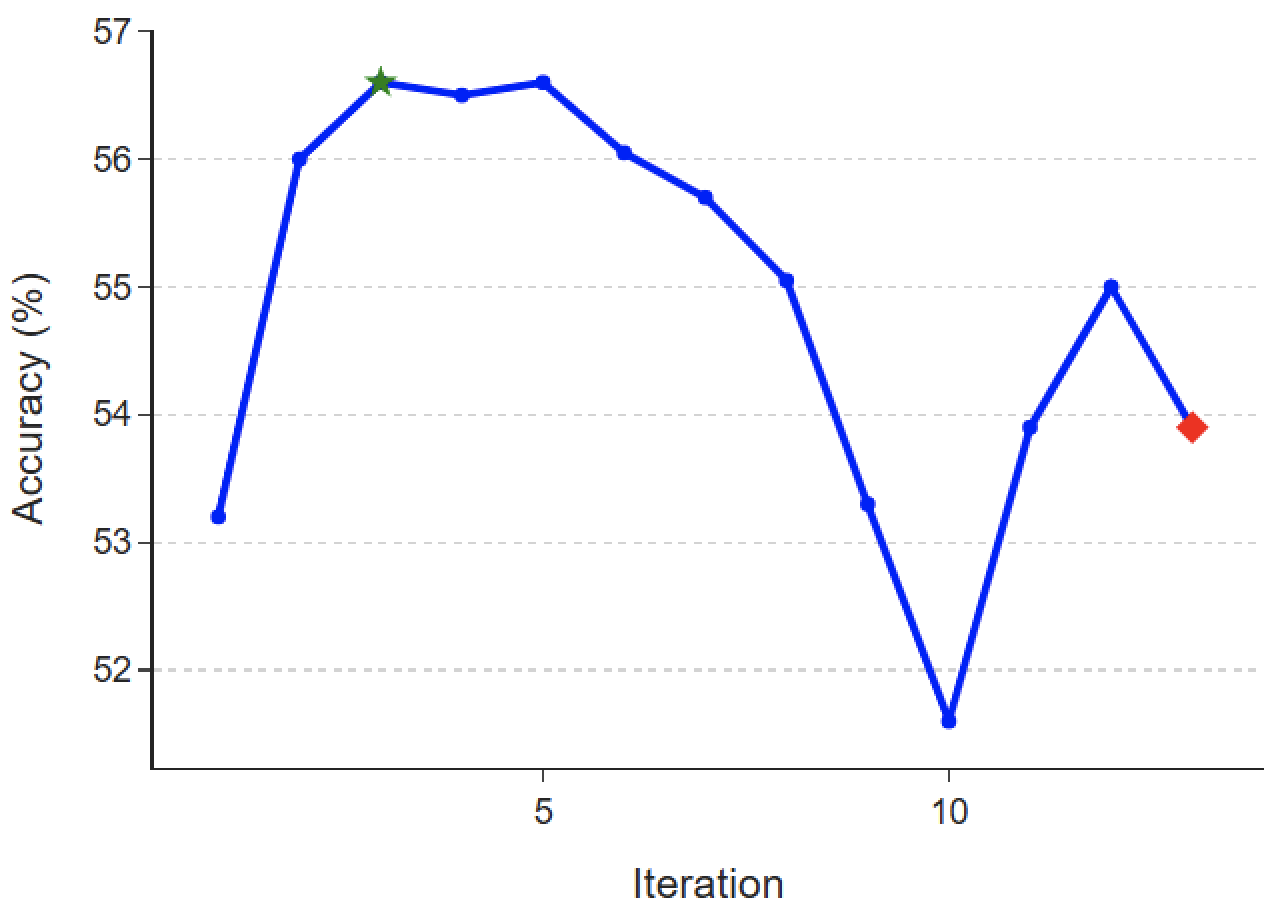}
        \caption{WinoGrande}
    \end{subfigure}

    \vspace{0.3cm}
    \begin{subfigure}[t]{0.32\textwidth}
        \centering
        \includegraphics[width=\linewidth]{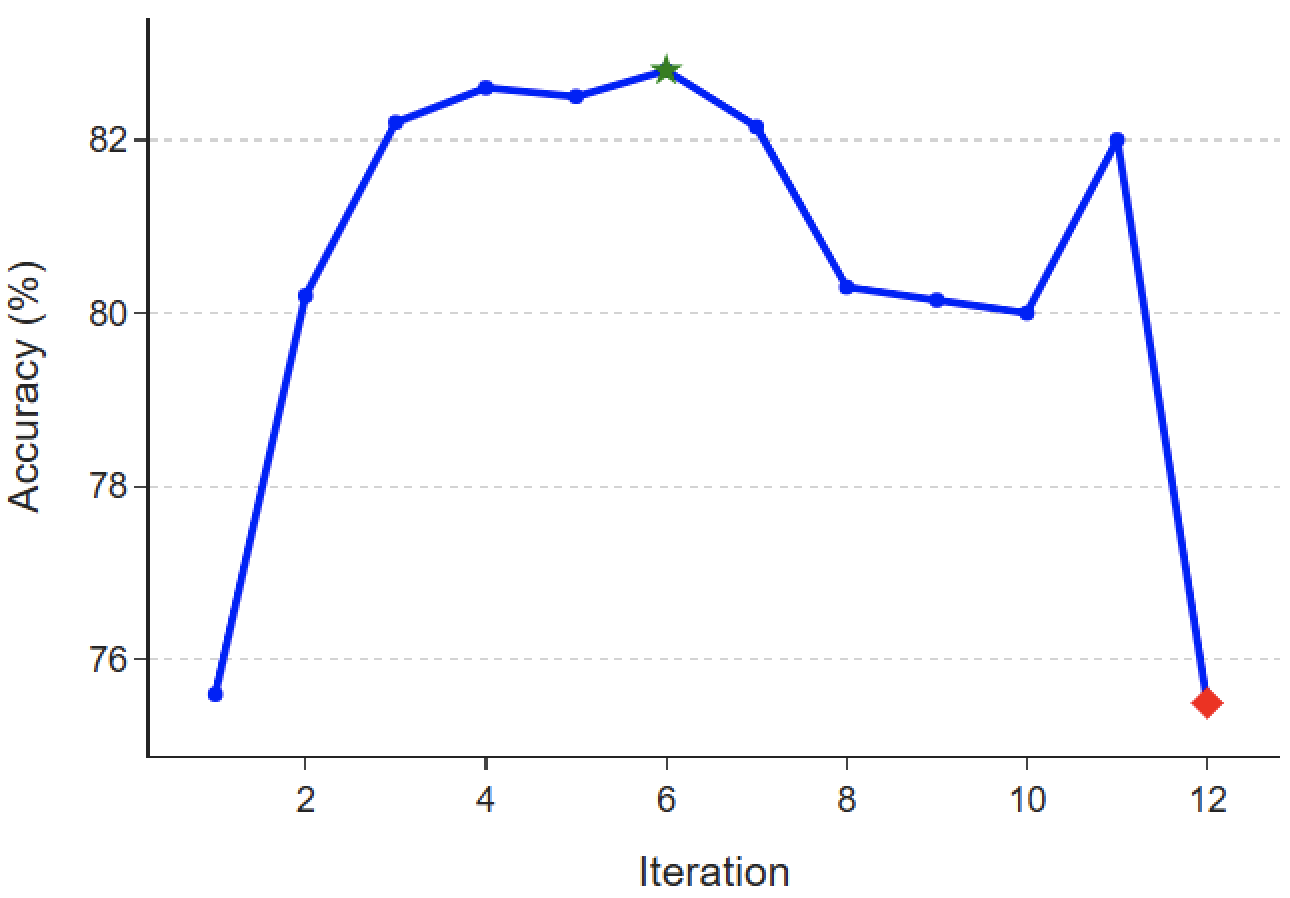}
        \caption{BIG-Bench}
    \end{subfigure}
    \hfill
    \begin{subfigure}[t]{0.32\textwidth}
        \centering
        \includegraphics[width=\linewidth]{img/plots/gsm8k_hard.png}
        \caption{GSM8K-Hard}
    \end{subfigure}
    \hfill
    \begin{subfigure}[t]{0.32\textwidth}
        \centering
        \includegraphics[width=\linewidth]{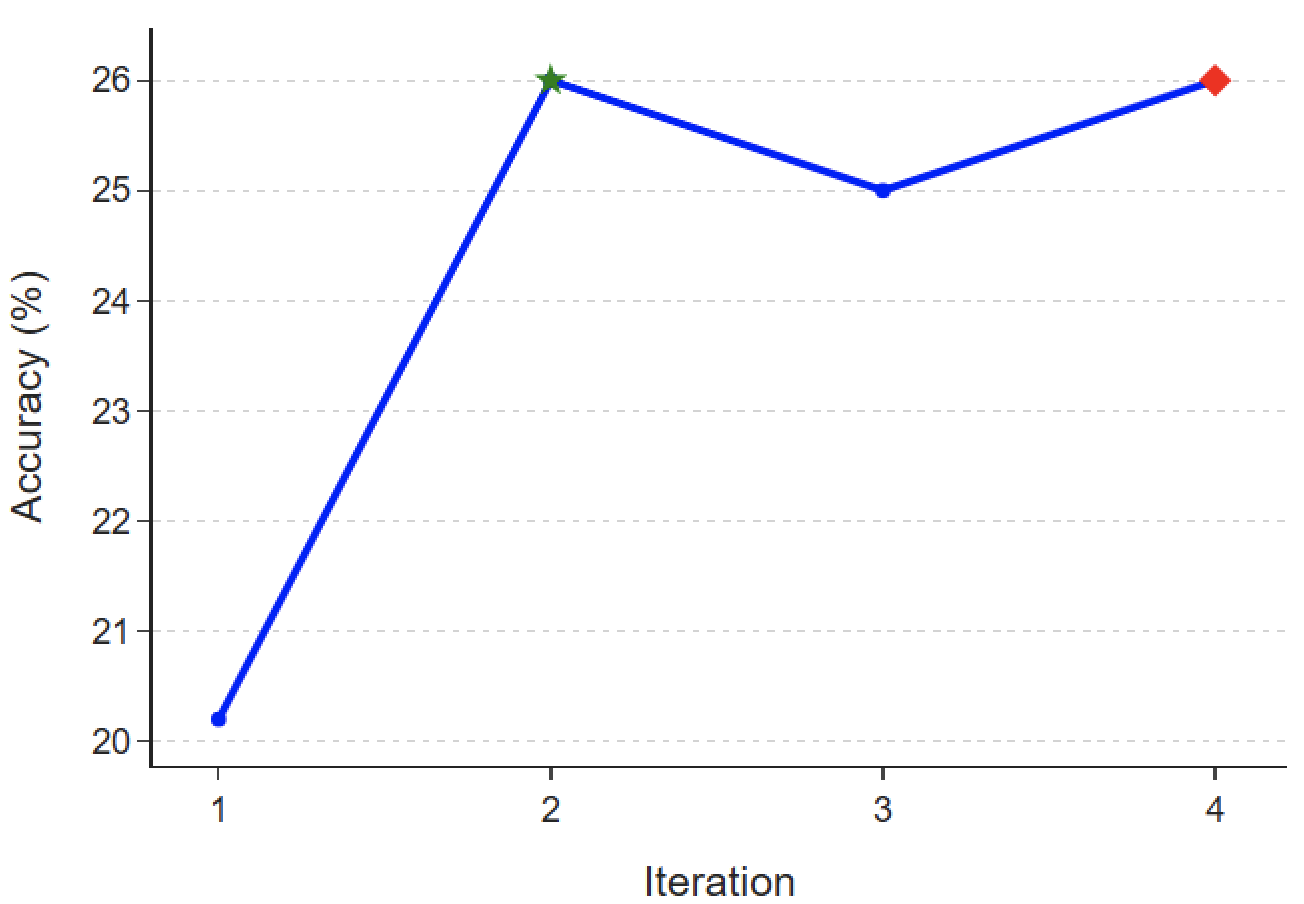}
        \caption{Math500}
    \end{subfigure}

    \caption{Accuracy progression of \method across 9 benchmark datasets for LLaMA 3.1 8B. Each curve represents the accuracy at successive iterations. The \textcolor{green}{$\star$} denotes the best-performing layer drop configuration, while the \textcolor{red}{$\square$} highlights the Best Speed up with at least Baseline Accuracy (BSBA) configuration.}
    \label{graph:dataset_results1}
\end{figure*}

See Figure \ref{graph:dataset_results1}.

\newpage

\section{General Pruning results}
\begin{table}[!ht]
\centering
\renewcommand{\arraystretch}{1.3}
\setlength{\tabcolsep}{4pt}
\resizebox{\linewidth}{!}{
\begin{tabular}{l|c|c|c|c}
\toprule
\textbf{Group} & Dataset & \textbf{Baseline } & \textbf{ Pruned Model} & speedup\\
\midrule
Common-sense &ARC-Easy      & 87.0 & 87.82 & 1.2\\
& ARC-Challenge & 75.86 & 75.00 & 1.21 \\
& CommonQA      & 72.20 & 64.70 & 1.1\\
& Winogrande    & 54.20 & 50.57 & 1.13\\
Reading & BoolQ & 85.0 & 85.5 & 1.17\\
& BIG-Bench     & 75.2 & 67.2 & 1.1\\
\bottomrule
\end{tabular}
}
\caption{Accuracy of LLaMA-3.1-8B (baseline) versus a pruned variant obtained by dropping layers selected through BSBA. For each task, BSBA identified removable layers, and we retained the intersection of layers that appeared in at least 75\% of tasks within the Common-sense group (layers 19, 22, 23, 27) and (layers 18, 21, 22, 28, 32) for Reading Comprehension tasks. These layers were then pruned globally from the model, and performance was re-evaluated across tasks. Speedup is reported relative to the baseline.}
\label{general-best}
\end{table}
\section{TALE evaluation with perplexity}
\label{appendix:perplexity}

\begin{table}[h!]
\centering
\renewcommand{\arraystretch}{1.3} 
\setlength{\tabcolsep}{6pt}
\resizebox{\linewidth}{!}{
\begin{tabular}{lcccc}
\toprule
\textbf{Model} &
\multicolumn{2}{c}{\textbf{WikiText2}} &
\multicolumn{2}{c}{\textbf{LAMBADA}} \\
\cmidrule(lr){2-3} \cmidrule(lr){4-5}
 & \textbf{Vanilla} & \textbf{TALE} & \textbf{Vanilla} & \textbf{TALE} \\
\midrule
LLaMA 3.1 8B & 24.6 & 24.9 & 28.1 & 28.9 \\
Lucie 7B     & 46.1 & 36.4 & 52.5 & 43.8 \\
\bottomrule
\end{tabular}
}
\caption{Perplexity scores for two models across WikiText2 and LAMBADA with Vanilla and TALE (sparisty 10\%) configurations.}
\label{tab:perplexity}
\end{table}



\section{More on pruning and a common pruned layers model}
\label{appendix:pruning}

\begin{figure*}[h]
    \centering
    \includegraphics[width=\linewidth]{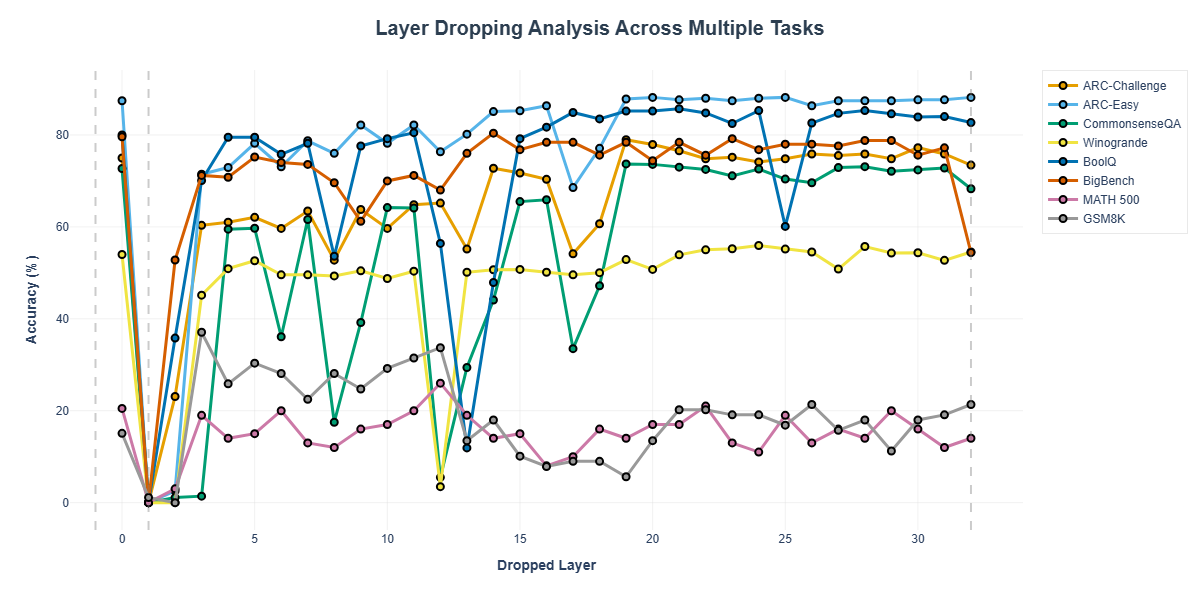}
    \caption{Nine benchmark tasks indicating performance after one layer is dropped from different positions in Llama3-8B.}
    \label{drop1}
\end{figure*}

See Figure \ref{drop1}

\newpage

\section{Comparison with Training-Free Pruning Technics}

\begin{table}[h]
\centering
\setlength{\tabcolsep}{4pt}
\resizebox{\linewidth}{!}{
\begin{tabular}{ll|c|ccc}
\hline
\rowcolor{teal!30}
\textbf{Model} & \textbf{Method} & \textbf{Sparsity} & \textbf{WinoGr} & \textbf{ARC-e} & \textbf{ARC-c} \\
\hline

& Baseline & 0\% & 41.2 & 51.7 & 40 \\
LLaMA-2-7B & SLEB & 10\% & 18 \percdd{56.3} & 29 \percdd{43.9} & 28.8 \percdd{28.0}\\

&  \cellcolor{blue!15}{TALE} &  \cellcolor{blue!15}{10\%} 
& \cellcolor{blue!15}{\bf 56 }\perc{35.9} 
& \cellcolor{blue!15}{\bf 62.3} \perc{20.5}
& \cellcolor{blue!15}{\bf 50 }\perc{25.0} \\

&  \cellcolor{blue!15}{TALE} &  \cellcolor{blue!15}{25\%} 
& \cellcolor{blue!15}{\bf 51}  \perc{23.8}
& \cellcolor{blue!15}{\bf 64.8 } \perc{25.3}
& \cellcolor{blue!15}{\bf 47.6 } \perc{19.0} \\
\hline

& Baseline & 0\% & 42 & 73.0 & 54.9 \\
LLaMA-2-13B & SLEB & 10\% & 24.2 \percdd{42.3} & 43.5  \percdd{40.4} & 29.8 \percdd{47.3}\\
& Blockpruner & 10\% & 24.2 \percdd{42.3} & 43.5  \percdd{40.4} & 29.8 \percdd{47.3}\\
&  \cellcolor{blue!15}{TALE} &  \cellcolor{blue!15}{10\% }
& \cellcolor{blue!15}{\bf 56.4}  \perc{34.3}
& \cellcolor{blue!15}{\bf 77.3}  \perc{5.9}
& \cellcolor{blue!15}{\bf 64.4 }\perc{17.1} \\

&  \cellcolor{blue!15}{TALE} & \cellcolor{blue!15} {25\%} 
& \cellcolor{blue!15}{\bf 55.2}  \perc{31.4}
& \cellcolor{blue!15}{\bf 75.3} \perc{3.2} 
& \cellcolor{blue!15}{\bf 64.1 }\perc{16.4} \\
\hline

\end{tabular}
}
\caption{Accuracies (\%) with Decoder Eval on 0-shot tasks for LLaMA-2-7B and LLaMA-2-13B}\label{tab:llama2_7b-13b_our-eval}
\end{table}

\section{Deleted Layers in each Model and Benchmark}

\begin{table}[!ht]
\centering
\renewcommand{\arraystretch}{1.3}
\resizebox{\linewidth}{!}{
\begin{tabular}{l|c|c}
\toprule
\textbf{Dataset} & \textbf{Best Model} & \textbf{BSBA} \\
\midrule
ARC-Easy      & \del{blue}{19} \del{blue}{20} \del{blue}{21} \del{blue}{29} \del{blue}{32} 
              & \del{orange}{19} \del{orange}{20} \del{orange}{21} \del{orange}{22} \del{orange}{25} \del{orange}{27} \del{orange}{29} \del{orange}{32} \\
ARC-Challenge & \del{blue}{19} \del{blue}{20} \del{blue}{23} \del{blue}{27} 
              & \del{orange}{19} \del{orange}{20} \del{orange}{21} \del{orange}{23} \del{orange}{25} \del{orange}{27} \del{orange}{28} \\
BoolQ         & \del{blue}{21} \del{blue}{23} \del{blue}{28} 
              & \del{orange}{18} \del{orange}{21} \del{orange}{22} \del{orange}{27} \del{orange}{28} \del{orange}{32} \\
MMLU          & \del{blue}{21} 
              & \del{orange}{19} \del{orange}{21} \del{orange}{22} \del{orange}{24} \del{orange}{25} \del{orange}{26} \del{orange}{27} \del{orange}{28} \del{orange}{31} \\
CommonQA      & \del{blue}{19} \del{blue}{23} \del{blue}{28} 
              & \del{orange}{19} \del{orange}{22} \del{orange}{23} \del{orange}{26} \del{orange}{27} \del{orange}{28} \\
Winogrande    & \del{blue}{23} \del{blue}{24} \del{blue}{26} \del{blue}{32} 
              & \del{orange}{20} \del{orange}{21} \del{orange}{22} \del{orange}{23} \del{orange}{24} \del{orange}{25} \del{orange}{26} \del{orange}{27} \del{orange}{29} \del{orange}{31} \del{orange}{32} \\
BIG-Bench     & \del{blue}{14} \del{blue}{20} \del{blue}{22} \del{blue}{28} \del{blue}{29} 
              & \del{orange}{14} \del{orange}{18} \del{orange}{20} \del{orange}{21} \del{orange}{22} \del{orange}{23} \del{orange}{24} \del{orange}{28} \del{orange}{29} \del{orange}{31} \del{orange}{32} \\
GSM8K-Hard    & \del{blue}{3} 
              & \del{orange}{3} \del{orange}{21} \del{orange}{22} \del{orange}{25} \del{orange}{26} \del{orange}{27} \del{orange}{29} \\
MATH500   & \del{blue}{11}
            \del{blue}{22}
              & \del{orange}{11} \del{orange}{22} \del{orange}{26} \\
        
\bottomrule
\end{tabular}}
\caption{Deleted layers represented as color-coded inline numbers. Blue = Best Model, Orange = BSBA for LlaMA 3.1 8B 0-shot.}
\label{deletedllama}
\end{table}

\begin{table}[!ht]
\centering
\renewcommand{\arraystretch}{1.3}
\resizebox{\linewidth}{!}{
\begin{tabular}{l|c|c}
\toprule
\textbf{Dataset} & \textbf{Best Model} & \textbf{BSBA} \\
\midrule
ARC-Easy      & \del{blue}{19} \del{blue}{22} \del{blue}{28} 
              & \del{orange}{6} \del{orange}{19} \del{orange}{22} \del{orange}{24} \del{orange}{26} \del{orange}{27} \del{orange}{28} \\
ARC-Challenge & \del{blue}{27} \del{blue}{28} 
              & \del{orange}{7} \del{orange}{22} \del{orange}{23} \del{orange}{26} \del{orange}{27} \del{orange}{28} \\
BoolQ         & \del{blue}{18} \del{blue}{21} \del{blue}{27} \del{blue}{28} 
              & \del{orange}{12} \del{orange}{19} \del{orange}{21} \del{orange}{22} \del{orange}{26} \del{orange}{27} \del{orange}{28} \\
MMLU          & \del{blue}{22} \del{blue}{23} \del{blue}{26} \del{blue}{27} \del{blue}{28} 
              & \del{orange}{18} \del{orange}{22} \del{orange}{23} \del{orange}{26} \del{orange}{27} \del{orange}{28} \\
CommonQA      & \del{blue}{22} \del{blue}{28} 
              & \del{orange}{6} \del{orange}{21} \del{orange}{22} \del{orange}{23} \del{orange}{27} \del{orange}{28} \\
Winogrande    & \del{blue}{22} \del{blue}{26} \del{blue}{27} 
              & \del{orange}{6} \del{orange}{20} \del{orange}{22} \del{orange}{25} \del{orange}{26} \del{orange}{27} \\
BIG-Bench     & \del{blue}{10} \del{blue}{19} \del{blue}{23} \del{blue}{25} \del{blue}{26} \del{blue}{27} 
              & \del{orange}{10} \del{orange}{19} \del{orange}{23} \del{orange}{25} \del{orange}{26} \del{orange}{27} \\

MATH500    & \del{blue}{17}
              & \del{orange}{5} \del{orange}{6} \del{orange}{17}
              \del{orange}{18}\\

\bottomrule
\end{tabular}}
\caption{ Deleted layers represented as color-coded inline numbers. Blue = Best Model, Orange = BSBA for Qwen 2.5 7B 0-shot.}
\label{deletedqwen}
\end{table}

\begin{table}[!ht]
\centering
\renewcommand{\arraystretch}{1.3}
\resizebox{\linewidth}{!}{
\begin{tabular}{l|c|c}
\toprule
\textbf{Dataset} & \textbf{Best Model} & \textbf{BSBA} \\
\midrule
ARC-Easy      & \del{blue}{15} \del{blue}{16} \del{blue}{23} \del{blue}{24} \del{blue}{27} \del{blue}{28} 
              & \del{orange}{13} \del{orange}{15} \del{orange}{16} \del{orange}{18} \del{orange}{19} \del{orange}{20} \del{orange}{21} \del{orange}{22} \del{orange}{23} \del{orange}{24} \del{orange}{25} \del{orange}{27} \del{orange}{28} \\
ARC-Challenge & \del{blue}{16} \del{blue}{18} \del{blue}{20} \del{blue}{21} \del{blue}{23} \del{blue}{25} \del{blue}{26} 
              & \del{orange}{15} \del{orange}{16} \del{orange}{18} \del{orange}{19} \del{orange}{20} \del{orange}{21} \del{orange}{22} \del{orange}{23} \del{orange}{25} \del{orange}{26} \del{orange}{28} \\
BoolQ         & \del{blue}{8} \del{blue}{17} \del{blue}{25} \del{blue}{28} \del{blue}{29} 
              & \del{orange}{5} \del{orange}{8} \del{orange}{11} \del{orange}{12} \del{orange}{13} \del{orange}{14} \del{orange}{15} \del{orange}{16} \del{orange}{17} \del{orange}{19} \del{orange}{20} \del{orange}{23} \del{orange}{25} \del{orange}{26} \del{orange}{27} \del{orange}{28} \del{orange}{29} \del{orange}{31} \\
MMLU          & \del{blue}{11} \del{blue}{12} \del{blue}{15} \del{blue}{16} \del{blue}{20} \del{blue}{21} \del{blue}{22} \del{blue}{28} 
              & \del{orange}{5} \del{orange}{10} \del{orange}{11} \del{orange}{12} \del{orange}{13} \del{orange}{14} \del{orange}{15} \del{orange}{16} \del{orange}{17} \del{orange}{18} \del{orange}{19} \del{orange}{20} \del{orange}{21} \del{orange}{22} \del{orange}{23} \del{orange}{24} \del{orange}{25} \del{orange}{26} \del{orange}{28} \del{orange}{30} \del{orange}{31} \\
CommonQA      & \del{blue}{11} \del{blue}{12} \del{blue}{27} 
              & \del{orange}{11} \del{orange}{12} \del{orange}{13} \del{orange}{15} \del{orange}{16} \del{orange}{17} \del{orange}{18} \del{orange}{19} \del{orange}{20} \del{orange}{21} \del{orange}{22} \del{orange}{23} \del{orange}{24} \del{orange}{25} \del{orange}{27} \del{orange}{28} \\
BIG-Bench     & \del{blue}{6} \del{blue}{7} \del{blue}{15} \del{blue}{17} \del{blue}{20} \del{blue}{21} \del{blue}{25} \del{blue}{26} \del{blue}{27} 
              & \del{orange}{6} \del{orange}{7} \del{orange}{13} \del{orange}{15} \del{orange}{17} \del{orange}{19} \del{orange}{20} \del{orange}{21} \del{orange}{22} \del{orange}{24} \del{orange}{25} \del{orange}{26} \del{orange}{27} \del{orange}{28} \del{orange}{29} \\
GSM8K-Hard    & \del{blue}{12} 
              & \del{orange}{12} \del{orange}{21} \del{orange}{23} \\

MATH500   & \del{blue}{12} 
              & \del{orange}{12} \del{orange}{21} \del{orange}{23} \\
\bottomrule
    \end{tabular}}
\caption{Deleted layers represented as color-coded inline numbers. Blue = Best Model, Orange = BSBA for Lucie 7B 0-shot.}\label{deletedlucie}
\end{table}

\begin{table}[!ht]
\centering
\renewcommand{\arraystretch}{1.3}
\resizebox{0.9\linewidth}{!}{
\begin{tabular}{l|c|c}
\toprule
\textbf{Dataset} & \textbf{Best Model} & \textbf{BSBA} \\
\midrule
ARC-Easy      & \del{blue}{21} \del{blue}{22} \del{blue}{24} \del{blue}{26} \del{blue}{29} 
              & \del{orange}{21} \del{orange}{22} \del{orange}{23} \del{orange}{24} \del{orange}{25} \del{orange}{26} \del{orange}{29} \del{orange}{30} \del{orange}{32} \\
ARC-Challenge & \del{blue}{22} \del{blue}{24} \del{blue}{25} \del{blue}{27} \del{blue}{28} \del{blue}{30} 
              & \del{orange}{21} \del{orange}{22} \del{orange}{24} \del{orange}{25} \del{orange}{26} \del{orange}{27} \del{orange}{28} \del{orange}{30} \\
BoolQ         & \del{blue}{17} \del{blue}{22} \del{blue}{23} \del{blue}{24} \del{blue}{27} \del{blue}{32} 
              & \del{orange}{12} \del{orange}{17} \del{orange}{21} \del{orange}{23} \del{orange}{24} \del{orange}{25} \del{orange}{27} \del{orange}{28} \del{orange}{32} \\
MMLU          & \del{blue}{24} \del{blue}{30} 
              & \del{orange}{22} \del{orange}{23} \del{orange}{24} \del{orange}{25} \del{orange}{26} \del{orange}{27} \del{orange}{30} \del{orange}{32} \\
CommonQA      & \del{blue}{19} \del{blue}{22} \del{blue}{25} \del{blue}{28} 
              & \del{orange}{19} \del{orange}{21} \del{orange}{22} \del{orange}{24} \del{orange}{25} \del{orange}{28} \del{orange}{32} \\
Winogrande    & \del{blue}{18} \del{blue}{19} \del{blue}{20} \del{blue}{22} \del{blue}{23} \del{blue}{24} \del{blue}{26} \del{blue}{27} \del{blue}{31} \del{blue}{32} 
              & \del{orange}{4} \del{orange}{13} \del{orange}{18} \del{orange}{19} \del{orange}{20} \del{orange}{22} \del{orange}{23} \del{orange}{24} \del{orange}{26} \del{orange}{27} \del{orange}{29} \del{orange}{31} \del{orange}{32} \\
BIG-Bench     & \del{blue}{3} \del{blue}{5} \del{blue}{15} \del{blue}{22} \del{blue}{23} \del{blue}{24} \del{blue}{26} \del{blue}{27} \del{blue}{28} 
              & \del{orange}{3} \del{orange}{5} \del{orange}{14} \del{orange}{15} \del{orange}{18} \del{orange}{22} \del{orange}{23} \del{orange}{24} \del{orange}{26} \del{orange}{27} \del{orange}{28} \\
GSM8K-Hard    & \del{blue}{6} \del{blue}{22} & \del{orange}{6} \del{orange}{11} \del{orange}{22} \del{orange}{28} \\
\bottomrule
\end{tabular}}
\caption{Deleted layers represented as color-ccdinline numbers. Blue = Best Model, Orange = BSBA for Mistral 0-shot.}
\end{table}

\hidden{
\begin{table}[!ht]
\centering
\renewcommand{\arraystretch}{1.3}
\resizebox{\linewidth}{!}{
\begin{tabular}{l|c|c}
\toprule
\textbf{Dataset} & \textbf{Best Model} & \textbf{BSBA} \\
\midrule
ARC-Easy      & \del{blue}{19} \del{blue}{25} \del{blue}{27} \del{blue}{28} & \del{orange}{19} \del{orange}{20} \del{orange}{21} \del{orange}{24} \del{orange}{25} \del{orange}{26} \del{orange}{27} \del{orange}{28} \\
ARC-Challenge & \del{blue}{19} \del{blue}{22} \del{blue}{27} & \del{orange}{19} \del{orange}{20} \del{orange}{21} \del{orange}{22} \del{orange}{23} \del{orange}{24} \del{orange}{26} \del{orange}{27} \del{orange}{28} \\
BoolQ         & \del{blue}{19} \del{blue}{25} \del{blue}{26} \del{blue}{32} & \del{orange}{15} \del{orange}{19} \del{orange}{21} \del{orange}{22} \del{orange}{25} \del{orange}{26} \del{orange}{30} \del{orange}{32} \\
MMLU          & \del{blue}{20} \del{blue}{21} \del{blue}{27} \del{blue}{28} & \del{orange}{20} \del{orange}{21} \del{orange}{22} \del{orange}{24} \del{orange}{27} \del{orange}{28} \del{orange}{32} \\
CommonQA      & \del{blue}{21} \del{blue}{22} \del{blue}{27} \del{blue}{28} \del{blue}{31} \del{blue}{32} & \del{orange}{21} \del{orange}{22} \del{orange}{23} \del{orange}{27} \del{orange}{28} \del{orange}{31} \del{orange}{32} \\
Winogrande    & \del{blue}{20} \del{blue}{22} \del{blue}{24} & \del{orange}{17} \del{orange}{19} \del{orange}{20} \del{orange}{22} \del{orange}{24} \del{orange}{26} \del{orange}{29} \del{orange}{32} \\
BIG-Bench     & \del{blue}{11} \del{blue}{16} \del{blue}{20} \del{blue}{21} \del{blue}{26} & \del{orange}{10} \del{orange}{11} \del{orange}{16} \del{orange}{20} \del{orange}{21} \del{orange}{22} \del{orange}{23} \del{orange}{24} \del{orange}{26} \del{orange}{27} \del{orange}{28} \del{orange}{29} \del{orange}{30} \del{orange}{31} \del{orange}{32}\\
MATH500       & \del{blue}{28} & \del{orange}{24} \del{orange}{28}\\
\bottomrule
\end{tabular}}
\caption{Deleted layers represented as color-coded inline numbers. Blue = Best Model, Orange = BSBA for LlaMA 3.1 8B with few-shot.}
\label{deletedllamafewshots}
\end{table}

\section{A tunable metric for finding accuracy vs. speed up optimization}
\label{appendix:aehm}

\textcolor{red}{Do we really need this section? - Because we didnt discuss anything about this in the main paper.}
 To systematically select among these candidates according to user priorities, we propose the Accuracy–Efficiency Harmonic Mean (AE-HM): 

\hidden{
If the best model corresponds to the pruning configuration that achieves the highest post-pruning accuracy, the BSBA model represents the most pruned architecture that still retains baseline-level accuracy but discarding as  many layers as is consistent with that objective to optimize speed up.  To better quantify the trade-off between accuracy retention and efficiency gains with \method, we define, \textbf{Accuracy-Efficiency Harmonic Mean (AE-HM)}, a composite metric:\footnote{Inspired by the F1-score’s use of a harmonic mean to penalize imbalance}   }
\begin{equation}
\begin{aligned}
r_A &= \frac{\text{Acc}(\text{Model})}{\text{Acc}(\text{Baseline})}, \\
\text{AE-HM}(\text{Model})
&= \frac{(1+\lambda^2) r_A\, S}{\lambda^2 S + r_A} \\
&= \frac{1+\lambda^2}{\frac{\lambda^2}{r_A}+ \frac{1}{S}}
\end{aligned}
\end{equation}
where $S$ denotes the relative inference speedup and $\lambda$ controls the relative importance of accuracy versus efficiency.  
The user can set AE-HE's parameter $\lambda$ to desired specifications: if $\lambda >1$, we prioritize $r_A$; if $\lambda <1$ we prioritize Speedup. 

By computing AE-HM for candidate models, we can automatically identify the model with the highest score for a given task or a set of tasks given a particular AE-HM parameter setting:
\begin{equation}
M_{\text{best-compromise}} = \arg\max_i \text{AE-HM}(M_i)
\end{equation}

\begin{figure}[!ht]
    \centering
    \includegraphics[width=\linewidth]{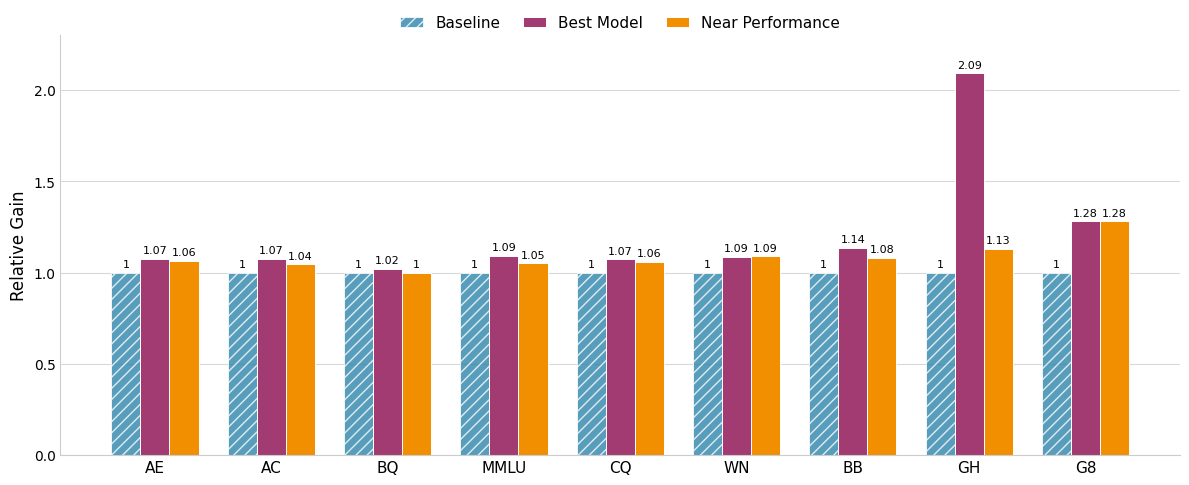}
    \caption{Relative Gain comparison across datasets. LLaMA $\beta=3$}
    \label{fig:relative_gain_comparison}
\end{figure}

}
\newpage

\hidden{
\begin{table*}[!ht]
 \renewcommand{\arraystretch}{1.3}  More breathing space
 \setlength{\tabcolsep}{6pt}  Column spacing

 \begin{subtable}[t]{\textwidth}
 \centering
 \resizebox{\linewidth}{!}{
 \begin{tabular}{l|c|ccc|ccc|c|ccc|ccc|}
 \toprule
 \rowcolor{orange!30}
 \multirow{2}{*}{\textbf{Dataset}} & 
 \multicolumn{7}{c|}{\textbf{LLaMA 3.1 8B (0-shot)}} & 
 \multicolumn{7}{c}{\textbf{Qwen 2.5 7B (0-shot)}}\\
 \cmidrule(lr){2-8} \cmidrule(lr){9-15} 
  & Baseline & \multicolumn{3}{c|}{Best Model} & \multicolumn{3}{c|}{BSBA} 
  & Baseline & \multicolumn{3}{c|}{Best Model} & \multicolumn{3}{c}{BSBA} \\
 \cmidrule(lr){2-2} \cmidrule(lr){3-5} \cmidrule(lr){6-8} 
 \cmidrule(lr){9-9} \cmidrule(lr){10-12} \cmidrule(lr){13-15}
  & Perf. & Perf. $\pm$ Std & \#D & Sp. & Perf. & \#D & Sp. 
  & Perf. & Perf. $\pm$ Std & \#D & Sp. & Perf. & \#D & Sp. \\
 \midrule
 ARC-Easy       & 89 & 90.4 $\pm$ 0.14 & 5 & \percd{14.6} & 88.8 & 8 & \percd{23.5} 
                & 90.04 & 93.2 $\pm$ 0.16 & 3 & \percd{10.0} & 90.08 & 7 & \percd{30.3} \\
 ARC-Challenge  & 79.4 & 80.6 $\pm$ 0.18 & 4 & \percd{11.7} & 77.6 & 7 & \percd{20.5} 
                & 86.55 & 92.00 $\pm$ 0.15 & 2 & \percd{6.7} & 86.55 & 6 & \percd{19.9} \\
 BoolQ          & 85.4 & 85.9 $\pm$ 0.12 & 3 & \percd{8.8} & 85.4 & 7 & \percd{17.6} 
                & 81.90 & 83.90 $\pm$ 0.14 & 4 & \percd{13.3} & 82.70 & 5 & \percd{23.2} \\
 MMLU           & 48.8 & 53.8 $\pm$ 0.22 & 1 & \percd{2.9} & 50.2 & 9 & \percd{26.4} 
                & 68.10 & 71.00 $\pm$ 0.19 & 5 & \percd{16.6} & 68.13 & 6 & \percd{19.9} \\
 CommonQA       & 72.9 & 73.8 $\pm$ 0.15 & 3 & \percd{8.8} & 73.10 & 6 & \percd{17.6} 
                & 80.30 & 84.40 $\pm$ 0.17 & 2 & \percd{6.6} & 80.50 & 6 & \percd{19.9} \\
 Winogrande     & 53.8 & 54.1 $\pm$ 0.20 & 4 & \percd{11.7} & 53.83 & 12 & \percd{32.2} 
                & 62.04 & 67.25 $\pm$ 0.21 & 3 & \percd{10.0} & 62.19 & 6 & \percd{19.9} \\
 BIG-Bench      & 77.2 & 85.6 $\pm$ 0.25 & 5 & \percd{14.4} & 76.4 & 11 & \percd{32.2} 
                & 79.20 & 81.60 $\pm$ 0.14 & 6 & \percd{19.9} & 81.60 & 6 & \percd{19.9} \\
 GSM8K-HARD     & 39.0 & 59.0 $\pm$ 0.30 & 1 & \percd{2.9} & 39.4 & 4 & \percd{11.7} 
                & 43.80 & 61.80 $\pm$ 0.28 & 2 & \percd{43.6} & 43.99 & 5 & \percd{17.6} \\
 Math500        & 25.4 & 28.2 $\pm$ 0.18 & 2 & \percd{6.0} & 27.4 & 3 & \percd{9.1} 
                & 31.00 & 38.20 $\pm$ 0.22 & 2 & \percd{6.6} & 32.10 & 4 & \percd{13.3} \\
 \bottomrule
 \end{tabular}
 }
 \end{subtable}

 \vspace{0.5cm}

 \begin{subtable}[t]{\textwidth}
 \centering
 \resizebox{\linewidth}{!}{
 \begin{tabular}{l|c|ccc|ccc|c|ccc|ccc|}
 \toprule
 \rowcolor{orange!30}
 \multirow{2}{*}{\textbf{Dataset}} & 
 \multicolumn{7}{c|}{\textbf{Lucie 7B (0-shot)}} & 
 \multicolumn{7}{c}{\textbf{Mistral 7B (0-shot)}}\\
 \cmidrule(lr){2-8} \cmidrule(lr){9-15}
  & Baseline & \multicolumn{3}{c|}{Best Model} & \multicolumn{3}{c|}{BSBA} 
  & Baseline & \multicolumn{3}{c|}{Best Model} & \multicolumn{3}{c}{BSBA} \\
 \cmidrule(lr){2-2} \cmidrule(lr){3-5} \cmidrule(lr){6-8}
 \cmidrule(lr){9-9} \cmidrule(lr){10-12} \cmidrule(lr){13-15}
  & Perf. & Perf. $\pm$ Std & \#D & Sp. & Perf. & \#D & Sp.
  & Perf. & Perf. $\pm$ Std & \#D & Sp. & Perf. & \#D & Sp. \\
 \midrule
 ARC-Easy        & 74.4 & 75.8 $\pm$ 0.15 & 6 & \percd{18.1} & 73.8 & 8 & \percd{23.5}
                 & 83.8 & 85.6 $\pm$ 0.14 & 5 & \percd{15.4} & 82.8 & 9 & \percd{27.7} \\
 ARC-Challenge   & 46.0 & 51.45 $\pm$ 0.22 & 7 & \percd{22.1} & 48.8 & 11 & \percd{33.1}
                 & 76.2 & 79.1 $\pm$ 0.18 & 6 & \percd{18.5} & 76.2 & 8 & \percd{24.6} \\
 BoolQ           & 53.0 & 74.0 $\pm$ 0.30 & 5 & \percd{17.2} & 63.0 & 19 & \percd{54.2}
                 & 81.3 & 84.4 $\pm$ 0.16 & 4 & \percd{18.5} & 80.8 & 5 & \percd{27.7} \\
 MMLU            & 13.0 & 54.0 $\pm$ 0.35 & 8 & \percd{24.1} & 15 & 22 & \percd{60.2}
                 & 39.4 & 40.8 $\pm$ 0.20 & 2 & \percd{6.2} & 39.0 & 8 & \percd{24.6} \\
 CommonQA        & 54.2 & 68.6 $\pm$ 0.27 & 3 & \percd{9.1} & 54.6 & 17 & \percd{48.2}
                 & 61.0 & 64.4 $\pm$ 0.19 & 4 & \percd{12.3} & 61.6 & 7 & \percd{21.5} \\
 Winogrande      & 51.6 & 53.1 $\pm$ 0.21 & 5 & \percd{27.1} & 53.0 & 15 & \percd{45.2}
                 & 53.2 & 54.3 $\pm$ 0.16 & 10 & \percd{30.7} & 52.4 & 13 & \percd{40.0} \\
 BIG-Bench       & 67.4 & 75.0 $\pm$ 0.25 & 9 & \percd{27.1} & 71 & 15 & \percd{45.1}
                 & 70.4 & 75.4 $\pm$ 0.22 & 9 & \percd{28.0} & 72.6 & 11 & \percd{33.8} \\
 GSM8K-HARD      & 32 & 39.0 $\pm$ 0.28 & 1 & \percd{3.1} & 37 & 3 & \percd{9.1}
                 & 24 & 33 $\pm$ 0.24 & 2 & \percd{6.2} & 26.1 & 4 & \percd{12.3} \\
 Math500         & 21.0 & 26.1 $\pm$ 0.20 & 2 & \percd{6.0} & 25.1 & 3 & \percd{9.1}
                 & 19 & 28 $\pm$ 0.23 & 1 & \percd{3.1} & 18.8 & 4 & \percd{12.3} \\
 \bottomrule
 \end{tabular}
 }
\end{subtable}
\caption{ Robustness study of the proposed layer-dropping method across multiple language models under 0-shot evaluation. For each dataset and model, results are reported over five random seeds to account for variability in decoding and sampling. We present the baseline model accuracy and the accuracy of the best pruned configuration, along with their corresponding standard deviations computed across the 5 seeds. The table also includes the number of transformer layers removed in the best-performing configuration (\textbf{\#D}) and the resulting inference speedup (\textbf{Sp.}) expressed as the percentage of total TFlops saved during evaluation. All experiments use 10\% of the training split for optimization and evaluate on the respective test sets. Bold values indicate the highest mean accuracy for each dataset.}
\end{table*}

}

\hidden{

\del{}{}
\begin{table}[!ht]
\centering
\renewcommand{\arraystretch}{1.3}
\resizebox{\linewidth}{!}{
\begin{tabular}{l|ccc|ccc|ccc}
\toprule
\multirow{3}{*}{\textbf{Dataset}} & 
\multicolumn{9}{c}{\textbf{Qwen 2.5 0.5B 0 shot}} \\
\cmidrule(lr){2-10}
 & \multicolumn{3}{c|}{Baseline} & \multicolumn{3}{c|}{Best Model} & \multicolumn{3}{c}{BSBA} \\
\cmidrule(lr){2-4} \cmidrule(lr){5-7} \cmidrule(lr){8-10}
& Perf. & \#D & Sp. & Perf. & \#D & Sp. & Perf. & \#D & Sp. \\
\midrule
ARC-Easy      & 40.00 & 0 & 1.00 & \gain{60.91}\textcolor{green}{(+48.49\% $\uparrow$)}  & 3 & 1.12 & 48.36 & 5 & 1.36 \\
ARC-Challenge & 35.52 & 0 & 1 & 40.34\textcolor{green}{(+13.57\% $\uparrow$)} & 1 & 1.11 &  37.24 & 4 & 1.43  \\
BoolQ         & 62.3 & 0 & 1.00 & 67.2\textcolor{green}{(+7.87\% $\uparrow$)} & 5 & 1.42 & 66.2 & 6 & 1.48 \\
MMLU          & 31.48 & 0 & 1.00 & 39.97\textcolor{green}{(+26.96\% $\uparrow$)} & 2 & 1.10 & 33.90 & 5 & 1.36 \\
COMMONQA      & 42.40 & 0 & 1.00 & 49.10\textcolor{green}{(+15.8\% $\uparrow$)} & 2 & 1.24 & 44.00 & 3 & 1.38 \\
WINOGRANDE    & 49.86 & 0 & 1.00 & 51.88\textcolor{green}{(+04.51\% $\uparrow$)} & 5 & 1.3 &49.87 & 17 & 3.9 \\
BIG-Bench     & 72.40 & 0 & 1.00 & 73.60\textcolor{green}{(+1.66\% $\uparrow$)} & 2 & - & 73.60 & 2 & 1.14 \\
GSM8K-HARD    & 6.74 & 0 & 1.00 & 11.24\textcolor{green}{(+66.77\% $\uparrow$)} & 1 & 1.15 & 8.99 & 2 & 1.16 \\
MATH500        & -- & 0 & 1.00 & -- & -- & -- & -- & -- & -- \\
\bottomrule
\end{tabular}}
\caption{Results of \textbf{Qwen 2.5 0.5B} across nine benchmarks. 
Performance (\%) cells are color-coded: \textcolor{green!50!black}{green = gain}, 
\textcolor{red!60!black}{red = decline}, and \textcolor{gray!70!black}{gray = near-neutral change} compared to baseline.}
\label{tab:quen-small-results}
\end{table}

\begin{table}[!ht]
\centering
\renewcommand{\arraystretch}{1.3}
\resizebox{0.8\linewidth}{!}{
\begin{tabular}{l|c|cc|cc}
\toprule
\multirow{3}{*}{\textbf{Dataset}} & 
\multicolumn{5}{c}{\textbf{Qwen 2.5 0.5B 0-shot}} \\
\cmidrule(lr){2-6}
 & \multicolumn{1}{c|}{Baseline} & \multicolumn{2}{c|}{Best Model} & \multicolumn{2}{c}{BSBA} \\
\cmidrule(lr){2-2} \cmidrule(lr){3-4} \cmidrule(lr){5-6}
& Perf. & Perf. & \#D & Perf. & \#D \\
\midrule
ARC-Easy      & 40.00 & 60.91 & 3 & 48.36 & 5 \\
ARC-Challenge &  35.52 & 40.34 & 1 & 37.24 & 4 \\
BoolQ           & 62.30 & \textbf{67.20}\perc{7.87} & 5 & \percd{15.5} & 66.20 & 6 & \percd{18.6} \\
MMLU            & 31.48 & \textbf{39.97}\perc{26.96} & 2 & \percd{6.2} & 33.90 & 5 & \percd{15.5} \\
CommonQA        & 42.40 & \textbf{49.10}\perc{15.80} & 2 & \percd{6.2} & 44.00 & 3 & \percd{9.3} \\
Winogrande      & 49.86 & \textbf{51.88}\perc{4.51}  & 5 & \percd{15.5} & 49.87 & 17 & \percd{52.6} \\
BIG-Bench       & 72.40 & \textbf{73.60}\perc{1.66}  & 2 & \percd{6.2}  & 73.60 & 2 & \percd{6.2} \\
GSM8K-HARD      &  6.74 & \textbf{11.24}\perc{66.77} & 1 & \percd{3.1} & 8.99 & 2 & \percd{6.2} \\
Math500         & 8.00   & \textbf{12.00}\perc{50} & 1 & \percd{3.1} & 9 & 2 & \percd{6.2} \\
\bottomrule
\end{tabular}}
\caption{Results of \textbf{Lucie 7B} across nine benchmarks. All tested on 5-shots, except gms8k on 8-shots 
Performance (\%) cells are color-coded: \textcolor{green!50!black}{green = gain}, 
\textcolor{red!60!black}{red = decline}, and \textcolor{gray!70!black}{gray = near-neutral change} compared to baseline.}
\label{tab:smallqwen_zeroshot}
\end{table}

\begin{table}[!ht]
\centering
\caption{Performance comparison under 0-shot evaluation. Accuracy (\textbf{Perf.}) uses Decoder Eval. We also report number of dropped layers (\textbf{\#D}), and relative inference speedup (\textbf{Sp.}) in terms of percentage of Tflops saved (Percentage saved = $\frac{\text{Tflops}_{\text{Baseline}} - \text{Tflops}_{\text{Pruned-model}}}{\text{Tflops}_{\text{Baseline}}} \times 100$).  Percentage gain = $\frac{\text{Acc}_{\text{Best}} - \text{Acc}_{\text{Baseline}}}{\text{Acc}_{\text{Baseline}}} \times 100$. Best accuracy is highlighted in \textbf{bold}; BSBA shows balanced trade-offs.}
\resizebox{\linewidth}{!}{ 
\begin{tabular}{l|c|ccc|ccc|}
\toprule
\rowcolor{orange!30}
\multirow{2}{*}{\textbf{Dataset}} & 
\multicolumn{7}{c}{\textbf{Qwen 2.5 0.5B (0-shot)}} \\
\cmidrule(lr){2-8}
 & Baseline & \multicolumn{3}{c|}{Best Model} & \multicolumn{3}{c}{BSBA} \\
\cmidrule(lr){2-2} \cmidrule(lr){3-5} \cmidrule(lr){6-8}
 & Perf. & Perf. & \#D & Sp. & Perf. & \#D & Sp. \\
\midrule
ARC-Easy        & 40.00 & \textbf{60.91}\perc{48.49} & 3 & \percd{9.3} & 48.36 & 5 & \percd{15.5} \\
ARC-Challenge   & 35.52 & \textbf{40.34}\perc{13.57} & 1 & \percd{3.1} & 37.24 & 4 & \percd{12.4} \\
BoolQ           & 62.30 & \textbf{67.20}\perc{7.87} & 5 & \percd{15.5} & 66.20 & 6 & \percd{18.6} \\
MMLU            & 31.48 & \textbf{39.97}\perc{26.96} & 2 & \percd{6.2} & 33.90 & 5 & \percd{15.5} \\
CommonQA        & 42.40 & \textbf{49.10}\perc{15.80} & 2 & \percd{6.2} & 44.00 & 3 & \percd{9.3} \\
Winogrande      & 49.86 & \textbf{51.88}\perc{4.51}  & 5 & \percd{15.5} & 49.87 & 17 & \percd{52.6} \\
BIG-Bench       & 72.40 & \textbf{73.60}\perc{1.66}  & 2 & \percd{6.2}  & 73.60 & 2 & \percd{6.2} \\
GSM8K-HARD      &  6.74 & \textbf{11.24}\perc{66.77} & 1 & \percd{3.1} & 8.99 & 2 & \percd{6.2} \\
Math500         & 8.00   & \textbf{12.00}\perc{50} & 1 & \percd{3.1} & 9 & 2 & \percd{6.2} \\
\bottomrule
\end{tabular}
\label{tab:qwen}
}
\end{table}
}




\hidden{
\begin{table}[!ht]
\centering
\renewcommand{\arraystretch}{1.3}
\resizebox{\linewidth}{!}{
\begin{tabular}{l|c|c}
\toprule
\textbf{Dataset} & \textbf{Best Model} & \textbf{BSBA} \\
\midrule
ARC-Easy      & \del{blue}{19} \del{blue}{20} \del{blue}{21} \del{blue}{29} \del{blue}{32} 
              & \del{orange}{19} \del{orange}{20} \del{orange}{21} \del{orange}{22} \del{orange}{25} \del{orange}{27} \del{orange}{29} \del{orange}{32} \\
ARC-Challenge & \del{blue}{19} \del{blue}{20} \del{blue}{23} \del{blue}{27} 
              & \del{orange}{19} \del{orange}{20} \del{orange}{21} \del{orange}{23} \del{orange}{25} \del{orange}{27} \del{orange}{28} \\
BoolQ         & \del{blue}{21} \del{blue}{23} \del{blue}{28} 
              & \del{orange}{18} \del{orange}{21} \del{orange}{22} \del{orange}{27} \del{orange}{28} \del{orange}{32} \\
MMLU          & \del{blue}{21} 
              & \del{orange}{19} \del{orange}{21} \del{orange}{22} \del{orange}{24} \del{orange}{25} \del{orange}{26} \del{orange}{27} \del{orange}{28} \del{orange}{31} \\
CommonQA      & \del{blue}{19} \del{blue}{23} \del{blue}{28} 
              & \del{orange}{19} \del{orange}{22} \del{orange}{23} \del{orange}{26} \del{orange}{27} \del{orange}{28} \\
Winogrande    & \del{blue}{23} \del{blue}{24} \del{blue}{26} \del{blue}{32} 
              & \del{orange}{20} \del{orange}{21} \del{orange}{22} \del{orange}{23} \del{orange}{24} \del{orange}{25} \del{orange}{26} \del{orange}{27} \del{orange}{29} \del{orange}{31} \del{orange}{32} \\
BIG-Bench     & \del{blue}{14} \del{blue}{20} \del{blue}{22} \del{blue}{28} \del{blue}{29} 
              & \del{orange}{14} \del{orange}{18} \del{orange}{20} \del{orange}{21} \del{orange}{22} \del{orange}{23} \del{orange}{24} \del{orange}{28} \del{orange}{29} \del{orange}{31} \del{orange}{32} \\
GSM8K-Hard    & \del{blue}{3} 
              & \del{orange}{3} \del{orange}{21} \del{orange}{22} \del{orange}{25} \del{orange}{26} \del{orange}{27} \del{orange}{29} \\
              
\bottomrule
\end{tabular}}
\caption{Deleted layers represented as color-coded inline numbers. Blue = Best Model, Orange = BSBA for LlaMA 3.1 8B 0 shot.}
\label{deletedllama}
\end{table}
}

\hidden{

\begin{table}[!ht]
\centering
\renewcommand{\arraystretch}{1.3}
\resizebox{\linewidth}{!}{
\begin{tabular}{l|c|c}
\toprule
\textbf{Dataset} & \textbf{Best Model} & \textbf{BSBA} \\
\midrule
ARC-Easy      & \del{blue}{15} \del{blue}{16} \del{blue}{23} \del{blue}{24} \del{blue}{27} \del{blue}{28} 
              & \del{orange}{13} \del{orange}{15} \del{orange}{16} \del{orange}{18} \del{orange}{19} \del{orange}{20} \del{orange}{21} \del{orange}{22} \del{orange}{23} \del{orange}{24} \del{orange}{25} \del{orange}{27} \del{orange}{28} \\
ARC-Challenge & \del{blue}{16} \del{blue}{18} \del{blue}{20} \del{blue}{21} \del{blue}{23} \del{blue}{25} \del{blue}{26} 
              & \del{orange}{15} \del{orange}{16} \del{orange}{18} \del{orange}{19} \del{orange}{20} \del{orange}{21} \del{orange}{22} \del{orange}{23} \del{orange}{25} \del{orange}{26} \del{orange}{28} \\
BoolQ         & \del{blue}{8} \del{blue}{17} \del{blue}{25} \del{blue}{28} \del{blue}{29} 
              & \del{orange}{5} \del{orange}{8} \del{orange}{11} \del{orange}{12} \del{orange}{13} \del{orange}{14} \del{orange}{15} \del{orange}{16} \del{orange}{17} \del{orange}{19} \del{orange}{20} \del{orange}{23} \del{orange}{25} \del{orange}{26} \del{orange}{27} \del{orange}{28} \del{orange}{29} \del{orange}{31} \\
MMLU          & \del{blue}{11} \del{blue}{12} \del{blue}{15} \del{blue}{16} \del{blue}{20} \del{blue}{21} \del{blue}{22} \del{blue}{28} 
              & \del{orange}{5} \del{orange}{10} \del{orange}{11} \del{orange}{12} \del{orange}{13} \del{orange}{14} \del{orange}{15} \del{orange}{16} \del{orange}{17} \del{orange}{18} \del{orange}{19} \del{orange}{20} \del{orange}{21} \del{orange}{22} \del{orange}{23} \del{orange}{24} \del{orange}{25} \del{orange}{26} \del{orange}{28} \del{orange}{30} \del{orange}{31} \\
CommonQA      & \del{blue}{11} \del{blue}{12} \del{blue}{27} 
              & \del{orange}{11} \del{orange}{12} \del{orange}{13} \del{orange}{15} \del{orange}{16} \del{orange}{17} \del{orange}{18} \del{orange}{19} \del{orange}{20} \del{orange}{21} \del{orange}{22} \del{orange}{23} \del{orange}{24} \del{orange}{25} \del{orange}{27} \del{orange}{28} \\
BIG-Bench     & \del{blue}{6} \del{blue}{7} \del{blue}{15} \del{blue}{17} \del{blue}{20} \del{blue}{21} \del{blue}{25} \del{blue}{26} \del{blue}{27} 
              & \del{orange}{6} \del{orange}{7} \del{orange}{13} \del{orange}{15} \del{orange}{17} \del{orange}{19} \del{orange}{20} \del{orange}{21} \del{orange}{22} \del{orange}{24} \del{orange}{25} \del{orange}{26} \del{orange}{27} \del{orange}{28} \del{orange}{29} \\
GSM8K-Hard    & \del{blue}{12} 
              & \del{orange}{12} \del{orange}{21} \del{orange}{23} \\
\bottomrule
    \end{tabular}}
\caption{Deleted layers represented as color-coded inline numbers. Blue = Best Model, Orange = BSBA for Lucie 7B 0 shot.}\label{deletedlucie}
\end{table}

\begin{table}[!ht]
\centering
\renewcommand{\arraystretch}{1.3}
\resizebox{\linewidth}{!}{
\begin{tabular}{l|c|c}
\toprule
\textbf{Dataset} & \textbf{Best Model} & \textbf{BSBA} \\
\midrule
ARC-Easy      & \del{blue}{21} \del{blue}{22} \del{blue}{24} \del{blue}{26} \del{blue}{29} 
              & \del{orange}{21} \del{orange}{22} \del{orange}{23} \del{orange}{24} \del{orange}{25} \del{orange}{26} \del{orange}{29} \del{orange}{30} \del{orange}{32} \\
ARC-Challenge & \del{blue}{22} \del{blue}{24} \del{blue}{25} \del{blue}{27} \del{blue}{28} \del{blue}{30} 
              & \del{orange}{21} \del{orange}{22} \del{orange}{24} \del{orange}{25} \del{orange}{26} \del{orange}{27} \del{orange}{28} \del{orange}{30} \\
BoolQ         & \del{blue}{17} \del{blue}{22} \del{blue}{23} \del{blue}{24} \del{blue}{27} \del{blue}{32} 
              & \del{orange}{12} \del{orange}{17} \del{orange}{21} \del{orange}{23} \del{orange}{24} \del{orange}{25} \del{orange}{27} \del{orange}{28} \del{orange}{32} \\
MMLU          & \del{blue}{24} \del{blue}{30} 
              & \del{orange}{22} \del{orange}{23} \del{orange}{24} \del{orange}{25} \del{orange}{26} \del{orange}{27} \del{orange}{30} \del{orange}{32} \\
CommonQA      & \del{blue}{19} \del{blue}{22} \del{blue}{25} \del{blue}{28} 
              & \del{orange}{19} \del{orange}{21} \del{orange}{22} \del{orange}{24} \del{orange}{25} \del{orange}{28} \del{orange}{32} \\
Winogrande    & \del{blue}{18} \del{blue}{19} \del{blue}{20} \del{blue}{22} \del{blue}{23} \del{blue}{24} \del{blue}{26} \del{blue}{27} \del{blue}{31} \del{blue}{32} 
              & \del{orange}{4} \del{orange}{13} \del{orange}{18} \del{orange}{19} \del{orange}{20} \del{orange}{22} \del{orange}{23} \del{orange}{24} \del{orange}{26} \del{orange}{27} \del{orange}{29} \del{orange}{31} \del{orange}{32} \\
BIG-Bench     & \del{blue}{3} \del{blue}{5} \del{blue}{15} \del{blue}{22} \del{blue}{23} \del{blue}{24} \del{blue}{26} \del{blue}{27} \del{blue}{28} 
              & \del{orange}{3} \del{orange}{5} \del{orange}{14} \del{orange}{15} \del{orange}{18} \del{orange}{22} \del{orange}{23} \del{orange}{24} \del{orange}{26} \del{orange}{27} \del{orange}{28} \\
GSM8K-Hard    & \del{blue}{6} \del{blue}{22} & \del{orange}{6} \del{orange}{11} \del{orange}{22} \del{orange}{28} \\
\bottomrule
\end{tabular}}
\caption{Deleted layers represented as color-ccdinline numbers. Blue = Best Model, Orange = BSBA for \textbf{Mistral} 0-shot.}
\end{table}

\begin{table}[!ht]
\centering
\renewcommand{\arraystretch}{1.3}
\resizebox{\linewidth}{!}{
\begin{tabular}{l|c|c}
\toprule
\textbf{Dataset} & \textbf{Best Model} & \textbf{BSBA} \\
\midrule
ARC-Easy      & \del{blue}{19} \del{blue}{25} \del{blue}{27} \del{blue}{28} & \del{orange}{19} \del{orange}{20} \del{orange}{21} \del{orange}{24} \del{orange}{25} \del{orange}{26} \del{orange}{27} \del{orange}{28} \\
ARC-Challenge & \del{blue}{19} \del{blue}{22} \del{blue}{27} & \del{orange}{19} \del{orange}{20} \del{orange}{21} \del{orange}{22} \del{orange}{23} \del{orange}{24} \del{orange}{26} \del{orange}{27} \del{orange}{28} \\
BoolQ         & \del{blue}{19} \del{blue}{25} \del{blue}{26} \del{blue}{32} & \del{orange}{15} \del{orange}{19} \del{orange}{21} \del{orange}{22} \del{orange}{25} \del{orange}{26} \del{orange}{30} \del{orange}{32} \\
MMLU          & \del{blue}{20} \del{blue}{21} \del{blue}{27} \del{blue}{28} & \del{orange}{20} \del{orange}{21} \del{orange}{22} \del{orange}{24} \del{orange}{27} \del{orange}{28} \del{orange}{32} \\
CommonQA      & \del{blue}{21} \del{blue}{22} \del{blue}{27} \del{blue}{28} \del{blue}{31} \del{blue}{32} & \del{orange}{21} \del{orange}{22} \del{orange}{23} \del{orange}{27} \del{orange}{28} \del{orange}{31} \del{orange}{32} \\
Winogrande    & \del{blue}{20} \del{blue}{22} \del{blue}{24} & \del{orange}{17} \del{orange}{19} \del{orange}{20} \del{orange}{22} \del{orange}{24} \del{orange}{26} \del{orange}{29} \del{orange}{32} \\
BIG-Bench     & \del{blue}{11} \del{blue}{16} \del{blue}{20} \del{blue}{21} \del{blue}{26} & \del{orange}{10} \del{orange}{11} \del{orange}{16} \del{orange}{20} \del{orange}{21} \del{orange}{22} \del{orange}{23} \del{orange}{24} \del{orange}{26} \del{orange}{27} \del{orange}{28} \del{orange}{29} \del{orange}{30} \del{orange}{31} \del{orange}{32}\\
MATH500       & \del{blue}{28} & \del{orange}{24} \del{orange}{28}\\
\bottomrule
\end{tabular}}
\caption{Deleted layers represented as color-coded inline numbers. Blue = Best Model, Orange = BSBA for LlaMA 3.1 8B with few-shot.}
\label{deletedllamafewshots}
\end{table}
}
\newpage
\hidden{
 ---------------------- LLaMA 3.1 8B ----------------------
\begin{table}[!ht]
\centering
\renewcommand{\arraystretch}{1.2}

\resizebox{\linewidth}{!}{
\begin{tabular}{l|ccc|ccc|ccc}
\toprule
\multirow{3}{*}{\textbf{Dataset}} & 
\multicolumn{9}{c}{\textbf{LLaMA 3.1 8B}} \\
\cmidrule(lr){2-10}
 & \multicolumn{3}{c|}{Baseline} & \multicolumn{3}{c|}{Best Model} & \multicolumn{3}{c}{BSBA} \\
\cmidrule(lr){2-4} \cmidrule(lr){5-7} \cmidrule(lr){8-10}
& Perf. & \#D & Sp. & Perf. & \#D & Sp. & Perf. & \#D & Sp. \\
\midrule
ARC-Easy       & 87.00 & 0 & 1.00 & \textbf{90.55} \textcolor{green}{(+3.55\% $\uparrow$)} & 5 & 1.27 & 87.82  & \textbf{8} & \textbf{1.37} \\
ARC-Challenge  & 75.86 & 0 & 1.00 & \textbf{78.62} \textcolor{green}{(+2.76\% $\uparrow$)} & 4 & 1.26 & 76.90 \textcolor{green}{(+1.04\% $\uparrow$)} & \textbf{7} & \textbf{1.41} \\
BoolQ          & 85.00 & 0 & 1.00 & \textbf{86.20} \textcolor{green}{(+1.20\% $\uparrow$)} & 3 & 1.10 & 85.70 & \textbf{7} & \textbf{1.36} \\
MMLU           & 54.87 & 0 & 1.00 & \textbf{59.90} \textcolor{green}{(+5.03\% $\uparrow$)} & 1 & 1.05 & 54.87 & \textbf{9} & \textbf{1.37} \\
COMMONQA       & 72.20 & 0 & 1.00 & \textbf{75.30} \textcolor{green}{(+3.10\% $\uparrow$)} & 3 & 1.21 & 73.10  & 6 & \textbf{1.34} \\
WINOGRANDE     & 53.83 & 0 & 1.00 & \textbf{56.67} \textcolor{green}{(+2.84\% $\uparrow$)} & 4 & 1.25 & 55.09 & \textbf{11} & \textbf{1.46} \\
BIG-Bench      & 75.20 & 0 & 1.00 & \textbf{83.60} \textcolor{green}{(+8.40\% $\uparrow$)} & 5 & 1.24 & 75.20 & 11 & 1.58 \\
GSM8K-HARD     & 15.07 & 0 & 1.00 & \textbf{38.02} \textcolor{green}{(+152.28\% $\uparrow$)} & 1 & 1.12 & 17.81 & \textbf{7} & \textbf{1.27} \\
MATH500         & 42.15 & 0 & 1.00 & \textbf{56.56} \textcolor{green}{(+14.41\% $\uparrow$)} & 1 & --   & 48.14 \textcolor{green}{(+5.99\% $\uparrow$)} & 3 (redo for 7)& -- \\
\bottomrule
\end{tabular}}
\caption{Results of \textbf{LLaMA 3.1 8B} across nine benchmarks. We report accuracy (\%), number of layers dropped, and relative inference speedup. 
Best accuracy per dataset is in \textbf{bold}. 
\textcolor{green}{(+X\% $\uparrow$)} shows accuracy improvement from baseline, 
\textcolor{red}{(-X\% $\downarrow$)} shows accuracy decline.}
\label{tab:llama_results}
\end{table}

\begin{table}[!ht]
\centering
\renewcommand{\arraystretch}{1.2}

\resizebox{\linewidth}{!}{
\begin{tabular}{l|ccc|ccc|ccc}
\toprule
\multirow{3}{*}{\textbf{Dataset}} & 
\multicolumn{9}{c}{\textbf{Qwen 2.5 7B}} \\
\cmidrule(lr){2-10}
 & \multicolumn{3}{c|}{Baseline} & \multicolumn{3}{c|}{Best Model} & \multicolumn{3}{c}{BSBA} \\
\cmidrule(lr){2-4} \cmidrule(lr){5-7} \cmidrule(lr){8-10}
& Perf. & \#D & Sp. & Perf. & \#D & Sp. & Perf. & \#D & Sp. \\
\midrule
ARC-Easy       & 90.04 & 0 & 1.00 & \textbf{94.40} & 3 & 1.17 & 90.08 & 7 & 1.50 \\
ARC-Challenge  & 86.55 & 0 & 1.00 & \textbf{92.00} \textcolor{green}{(+5.45\% $\uparrow$)} & 2 & 1.30 & 86.55 & 6 & 1.37 \\
BoolQ          & 81.90 & 0 & 1.00 & 83.90 & 4 & 1.30 & 80.00 & 7 & 1.45 \\
MMLU           & 68.10 & 0 & 1.00 & 71.00 & 5 & 1.24 & 68.13 & 6 & 1.31 \\
COMMONQA       & 80.30 & 0 & 1.00 & 84.40 & 2 & 1.12 & 80.50 & 6 & 1.27 \\
WINOGRANDE     & 62.04 & 0 & 1.00 & 67.25 & 3 & 1.16 & 62.19 & 6 & 1.48 \\
BIG-Bench      & 79.20 & 0 & 1.00 & 81.60 & 6 & 1.47 & 81.60 & 6 & 1.47 \\
GSM8K-HARD     & - & - & - & - & - & - & - & - & - \\
GSM8K*         & - & - & - & - & - & - & - & - & - \\
\bottomrule
\end{tabular}}
\caption{Results of \textbf{Qwen 2.5 7B} across nine benchmarks. We report accuracy (\%), number of layers dropped, and relative inference speedup. 
Best accuracy per dataset is in \textbf{bold}. 
\textcolor{green}{(+X\% $\uparrow$)} shows accuracy improvement from baseline, 
\textcolor{red}{(-X\% $\downarrow$)} shows accuracy decline.}
\label{tab:qwen_results}
\end{table}
}


\hidden{

\definecolor{rowblue}{RGB}{203, 206, 251}
\begin{table}[!ht]
\centering
\renewcommand{\arraystretch}{1.2}
\resizebox{\linewidth}{!}{
\begin{tabular}{l|c|ccc|ccc|c|ccc|ccc}
\toprule
\multirow{2}{*}{\textbf{Dataset}} & 
\multicolumn{7}{c|}{\textbf{LLaMA 3.1 8B}} & 
\multicolumn{7}{c}{\textbf{Qwen 2.5 7B}} \\
\cmidrule(lr){2-8} \cmidrule(lr){9-15}
 & \multicolumn{1}{c|}{Baseline} & \multicolumn{3}{c|}{Best Model} & \multicolumn{3}{c|}{BSBA} & 
   \multicolumn{1}{c|}{Baseline} & \multicolumn{3}{c|}{Best Model} & \multicolumn{3}{c}{BSBA} \\
\cmidrule(lr){2-2} \cmidrule(lr){3-5} \cmidrule(lr){6-8} \cmidrule(lr){9-9} \cmidrule(lr){10-12} \cmidrule(lr){13-15}
 & Perf. & Perf. & \#D & Sp. & Perf. & \#D & Sp. & Perf. & Perf. & \#D & Sp. & Perf. & \#D & Sp. \\
\midrule
ARC-Easy       & 87.00 & \textbf{90.55} & 5 & 1.27 & 85.09 & 9 & 1.48 & 90.04 & \textbf{94.40} & 3 & 1.17 & 90.08 & 7 & 1.50 \\
ARC-Challenge  & 75.86 & \textbf{78.62} & 4 & 1.26 & 76.90 & 7 & 1.41 & 86.55 & \textbf{92.00} & 2 & 1.30 & 86.55 & 6 & 1.37 \\
BoolQ          & 85.00 & \textbf{86.20} & 3 & 1.10 & 83.00 & 8 & 1.36 & 81.90 & 83.90 & 4 & 1.30 & 80.00 & 7 & 1.45 \\
MMLU           & 54.87 & \textbf{59.90} & 1 & 1.05 & 54.87 & 9 & 1.37 & 68.10 & 71.00 & 5 & 1.24 & 68.13 & 6 & 1.31 \\
COMMONQA       & 72.20 & 75.30 & 3 & 1.21 & 70.10 & 8 & 1.34 & 80.30 & 84.40 & 2 & 1.12 & 80.50 & 6 & 1.27 \\
WINOGRANDE     & 53.83 & 56.67 & 4 & 1.25 & 53.83 & 12 & 1.47 & 62.04 & 67.25 & 3 & 1.16 & 62.19 & 6 & 1.48 \\
BIG-Bench      & 75.20 & 83.60 & 5 & 1.24 & 75.20 & 11 & 1.58 & 79.20 & 81.60 & 6 & 1.47 & 81.60 & 6 & 1.47 \\
GSM8K-HARD     & 15.07 & 38.02 & 1 & 1.12 & 17.31 & 6 & 1.35 & - & - & - & - & - & - & - \\
GSM8K*         & 42.15 & 56.56 & 1 & 34\% & 48.14 & 3 & -- & - & - & - & - & - & - & - \\
\bottomrule
\end{tabular}}
\caption{Comparison of \textbf{LLaMA 3.1 8B} and \textbf{Qwen 2.5 7B} across nine benchmarks. 
We report accuracy (\%), number of layers dropped, and relative inference speedup. 
Best accuracy per dataset is in \textbf{bold}. 
\textcolor{green}{(+X\% $\uparrow$)} shows accuracy improvement from baseline, 
\textcolor{red}{(-X\% $\downarrow$)} shows accuracy decline.}
\label{tab:main_results}
\end{table}

\definecolor{rowblue}{RGB}{203, 206, 251}

\begin{table}[!ht]
\centering
\renewcommand{\arraystretch}{1.2} 

\resizebox{\linewidth}{!}{
\begin{tabular}{l|c|ccc|ccc|c|ccc|ccc}
\toprule
\multirow{3}{*}{\textbf{Dataset}} & 
\multicolumn{7}{c|}{\textbf{LLaMA 3.1 8B}} & 
\multicolumn{7}{c}{\textbf{Qwen 2.5 7B}} \\
\cmidrule(lr){2-8} \cmidrule(lr){9-15}
 & {Baseline} & \multicolumn{3}{c|}{Best Model} & \multicolumn{3}{c|}{BSBA} 
 & {Baseline} & \multicolumn{3}{c|}{Best Model} & \multicolumn{3}{c}{BSBA} \\
\cmidrule(lr){2-2} \cmidrule(lr){3-5} \cmidrule(lr){6-8}
\cmidrule(lr){9-9} \cmidrule(lr){10-12} \cmidrule(lr){13-15}
& Perf. & Perf. & \#D & Sp. & Perf. & \#D & Sp. 
& Perf. & Perf. & \#D & Sp. & Perf. & \#D & Sp. \\
\midrule
ARC-Easy       & 87.00  & \textbf{90.55} \textcolor{green}{(+3.55\%)} & 5 & 1.27 & 85.09 & 9 & 1.48 & 
90.04 & \textbf{94.40} & 3 & 1.17 & 90.08 & 7 & 1.5 \\
ARC-Challenge  & 75.86  & \textbf{78.62} \textcolor{green}{(+2.76\%)} & 4 & 1.26 & 76.90 & 7 & 1.41 & 
86.55 & \textbf{92.00} \textcolor{green}{(+5.45\%)} & 2 & 1.3 & 86.55 & 6 & 1.37 \\
BoolQ          & 85.00  & \textbf{86.20} \textcolor{green}{(+1.20\%)} & 3 & 1.10 & 83.00 & 8 & 1.36 &
81.9 & 83.9 & 4 & 1.3 & 80.00 & 7 & 1.45 \\
MMLU           & 54.87  & \textbf{59.90} \textcolor{green}{(+5.03\%)} & 1 & 1.05 & 54.87 & 9 & 1.37 &
68.10 & 71.00 & 5 & 1.24 & 68.13 & 6 & 1.31 \\
COMMONQA       & 72.20 & \textbf{75.30} \textcolor{green}{(+3.10\%)} & 3 & 1.21 & 70.10 & 8 & 1.34 &
80.30 & 84.4 & 2 & 1.12 & 80.50 & 6 & 1.27 \\
WINOGRANDE     & 53.83 & \textbf{56.67} \textcolor{green}{(+2.84\%)} & 4 & 1.25 & 53.83 & 12 & 1.47 & 
62.04 & 67.25 & 3 & 1.16 & 62.19 & 6 & 1.48 \\
BIG-Bench      & 75.20 & \textbf{83.60} \textcolor{green}{(+8.40\%)} & 5 & 1.24 & 75.20 & 11 & 1.58 &
79.2 & 81.60 & 6 & 1.47 & 81.60 & 6 & 1.47 \\
GSM8K-HARD     & 15.07 & \textbf{38.02} \textcolor{green}{(+22.95\%)} & 1 & 1.12 & 17.31 & 6 & 1.35 &
- & - & - & - & - & - & - \\
GSM8K*         & 42.15 & \textbf{56.56} \textcolor{green}{(+14.41\%)} & 1 & -- & 48.14 & 3 & -- &
- & - & - & - & - & - & - \\
\bottomrule
\end{tabular}
}
\caption{Comparison of \textbf{LLaMA 3.1 8B} and \textbf{Qwen 2.5 7B} across nine benchmarks. 
We report accuracy (\%), number of layers dropped, and relative inference speedup. 
Best accuracy per dataset is in \textbf{bold}. 
\textcolor{green}{(+X\%)} shows accuracy improvement from baseline.}
\label{tab:main_results}
\end{table}

}

\definecolor{rowblue}{RGB}{203, 206, 251}

\hidden{
\begin{table}[!ht]
\centering
\renewcommand{\arraystretch}{1.2}

\resizebox{\linewidth}{!}{
\begin{tabular}{l|ccc|ccc|ccc|ccc|ccc|ccc}
\toprule
\multirow{2}{*}{\textbf{Dataset}} & 
\multicolumn{9}{c|}{\textbf{LLaMA 3.1 8B}} & 
\multicolumn{9}{c}{\textbf{Qwen 2.5 7B}} \\
\cmidrule(lr){2-10} \cmidrule(lr){11-19}
 & \multicolumn{3}{c|}{Baseline} & \multicolumn{3}{c|}{Best Model} & \multicolumn{3}{c|}{BSBA} & 
   \multicolumn{3}{c|}{Baseline} & \multicolumn{3}{c|}{Best Model} & \multicolumn{3}{c}{BSBA} \\
\cmidrule(lr){2-4} \cmidrule(lr){5-7} \cmidrule(lr){8-10} \cmidrule(lr){11-13} \cmidrule(lr){14-16} \cmidrule(lr){17-19}
 & Perf. & \#D & Sp. & Perf. & \#D & Sp. & -- & -- & -- & Perf. & \#D & Sp. & Perf. & \#D & Sp. & Perf. & \#D & Sp. \\
\midrule
ARC-Easy       & 87.00  & - & - & \textbf{90.55} & 5 & 1.27 & 85.09 & 9 & 1.48 & 90.04 & 0 & 1.00 & \textbf{94.40} & 3 & 1.17 & 90.08 & 7 & 1.50 \\
ARC-Challenge  & 75.86  & - & - & \textbf{78.62} & 4 & 1.26 & 76.90 & 7 & 1.41 & 86.55 & 0 & 1.00 & \textbf{92.00} & 2 & 1.30 & 86.55 & 6 & 1.37 \\
BoolQ          & 85.00  & - & - & \textbf{86.20} & 3 & 1.10 & 83.00 & 8 & 1.36 & 81.90 & 0 & 1.00 & 83.90 & 4 & 1.30 & 80.00 & 7 & 1.45 \\
MMLU           & 54.87  & - & - & \textbf{59.90} & 1 & 1.05 & 54.87 & 9 & 1.37 & 68.10 & 0 & 1.00 & 71.00 & 5 & 1.24 & 68.13 & 6 & 1.31 \\
COMMONQA       & 72.20  & - & - & 75.30 & 3 & 1.21 & 70.10 & 8 & 1.34 & 80.30 & 0 & 1.00 & 84.40 & 2 & 1.12 & 80.50 & 6 & 1.27 \\
WINOGRANDE     & 53.83  & - & - & 56.67 & 4 & 1.25 & 53.83 & 12 & 1.47 & 62.04 & 0 & 1.00 & 67.25 & 3 & 1.16 & 62.19 & 6 & 1.48 \\
BIG-Bench      & 75.20  & - & - & 83.60 & 5 & 1.24 & 75.20 & 11 & 1.58 & 79.20 & 0 & 1.00 & 81.60 & 6 & 1.47 & 81.60 & 6 & 1.47 \\
GSM8K-HARD     & 15.07  & - & - & 38.02 & 1 & 1.12 & 17.31 & 6 & 1.35 & - & - & - & - & - & - & - & - & - \\
GSM8K*         & 42.15  & - & - & 56.56 & 1 & -- & 48.14 & 3 & -- & - & - & - & - & - & - & - & - & - \\
\bottomrule
\end{tabular}}

\caption{Comparison of \textbf{LLaMA 3.1 8B} and \textbf{Qwen 2.5 7B} across nine benchmarks. 
We report accuracy (\%), number of layers dropped, and relative inference speedup. 
Best accuracy per dataset is in \textbf{bold}. 
\textcolor{green}{(+X\% $\uparrow$)} shows accuracy improvement from baseline, 
\textcolor{red}{(-X\% $\downarrow$)} shows accuracy decline.}
\label{tab:main_results}
\end{table}

}


\section{Scores with relative increases in accuracy}

 \begin{table*}[!ht]
 \centering
 \renewcommand{\arraystretch}{1.3}
 \setlength{\tabcolsep}{6pt}

 \begin{subtable}[t]{\textwidth}
 \centering
 \resizebox{\linewidth}{!}{
 \begin{tabular}{l|c|ccc|ccc|c|ccc|ccc|}
 \toprule
 \rowcolor{teal!30}
 \multirow{2}{*}{\textbf{Dataset}} & 
 \multicolumn{7}{c|}{\textbf{LLaMA 3.1 8B (0-shot)}} & 
 \multicolumn{7}{c}{\textbf{Qwen 2.5 7B (0-shot)}}\\
 \cmidrule(lr){2-8} \cmidrule(lr){9-15} 
  & Baseline & \multicolumn{3}{c|}{Best Model} & \multicolumn{3}{c|}{BSBA} 
  & Baseline & \multicolumn{3}{c|}{Best Model} & \multicolumn{3}{c}{BSBA} \\
 \cmidrule(lr){2-2} \cmidrule(lr){3-5} \cmidrule(lr){6-8} 
 \cmidrule(lr){9-9} \cmidrule(lr){10-12} \cmidrule(lr){13-15}
  & Perf. & Perf. & \#D & Sp. & Perf. & \#D & Sp. 
  & Perf. & Perf. & \#D & Sp. saved & Perf. & \#D & Sp. \\
 \midrule
ARC-Easy        & 87.00 & \textbf{90.55}\perc{4.08} & 5 & \percd{14.6} & 87.82 & 8 & \percd{23.5} & 91.01 & \textbf{91.82}\perc{0.89} & 2 & \percd{10.0} & 90.91 & 5 & \percd{30.3} \\
 ARC-Challenge   & 75.86 & \textbf{78.62}\perc{3.63} & 4 & \percd{11.7} & 76.90 & 7 & \percd{20.5} & 86.55 & \textbf{92.00}\perc{6.45} & 2 & \percd{6.7} & 86.55 & 6 & \percd{19.9} \\
 BoolQ           & 85.00 & \textbf{86.20}\perc{1.40} & 3 & \percd{8.8} & 85.70 & 7 & \percd{17.6} & 84.10 & \textbf{86.90}\perc{3.22} & 4 & \percd{13.3} & 82.70 & 5 & \percd{23.2} \\
 MMLU            & 54.87 & \textbf{59.90}\perc{9.17} & 1 & \percd{2.9} & 54.87 & 9 & \percd{26.4} & 68.10 & \textbf{71.00}\perc{4.26} & 5 & \percd{16.6} & 68.13 & 6 & \percd{19.9} \\
 CommonQA        & 72.20 & \textbf{75.30}\perc{4.29} & 3 & \percd{8.8} & 73.10 & 6 & \percd{17.6} & 80.30 & \textbf{84.40}\perc{5.11} & 2 & \percd{6.6} & 80.50 & 6 & \percd{19.9} \\
 Winogrande      & 53.83 & \textbf{56.67}\perc{5.28} & 4 & \percd{11.7} & 53.83 & 12 & \percd{32.2} & 62.04 & \textbf{67.25}\perc{8.40} & 3 & \percd{10.0} & 62.19 & 6 & \percd{19.9} \\
 BIG-Bench       & 75.20 & \textbf{83.60}\perc{11.17}& 5 & \percd{14.4} & 75.20 & 11 & \percd{32.2} & 79.20 & \textbf{81.60}\perc{3.03} & 6 & \percd{19.9} & 81.60 & 6 & \percd{19.9} \\
 GSM8K-HARD      & 15.07 & \textbf{37.08}\perc{146.05} & 1 & \percd{2.9} & 35.0 & 4 & \percd{11.7} & 7.9 & \textbf{27.0}\perc{243.58} & 2 & \percd{6.6} & 19.1 & 4 & \percd{13.3} \\
 Math500         & 20.50 & \textbf{26.00}\perc{26.83} & 1 & \percd{2.9} & 26.00 & 3 & \percd{8.8} & 18.00 & \textbf{27.00}\perc{50.0} & 2 & \percd{6.6} & 21.00 & 4 & \percd{13.3} \\
 \bottomrule
 \end{tabular}
 }
 \end{subtable}

 \vspace{0.5cm}

 \vspace{0.5cm}

 \begin{subtable}[t]{\textwidth}
 \centering
 \resizebox{\linewidth}{!}{
 \begin{tabular}{l|c|ccc|ccc|c|ccc|ccc|}
 \toprule
 \rowcolor{teal!30}
 \multirow{2}{*}{\textbf{Dataset}} & 
 \multicolumn{7}{c|}{\textbf{Lucie 7B (0-shot)}} & 
 \multicolumn{7}{c}{\textbf{Mistral 7B (0-shot)}}\\
 \cmidrule(lr){2-8} \cmidrule(lr){9-15}
  & Baseline & \multicolumn{3}{c|}{Best Model} & \multicolumn{3}{c|}{BSBA} 
  & Baseline & \multicolumn{3}{c|}{Best Model} & \multicolumn{3}{c}{BSBA} \\
 \cmidrule(lr){2-2} \cmidrule(lr){3-5} \cmidrule(lr){6-8}
 \cmidrule(lr){9-9} \cmidrule(lr){10-12} \cmidrule(lr){13-15}
  & Perf. & Perf. & \#D & Sp. & Perf. & \#D & Sp.
  & Perf. & Perf. & \#D & Sp. & Perf. & \#D & Sp. \\
 \midrule
 ARC-Easy        & 72.45 & \textbf{76.55}\perc{5.66} & 6 & \percd{18.1} & 72.55 & 13 & \percd{39.2} & 81.02 & \textbf{83.45}\perc{4.23} & 5 & \percd{15.4} & 81.09 & 9 & \percd{27.7} \\
 ARC-Challenge   & 48.00 & \textbf{53.79}\perc{12.06} & 7 & \percd{21.1} & 51.38 & 11 & \percd{33.1} & 72.20 & \textbf{74.83}\perc{3.64} & 6 & \percd{18.5} & 72.41 & 8 & \percd{24.6} \\
 BoolQ           & 53.70 & \textbf{77.50}\perc{44.32} & 5 & \percd{17.2} & 60.60 & 19 & \percd{54.2} & 80.36 & \textbf{83.20}\perc{3.53} & 6 & \percd{18.5} & 80.60 & 10 & \percd{27.7} \\
 MMLU            & 21.36 & \textbf{42.98}\perc{101.2} & 8 & \percd{24.1} & 39.39 & 15 & \percd{45.2} & 52.73 & \textbf{57.81}\perc{9.63} & 2 & \percd{6.2} & 52.91 & 8 & \percd{24.6} \\
 CommonQA        & 55.50 & \textbf{69.70}\perc{25.59} & 3 & \percd{9.1} & 57.10 & 17 & \percd{48.2} & 57.32 & \textbf{61.40}\perc{7.12} & 4 & \percd{12.3} & 57.40 & 7 & \percd{21.5} \\
 Winogrande      & 54.20 & \textbf{57.80}\perc{6.64} & 5 & \percd{27.1} & 54.30 & 15 & \percd{45.2} & 52.55 & \textbf{58.80}\perc{11.53} & 10 & \percd{30.7} & 53.43 & 13 & \percd{40.0} \\
 BIG-Bench       & 69.60 & \textbf{77.20}\perc{9.84} & 9 & \percd{27.1} & 72.00 & 15 & \percd{45.1} & 70.00 & \textbf{76.40}\perc{9.14} & 9 & \percd{28.0} & 72.80 & 11 & \percd{33.8} \\
 GSM8K-HARD      & 14.20 & \textbf{17.80}\perc{25.35} & 1 & \percd{3.1} & 17.40 & 3 & \percd{9.1} & 11.24 & \textbf{19.10}\perc{69.92} & 2 & \percd{6.2} & 15.73 & 4 & \percd{12.3} \\
 Math500         & 19.00 & \textbf{27.00}\perc{42.11} & 2 & \percd{6.0}  & 26.00 & 3 & \percd{9.1}  & 8.00 & \textbf{16.00}\perc{100}  & 1 & \percd{3.1} & 10.00 & 4 & \percd{12.3} \\
 \bottomrule
 \end{tabular}
 }
 \end{subtable}

\begin{subtable}[t]{\textwidth}
\centering
\resizebox{0.6\linewidth}{!}{
\begin{tabular}{l|c|ccc|ccc|}
\toprule
\rowcolor{teal!30}
\multirow{2}{*}{\textbf{Dataset}} & 
\multicolumn{7}{c}{\textbf{Qwen 2.5 0.5B (0-shot)}} \\
\cmidrule(lr){2-8}
 & Baseline & \multicolumn{3}{c|}{Best Model} & \multicolumn{3}{c}{BSBA} \\
\cmidrule(lr){2-2} \cmidrule(lr){3-5} \cmidrule(lr){6-8}
 & Perf. & Perf. & \#D & Sp. & Perf. & \#D & Sp. \\
\midrule
ARC-Easy        & 40.00 & \textbf{60.91}\perc{48.49} & 3 & 1.1 & 48.36 & 5 & 1.4 \\
ARC-Challenge   & 35.52 & \textbf{40.34}\perc{13.57} & 1 & 1.1 & 37.24 & 4 & 1.5 \\
BoolQ           & 62.30 & \textbf{67.20}\perc{7.87}  & 5 & 1.4 & 66.20 & 6 & 1.5 \\
MMLU            & 31.48 & \textbf{39.97}\perc{26.96} & 2 & 1.1 & 33.90 & 5 & 1.4 \\
CommonQA        & 42.40 & \textbf{49.10}\perc{15.80} & 2 & 1.3 & 44.00 & 3 & 1.4 \\
Winogrande      & 49.86 & \textbf{51.88}\perc{4.51}  & 5 & 1.3 & 49.87 & 17 & 3.9 \\
BIG-Bench       & 72.40 & \textbf{73.60}\perc{1.66}  & 2 & 1.2 & 73.60 & 2 & 1.2 \\
GSM8K-HARD      & 6.74  & \textbf{11.24}\perc{66.77} & 1 & 1.2 & 8.99  & 2 & 1.2 \\
Math500         & 8.00  & \textbf{12.00}\perc{50.00} & 1 & 1.1 & 9.00  & 2 & 1.1 \\
\bottomrule
\end{tabular}
}
\end{subtable}

 \caption{Performance comparison for LLaMA 3.1 8B, Qwen 2.5 7B, Lucie 7b, Mistral 7b and Qwen 2.5 0.5B under 0-shot evaluation on the 9 benchmarks with a different training test split from that used for Table \ref{tab:robustness_study}).  We report accuracy
(\%), number of layers dropped, and relative inference
speed in time.}
 \label{tab:combined_model_comparison_placeholder}
 \end{table*}

\end{document}